\documentclass[twoside,11pt]{article}

\usepackage{blindtext}

\usepackage{jmlr2e}

\usepackage{microtype}
\usepackage{graphicx}
\usepackage{subfigure}
\usepackage{booktabs} 

\usepackage{hyperref}

\usepackage{amssymb}
\usepackage{mathtools}
\usepackage{enumitem}

\usepackage{xspace}

\usepackage{bbm}
\usepackage{dsfont}

\renewcommand{\d}[1]{\ensuremath{\operatorname{d}\!{#1}}}

\newcommand{\ie}{\textit{i.e.}\xspace}
\newcommand{\eg}{\textit{e.g.}\xspace}
\newcommand{\FancyName}{SuaVE\xspace}

\usepackage[textsize=tiny]{todonotes}

\usepackage{mathtools}

\DeclareMathOperator{\vect}{vec}

\makeatletter
 \def\namedlabel#1#2{\begingroup
    #2%
    \def\@currentlabel{#2}%
    \phantomsection\label{#1}\endgroup
}
\makeatother

\usepackage{lastpage}
\jmlrheading{27}{2026}{1-\pageref{LastPage}}{8/23; Revised
5/25}{1/26}{23-1023}{Yuhang Liu, Zhen Zhang, Dong Gong, Mingming Gong, Biwei Huang, Anton van den Hengel, Kun Zhang, Javen Qinfeng Shi}
\ShortHeadings{Identifying Weight-Variant Latent Causal Models}{Liu, Zhang, Gong, Gong, Huang, Hengel, Zhang and Shi}
\firstpageno{1}

\begin{document}

\title{Identifying Weight-Variant Latent Causal Models}
\author{\name Yuhang Liu$^{1,2}$ \email yuhang.liu01@adelaide.edu.au \\
\name Zhen Zhang$^{1,2}$ \email zhen.zhang02@adelaide.edu.au \\
\name Dong Gong$^{3}$ \email edgong01@gmail.com \\
\name Mingming Gong$^{4}$ \email mingming.gong@unimelb.edu.au \\
\name Biwei Huang$^{5}$ \email bih007@ucsd.edu \\
\name Anton van den Hengel$^{1,2}$ \email anton.vandenhengel@adelaide.edu.au \\
\name Kun Zhang$^{6}$ \email kunz1@cmu.edu \\
\name Javen Qinfeng Shi$^{1,2}$ \email javen.shi@adelaide.edu.au \\[0.8ex]
\addr $^{1}$ Responsible AI Research Centre, Australia \\
\addr $^{2}$ Australian Institute for Machine Learning, Adelaide University, Australia \\
\addr $^{3}$ School of Computer Science and Engineering, The University of New South Wales, Australia \\
\addr $^{4}$ School of Mathematics and Statistics, The University of Melbourne, Australia \\
\addr $^{5}$ Halicioğlu Data Science Institute, University of California San Diego, USA \\
\addr $^{6}$ Department of Philosophy, Carnegie Mellon University, USA
}
\editor{Eric Laber}

\maketitle

\begin{abstract}
The task of causal representation learning aims to uncover latent higher-level causal variables that affect lower-level observations. Identifying the true latent causal variables from observed data, while allowing instantaneous causal relations among latent variables, remains a challenge, however. To this end, we start with the analysis of three intrinsic indeterminacies in identifying latent variables from observations: transitivity, permutation indeterminacy, and scaling indeterminacy. We find that transitivity acts as a key role in impeding the identifiability of latent causal variables. To address the unidentifiable issue due to transitivity, we introduce a novel identifiability condition where the underlying latent causal model satisfies a linear-Gaussian model, in which the causal coefficients and the distribution of Gaussian noise are modulated by an additional observed variable. Under certain assumptions, including the existence of a reference condition under which latent causal influences vanish, we can show that the latent causal variables can be identified up to trivial permutation and scaling, and that partial identifiability results can still be obtained when this reference condition is violated for a subset of latent variables. Furthermore, based on these theoretical results, we propose a novel method, termed Structural caUsAl Variational autoEncoder (\FancyName), which directly learns causal representations and causal relationships among them, together with the mapping from the latent causal variables to the observed ones. Experimental results on synthetic and real data demonstrate the identifiability and consistency results and the efficacy of \FancyName in learning causal representations. \paragraph{Project Page:} \href{https://sites.google.com/view/yuhangliu/projects/suave}{\texttt{https://sites.google.com/view/yuhangliu/projects/suave}}
\end{abstract}


\section{Introduction} \label{sec:intro}
While there is no universal formal definition, one widely accepted feature of disentangled representations is that a change in one dimension corresponds to a change in one factor of variation in the underlying model of the data, while having little effect on others \citep{bengio2013representation}. The underlying model is rarely available for interrogation, however, which makes learning disentangled representations challenging. Several excellent works for disentangled representation have been proposed that focus on enforcing independence over the latent variables that control the factors of variation \citep{higgins2016beta,chen2018isolating,locatello2019challenging,kim2018disentangling,locatello2020weakly}. In many applications, however, the latent variables are not statistically independent, which is at odds with the notion of disentanglement, \ie, foot length and body height exhibit strong correlation in the observed data \citep{trauble2021disentangled}.

Causal representation learning avoids the aforementioned limitation, as it aims to learn a representation that exposes the unknown high-level latent causal variables, and the relationships between them, from a set of low-level observations~\citep{scholkopf2021toward}. Unlike disentangled representation learning, it allows the possible causal relations between latent variables. In fact, disentangled representation learning can be viewed as a special case of causal representation learning where the latent variables have no causal influences~\citep{scholkopf2021toward}. One of the most prominent additional capabilities of causal representations is the ability to represent interventions and to make predictions regarding such interventions \citep{Pearl00}, which enables the generation of new samples that do not lie within the distribution of the observed data. This can be particularly useful to improve the generalization of the resulting model. Causal representations also enable answering counterfactual questions, \eg, would a given patient have suffered heart failure if they had started exercising a year earlier? 

Despite its advantages, causal representation learning is a notoriously hard problem. Without certain assumptions, identifying the true latent causal model from observed data is generally not possible. There are four primary approaches to achieve identifiability: 1) adapting (weakly) supervised methods with given latent causal graphs or/and labels \citep{kocaoglu2018causalgan,yang2020causalvae}, 2) imposing special graphical conditions with bottleneck graphical conditions \citep{adams2021identification,xie2020generalized,lachapelle2021disentanglement}, 3) using temporal information \citep{yao2021learning,lippe2022citris}, 4) using hard interventional data \citep{ahuja2023interventional,seigal2022linear}. A brief review is provided in Section \ref{sec:rw}. In supervised approaches, when either the labels for latent variables or the true latent causal graph are available, the challenging identifiability problem in the latent space can be alleviated, or even bypassed by leveraging existing identifiability results in observed sapce \citep{zhang2012identifiability, peters14a}. However, such methods tend to rely heavily on domain-specific prior knowledge and are often associated with high annotation costs. For the second approach, many true latent causal graphs do not satisfy the assumed special graph structure. The temporal approach is only applicable when temporal information or temporal intervened information between latent factors is available. For using intervention data, most studies focus on hard interventions, particularly single-node and paired hard interventions. In contrast, our work considers the changes of causal influences, which is more aligned with soft interventions.

In this work, we explore a new direction in the identifiability of latent causal variables by examining the changes in causal influences between latent causal variables. Our approach is motivated by recent advances in nonlinear ICA, which, broadly speaking, demonstrate that latent independent variables can be identified when an additional observed variable $\mathbf{u}$ is introduced to modulate their distributions \citep{hyvarinen2019nonlinear, khemakhem2020variational}. This raises a natural question: 

\emph{Beyond the independent latent variables in nonlinear ICA, with causal relationships between latent variables, what additional assumptions are required for identifiability of latent causal variables?}
\begin{itemize}[leftmargin=*]
    \item To answer the question above, we begin by analyzing three intrinsic indeterminacies in the latent space (see Section \ref{sec:indeterminacies}): transitivity, permutation indeterminacy, and scaling indeterminacy. This analysis leads to the following key insights. 1) Transitivity is the scourge of identifiability of latent causal models. 2) Permutation indeterminacy nature enables us to pre-define a causal order, avoiding troublesome directed acyclic graph (DAG) constraints. 3) Scaling and permutation indeterminacy only allow recovering latent causal variables up to permutation and scaling, not the exact values.
    \item To overcome the challenge of transitivity, we model the underlying causal generative process with weight-variate linear Gaussian models (see Section \ref{sec:identifiability}). In this model, both the weights (\ie, causal influences) and the mean and variance of the Gaussian noise are modulated by an additional observed variable $\mathbf{u}$.
    \item By exploring the changes in causal influences, i.e., the existence of a reference environment under which certain causal influences between latent variables vanish, we demonstrate that the latent causal variables can be recovered up to a trivial permutation and scaling. Combined with the independent causal mechanism, a common assumption in causality, we show that the latent causal structure is also identifiable. Further, realizing that requiring all weights to change may be challenging in real applications, we show partial identifiability results if only part of the weights change in Section \ref{sec:partial}.
    \item Based on our analysis, in Section \ref{sec:suave}, we additionally propose a novel method, Structural caUsAl Variational autoEncoder (SuaVE), for learning latent causal variables and latent causal graph among them. Section \ref{sec:exp} verifies the efficacy of the proposed method on both synthetic and real fMRI data.
\end{itemize}
   
The key to the identifiability results in this work lies in leveraging changes in the causal influences (\eg, weights in linear models) between latent causal variables. These changes correspond to distribution shifts, which can be interpreted as interventions acting on the latent variables. Such distribution shifts commonly occur across various domains, including medical imaging \citep{chandrasekaran2021image}, biogeography \citep{pinsky2020climate}, and finance \citep{gibbs2021adaptive}. Analogous to studies of distribution shifts in observed data \citep{ghassami2018multi, CDNOD_20}, shifts in the latent space enable comparative analyses that reveal the underlying causal mechanisms governing relationships between latent variables. Thus, investigating distribution shifts offers a promising avenue for achieving identifiability in causal representation learning.



\section{Related work} \label{sec:rw}

Due to the inherent challenges of identifiability in causal representation learning, most existing methods address this issue by imposing specific assumptions. We therefore provide a brief review of related work grounded on this basis.

\paragraph{(Weakly) Supervised Causal Representation Learning} Approaches in this category assume either known latent causal graphs or access to labels. For example, CausalGAN \citep{kocaoglu2018causalgan} requires prior knowledge of the latent causal graph structure, which limits its practical applicability. CausalVAE \citep{yang2020causalvae} relies on additional labels to supervise latent variable learning. However, such labels are often unavailable, and manual annotation can be costly and error-prone. \citet{von2021self} explore self-supervised causal representation learning using a known but complex causal graph between content and style factors. Meanwhile, \citet{brehmer2022weakly} propose a weakly supervised approach that assumes access to paired data before and after unknown interventions on the system.

\paragraph{Special Graphical Structure}~Most recent advances in identifiability focus on imposing special graphical structure constraints \citep{silva2006learning, shimizu2009estimation, anandkumar2013learning, frot2019robust, NEURIPS2019_8c66bb19, xie2020generalized, xie2022identification, kivva2021learning}. \citet{adams2021identification} provided a unifying perspective, advocating for sparser models that fit the observations better. However, their analysis mainly considers linear causal relationships both between latent causal variables and between latent and observed variables, which often violated in real-world complex scenarios. \citet{lachapelle2021disentanglement} addressed nonlinear causal relations by assuming special sparse graphical structures. Yet, latent causal graphs encountered in practice are more complex and may not conform to strict sparsity assumptions. In contrast, our approach defines a function class for latent variables without imposing restrictive constraints on the underlying graph structure. Additionally, \citet{kivva2021learning} focus on discrete latent causal variables within relatively limited graph structures, while our work considers continuous latent causal variables with flexible general graph configurations.


\paragraph{Temporal Information}~The temporal constraint that the effect cannot precede the cause has been used repeatedly in latent causal representation learning \citep{yao2021learning,lippe2022citris,yao2022learning}. For example, \citet{yao2021learning} recover latent causal variables and the relations between them using Variational AutoEncoders and enforcing constraints in causal process prior. \citet{lippe2022citris} learn causal representations from temporal sequences, which requires the underlying causal factors to be intervened. All of these works can be regarded as special cases of exploring the change of causal influences among latent variables in time series data. The proposed approach in this work tends to be more general, since the auxiliary variable $\mathbf{u}$ can represent not only time indices but also domain indices or virtually any additional side information.

\paragraph{Data from Hard Interventions} Very recently, there have been some works exploring interventional data \citep{brehmer2022weakly,ahuja2023interventional,seigal2022linear,buchholz2023learning,varici2023score}. Most of them consider hard intervention, and more restricted single-node and paired hard interventions \citep{ahuja2023interventional,seigal2022linear,buchholz2023learning,varici2023score,varici2024general,buchholz2024learning,von2024nonparametric}, which can only capture some special changes of causal influences. By contrast, this work studies unpaired data, and employs soft interventions to model a broader range of possible changes. This is especially important considering that distribution shifts in real applications may be arbitrary. From this perspective, soft interventions could be easier to achieve for latent variables than hard interventions. To clarify this distinction, we introduce hard and soft interventions in the following.

\paragraph{Preliminaries: Hard Intervention and Soft Intervention} Generally speaking, hard intervention (sometimes called perfect intervention) fixes a causal variable to a specific value, thereby breaking its dependence on all of its parent variables. In contrast, soft intervention modifies the distribution of a causal variable conditional on its parent nodes, such as by altering its parameters or noise distribution, without completely severing the causal influence from its parents. This added flexibility allows soft interventions to capture a broader and more realistic range of changes in causal influences, enabling the model to account for a broader array of changes in causal influences. This is particularly valuable for capturing changes in causal influences driven by uncontrollable behaviors, such as environmental fluctuations, within a causal system. Moreover, soft interventions generalize hard interventions, since a hard intervention can be viewed as a special case of a soft intervention where the distribution of the targeted causal variable collapses to a Dirac delta distribution.

\section{{Indeterminacies in Latent Causal Models}}
\label{sec:indeterminacies}
In this section, we first build a connection between nonlinear ICA and causal representation learning, by exploiting the correspondence between the independence of latent variables in nonlinear ICA and the independence of latent noise variables in causal systems. We then consider three indeterminacies in latent space: transitivity, permutation indeterminacy, and scaling indeterminacy, and analyze their impact on identifiability.

\subsection{Relating Causal Representation Learning with Identifiable Nonlinear ICA}
\begin{figure}[h]
\centering
\begin{minipage}[t]{0.8\textwidth}
\centering
\includegraphics[width=0.4\textwidth]{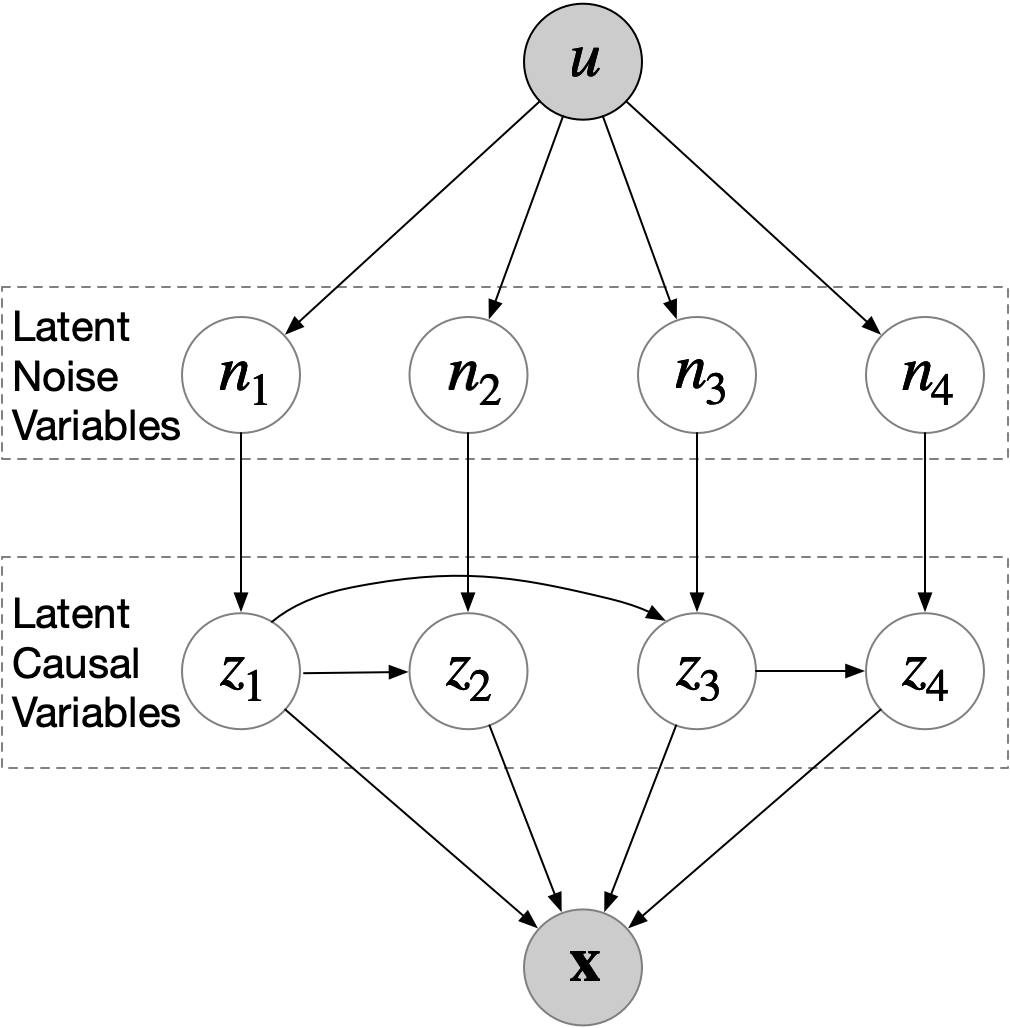}
\caption{Latent Causal Models with $\mathbf{u}$.}
\label{fig:problem}
\end{minipage}
\end{figure}
Causal representation learning aims to uncover latent higher-level causal variable that can explain lower-level raw observations. Specifically, we assume that the observed variables $\mathbf{x}$ are influenced by latent causal variables ${z_i}$, and the causal structure among ${z_i}$ can be any directed acyclic graph (which is unknown). For each latent causal variable $z_i$, there is a corresponding latent noise variable $n_i$, as shown in Figure \ref{fig:problem}, which represents some unmeasured factors that influence $z_i$. The latent noise variables $n_i$ are assumed to be independent with each other, \textit{conditional on the observed variable $\mathbf{u}$}\footnote{{For convenience in the later parts, with a slight abuse of definition, independent $n_i$ means that $n_i$ are mutually independent conditional on the observed variable $\mathbf{u}$.}}, in a causal system \citep{peters2017elements}, so it is natural to leverage recent progress in nonlinear ICA \citep{hyvarinen2019nonlinear,khemakhem2020variational}, which has shown that the independent latent noise variables $n_i$ are identifiable under relatively mild assumptions, \ie, one of the main assumptions is that $\{n_i\}$ are Gaussian distributed with their mean and variance modulated by an observed variable $\mathbf{u}$. Taking one step further, our goal is to recover the latent causal variables $z_i$. However, with the assumptions for the identifiability of latent noise variables $n_i$ from nonlinear ICA, it is still insufficient to identify the latent causal variables $z_i$. The reason of such non-identifiability will be given in Section \ref{sec:Transitivity}. Given this fact, a further question is what additional conditions are needed to recover the latent causal variables $z_i$. The corresponding identifiability conditions will be given in Section \ref{sec:identifiability}.

\paragraph{Understanding $\mathbf{u}$} Essentially, the introduced $\mathbf{u}$ as a surrogate variable is to characterize the changes of independent noise variables $n_i$ conditional on $\mathbf{u}$. In some cases, such a surrogate variable may be unavailable or hard to obtain in practice \citep{liu2021heterogeneous,creager2021environment}. Fortunately, it has been shown that learning or inferring such a surrogate variable is possible with additional auxiliary information \citep{lin2022zin}. These information is often cheaply available for every input in practice \citep{xie2020n,wang2020weakly}. Examples include time index of the data in time series forecasting tasks \citep{mudelsee2019trend}, locations (longitude and latitude) of collected satellite data in remote sensing \citep{russwurm2020meta}, modality index in multi-modality dataset, and label in natural images.

\subsection{Transitivity: the Challenge of Identifying Causal Representations} \label{sec:Transitivity}
\begin{figure} 
\centering
\begin{minipage}[t]{0.8\textwidth}
\centering
\includegraphics[width=0.4\textwidth]{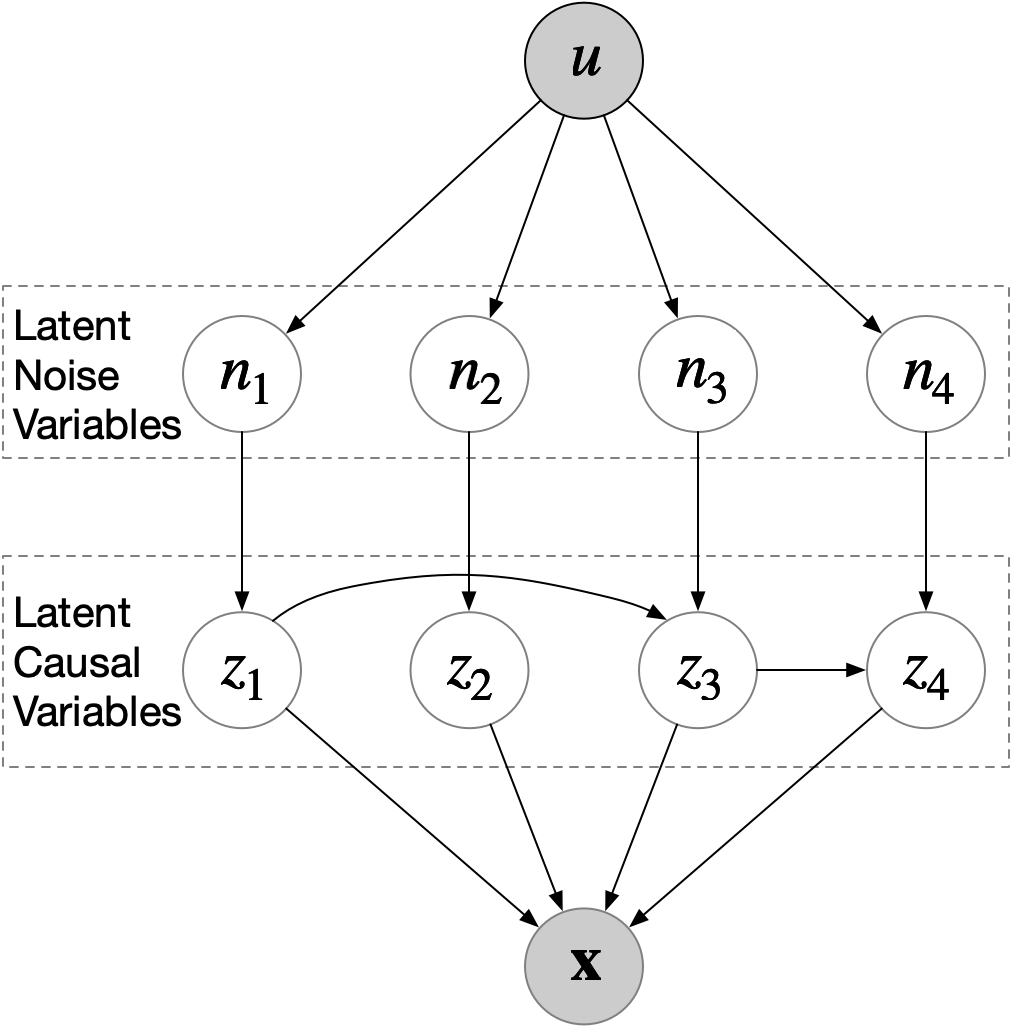}
\includegraphics[width=0.4\textwidth]{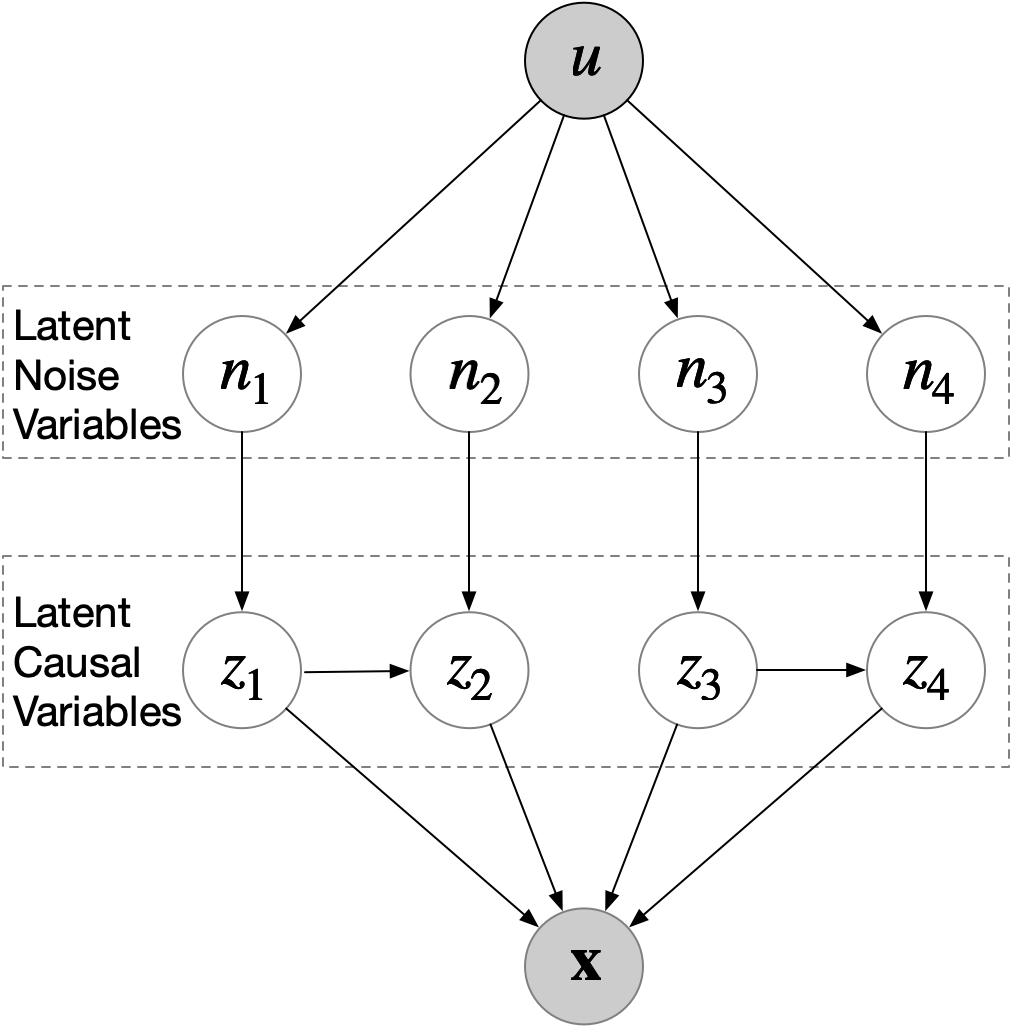}
\caption{Two equivalent structures.}
\label{fig:problem2}
\end{minipage}
\end{figure}

Even with the identifiable $n_i$, it is impossible to identify the latent causal variables $z_i$ without additional assumptions. To interpret this point, for simplicity, let us only consider the influences of $z_1$ and $z_2$ on $\mathbf{x}$ in Figure \ref{fig:problem}. According to the graph structure in Figure \ref{fig:problem}, assume that $z_1:=n_1$, $z_2:= z_1+n_2$ and $\mathbf{x}:= \mathbf{f}(z_1,z_2)$ (case 1). We then consider the graph structure shown in the left column of Figure \ref{fig:problem2}, where the edge $z_1 \rightarrow z_2$ has been removed, and assume that $z_1:=n_1$, $z_2:= n_2$ and $\mathbf{x}:= \mathbf{f} \circ \mathbf{g} (z_1,z_2)$ where $\mathbf{g} (z_1,z_2)=[z_1,z_1+z_2]$ (case 2). Interestingly, we find that the causal models in case 1 and case 2 generate the same observed data $\mathbf{x}$, which implies that there are two different causal models to interpret the same observed data. Clearly, $z_2$ in both the two equivalent causal structures are different, and thus $z_2$ is unidentifiable. 


Similarly, we can cut the edge $z_1 \rightarrow z_3$ in Figure \ref{fig:problem}, obtaining another equivalent causal graph as shown in the right of Figure \ref{fig:problem2}. That is, we can have two different $z_3$ to interpret the same observed data, and thus $z_3$ is unidentifiable. Since there exist many different equivalent causal structures, the latent causal variables in Figure \ref{fig:problem} are unidentifiable. Such a result is because the effect of $z_1$ on $z_2$ (or $z_3$) in latent space can be `absorbed' by the nonlinear function from $\mathbf{z}$ to $\mathbf{x}$. We term this indeterminacy transitivity in this work. We will show how to handle this challenge in Section \ref{sec:identifiability}. Before that, we introduce the other two indeterminacies in latent space, which assists in understanding our identifiability result.

\subsection{Scaling Indeterminacy in Causal Representations}
The scaling indeterminacy of latent causal variables is also an intrinsic indeterminacy in latent space. Again, for simplicity, we only consider the influence of $z_1$ and $z_2$ on $\mathbf{x}$ in Figure \ref{fig:problem}, and assume that $z_1:=n_1$, $z_2:= z_1+n_2$ and $\mathbf{x}:= \mathbf{f}(z_1,z_2)$. Under this setting, if the value of $z_1$ is scaled by $s$, \ie, $s \times z_1$, we can easily obtain the same observed data $\mathbf{x}$ by: 1) letting $z_2:= \frac{1}{s} \times z_1+n_2$ and 2) $\mathbf{x}:= \mathbf{f} \circ \mathbf{g} (z_1,z_2)$ where $\mathbf{g} (z_1,z_2)=[s \times z_1, z_1]$. This indeterminacy is because the scaling of the latent variables $z_i$ can be `absorbed' by the nonlinear function from $\mathbf{z}$ to $\mathbf{x}$ and the causal functions among the latent causal variables. Therefore, without additional information to determine the values of the latent causal variables $z_i$, it is only possible to identify the latent causal variable up to scaling, not exactly recovering the values. In general, this scaling does not affect identifying the causal structure among the latent causal variables. We will further discuss this point in Section \ref{sec:identifiability}.

\subsection{Permutation Indeterminacy in Causal Representations} \label{sec:permutation}
Due to the nature of ill-posedness, latent causal representation learning suffers from permutation indeterminacy, where the recovered latent causal representations have an arbitrary permutation. For example, assume that the underlying (synthetic) latent causal representations are corresponding to the size, color, shape, and location of an object, and we obtain the recovered latent causal representations $z_1,z_2,z_3,z_4$. Permutation indeterminacy means that we can not ensure that the recovered $z_1$ represents which specified semantic information, \eg, the size or the color. Therefore, without additional information, it is only possible to identify the latent causal representations up to permutation.

\begin{figure}[h]
\centering
\includegraphics[width=0.35\textwidth]{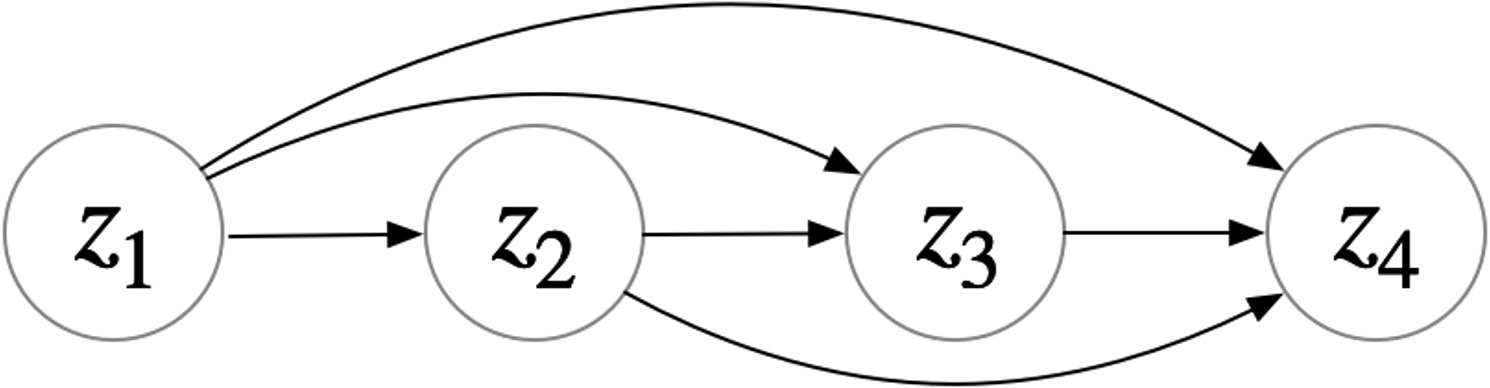}
  \caption{A Causal Fully-connected Graph, based on pre-defined causal order $z_1,z_2,z_3,z_4$, thanks to permutation indeterminacy in latent space.}
  \label{fig:Supergraph}
\end{figure}

However, the flexibility and ambiguity above in latent space also offer an advantage, i.e., by pre-defining a causal order, we can avoid constraints like those proposed in \citet{zheng2018dags}, facilitating the learning of latent causal representations and causal structure. For example, For example, assume that the true latent causal relations for generating data are: size $\rightarrow$ color $\rightarrow$ shape $\rightarrow$ location. During the learning process, we can pre-define a causal order $z_1,z_2,z_3,z_4$, without specifying semantic information for the nodes. Since $z_1$ is the first node in the predefined causal order, $z_1$ is enforced to the semantic information of the first node in the true underlying causal order, \ie, the size. Similarly, $z_2$ is enforced to the semantic information of the second node in the true underlying causal order, \ie, the color. Furthermore, the predefined causal order allows us to naturally establish a causal fully-connected graph as depicted in Figure \ref{fig:Supergraph}, ensuring the estimation of a directed acyclic graph (DAG) in learning causal representation and avoiding DAG constraints, such as those proposed by \citet{zheng2018dags}. We will further demonstrate how to implement this in the proposed method in Section \ref{sec:suave}.


\section{Identifiable Causal Representations with Weight-variant Linear Gaussian Models} \label{sec:identifiability}
\begin{figure}[h]
\centering
\includegraphics[width=0.35\textwidth]{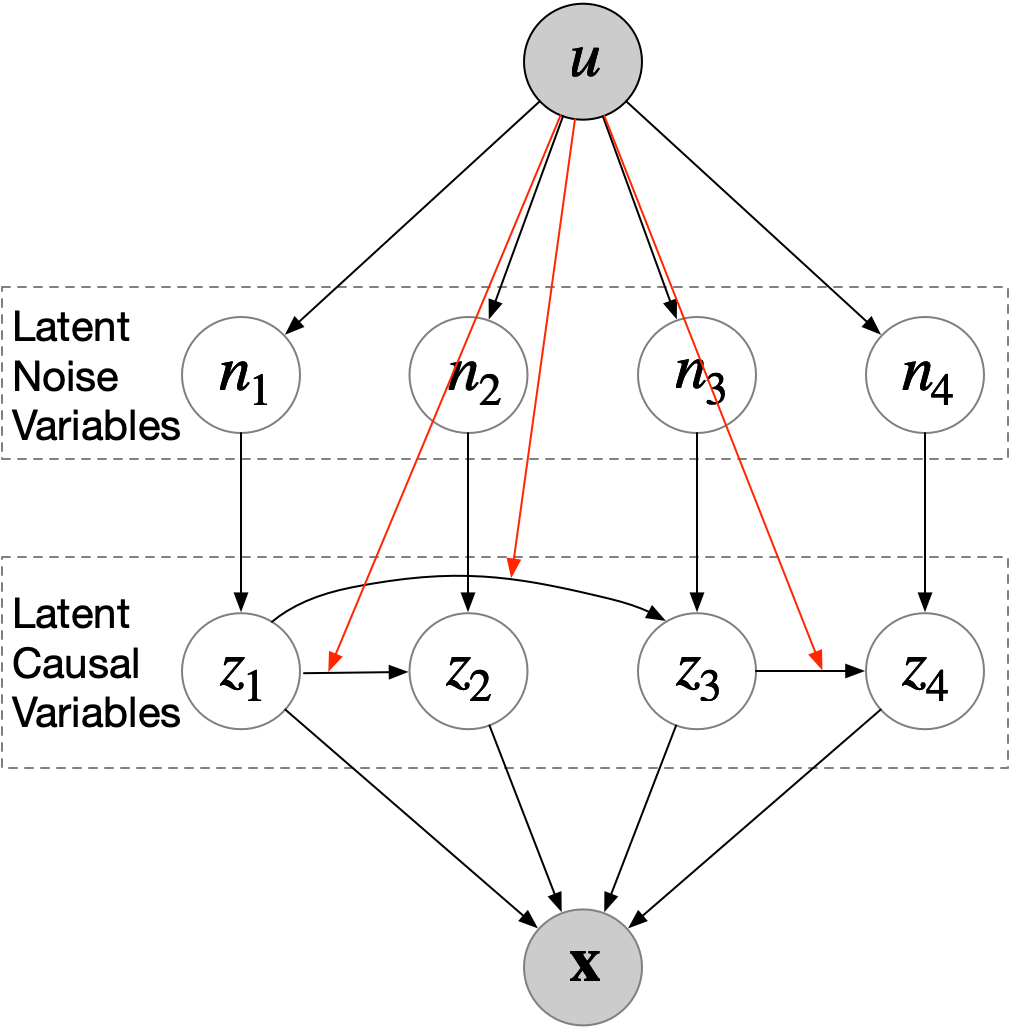}
  \caption{The proposed Latent Causal Models. Here we use the 'red' lines to indicate the changes of weights across $\mathbf{u}$, which is different from the definition of edges (\ie, causal direction) in standard causal model graphs.}
  \label{fig:problem3}
\end{figure}

As discussed in Section \ref{sec:Transitivity}, the key factor that impedes identifiable causal representations is the transitivity in latent space. Note that the transitivity is because the causal influences between the latent causal variables may be `absorbed' by the nonlinear mapping from latent variables $\mathbf{z}$ to the observed variable $\mathbf{x}$. To address this issue, motivated by identifying causal structures with the change of causal influences in observed space \citep{ghassami2018multi,SSM_Huang19, CDNOD_20}, we allow causal influences between latent causal variables to be modulated by an additionally observed variable $\mathbf{u}$, represented by the red edge in Figure \ref{fig:problem3}. 
Intuitively, variant causal influences between latent causal variables cannot be `absorbed' by an invariant nonlinear mapping from $\mathbf{z}$ to $\mathbf{x}$, breaking the transitivity, resulting in identifiable causal representations. Specifically, we explore latent causal generative models where the observed data $\mathbf{x}$ is generated by the latent causal variables $\mathbf{z}$, allowing for any potential graph structures among $\mathbf{z}$. 
In addition, there exist latent noise variables $\mathbf{n}$, known as exogenous variables in causal systems, corresponding to latent causal variables. We introduce a surrogate variable $\mathbf{u}$ characterizing the changes in the distribution of $\mathbf{n}$, as well as the causal influences among latent causal variables $\mathbf{z}$. Here $\mathbf{u}$ could be environment, domain, or time index. More specifically, we parameterize the latent causal generative models as follows:
\begin{align}
&n_i: \sim {\cal N}( {\beta_{i,1}(\mathbf{u})}, {\beta_{i,2}(\mathbf{u})}),\label{eq:Generative1}\\
&{z_i} := {\boldsymbol{\lambda}^{T}_{i}(\mathbf{u})( \mathbf{z})}+n_i,  \label{eq:Generative2} \\
&\mathbf{x} :=\mathbf{f}(\mathbf{z},  \boldsymbol{\varepsilon}).
 \label{eq:Generative}
\end{align}
where 
\begin{itemize}[leftmargin=*]
    \item each noise term $n_i$ is Gaussian distributed with mean ${\beta_{i,1}(\mathbf{u})}$ and variance ${\beta_{i,2}(\mathbf{u})}$, $\mathbf{n} \in \mathbb{R}^{n}$; both ${\beta_{i,1}}$ and ${\beta_{i,2}}$ can be nonlinear mappings. Moreover, the distribution of $n_i$ is modulated by the observed variable $\mathbf{u}$. 
    \item In Eq. \eqref{eq:Generative2}, $\boldsymbol{\lambda_i}(\mathbf{u})$ denote the vector corresponding to the causal weights from $\mathbf{z}$, $\mathbf{z} \in \mathbb{R}^{n}$, to $z_i$. Specifically, $\boldsymbol{\lambda_i}(\mathbf{u}) = [\lambda_{1,i}(\mathbf{u}),...,\lambda_{n,i}(\mathbf{u})]$, where each ${\lambda_{j,i}}$ could be a nonlinear mapping. 
    \item In Eq. \eqref{eq:Generative}, $\mathbf{f}$ denote a nonlinear mapping from $\mathbf{z}$ to $\mathbf{x}$, $\mathbf{x} \in \mathbb{R}^{d}$ and $\boldsymbol{\varepsilon}$ is independent noise with probability density function $p_{\boldsymbol{\varepsilon}}(\boldsymbol{\varepsilon})$, $\boldsymbol{\varepsilon} \in \mathbb{R}^{d-n}$.
\end{itemize}
In Eq. \eqref{eq:Generative1}, we assume the latent noise variables to be Gaussian, thus the joint distribution can be expressed by the following exponential family: 
\begin{equation}
p_{\mathbf{(\mathbf{T}_\mathbf{n},\boldsymbol{ \beta})}}({\mathbf{n}|\mathbf u}) =
  \frac{1}{Z_\mathbf{n}(\boldsymbol{\beta}, \mathbf{u})}\exp{{\Big(\mathbf{T_n}^{T}(\mathbf{n})\boldsymbol{\eta}_\mathbf{n}(\mathbf u)\Big)}},
 \label{nef}
\end{equation}
where
$Z_\mathbf{n}(\boldsymbol{ \beta},\mathbf{u})$ denotes the normalizing constant, and $\mathbf{T}_\mathbf{n}(\mathbf{n})$ denotes the sufficient statistic for $\mathbf{n}$, whose the natural parameter $\boldsymbol{\eta}_\mathbf{n}(\mathbf u)$ depends on $\beta_{i,1}$ and $\beta_{i,2}$. 

In Eq. \eqref{eq:Generative2}, we assume the latent causal model of each $z_i$ satisfies a linear causal model with the causal weights being modulated by $\mathbf{u}$, and we call it  \textit{weight-variant linear Gaussian models}. Therefore, $p({\mathbf{z}|\mathbf u})$ satisfies the following multivariate Gaussian distribution: 
\begin{equation}
  p_{\mathbf{(\boldsymbol{\lambda},\boldsymbol{ \beta})}}({\mathbf{z}|\mathbf u}) =
  {\cal N}( \boldsymbol{\mu}, \Sigma),
 \label{mgd}
\end{equation}
with the mean $\boldsymbol{\mu}$ and the covariance matrix $\Sigma$ computed by the following recursion relations \citep{bishop2006pattern,koller2009probabilistic}:
\begin{equation}
\begin{aligned}
&{\mu}_i = \sum_{j\in \mathrm{pa}_{i}}\lambda_{j,i}(\mathbf{u}){\mu}_j + {\beta_{i,1}(\mathbf{u})}, \\
&\Sigma_{i,i}=\sum_{j\in \mathrm{pa}_{i}}\lambda^2_{j,i}(\mathbf{u}) \Sigma_{j,j} + {\beta_{i,2}(\mathbf{u})},\\
&\Sigma_{i,j}=\sum_{k\in \mathrm{pa}_{j}}\lambda_{k,j}(\mathbf{u})\Sigma_{i,k}, \quad for \quad i \neq j,\label{lineargaussian}
\end{aligned}
\end{equation}
where $\mathrm{pa}_{i}$ denotes the parent nodes of $z_i$. 
Furthermore, this multi-variate Gaussian can be re-formulated with the following exponential family:
\begin{equation}  p_{\mathbf{(\mathbf{T}_\mathbf{z},\boldsymbol{\lambda},\boldsymbol{ \beta})}}({\mathbf{z}|\mathbf u}) =
  \frac{1}{Z_\mathbf{z}(\boldsymbol{ \lambda},\boldsymbol{ \beta}, \mathbf{u})}\exp{{\Big(\mathbf{T}^{T}_\mathbf{z}(\mathbf{z})\boldsymbol{\eta}_\mathbf{z}(\mathbf u)\Big)}},
 \label{mgef}
\end{equation}
where the
parameter $\boldsymbol{\eta}_\mathbf{z}(\mathbf u)=[{\Sigma}^{-1}\boldsymbol{\mu};-\frac{1}{2}\vect({{\Sigma}^{-1}})]$,
$Z_\mathbf{z}(\boldsymbol{ \lambda},\boldsymbol{ \beta},\mathbf{u})$ denotes the normalizing constant, and the sufficient statistic $\mathbf{T}_\mathbf{z}(\mathbf{z})=[\mathbf{z};\vect(\mathbf{z}\mathbf{z}^{T})]$ (`vec' denotes the vectorization of a matrix). We further denote by $\mathbf{T}_{\text{min}}$ the minimal sufficient statistic and by $k$ its dimension, with $2n \leq k \leq n + (n(n+1))/2$. In particular, $k = 2n$ corresponds to the case when $\mathbf{T}_{\text{min}}(\mathbf{z})=[{z_1},z_2,...,z_n,{z^2_1},{z^2_2},...,{z^2_n}]$, or in other words,  there is no edges among the latent variables $\mathbf{z}$, while $k = n + (n(n+1))/2$ corresponds to
a full-connected causal graph over $\mathbf{z}$. 
So with different causal structures over $\mathbf{z}$, the dimension $k$ may vary. For the graph structure in Figure \ref{fig:problem}, $\mathbf{T}_{\text{min}}(\mathbf{z})=[{z_1},z_2,z_3,z_4,{z^2_1},{z^2_2},{z^2_3},{z^2_4},z_1z_2, z_1z_3,z_3z_4]$ and $k=11$.


Given the above, the following theorem shows that under certain assumptions on the nonlinear mapping $\mathbf{f}$ and the variability of $\mathbf{u}$, the latent variables $\mathbf{z}$ can be identifiable up to trivial permutation and scaling.
\begin{theorem} Suppose latent causal variables $\mathbf{z}$ and the observed variable $\mathbf{x}$ follow the generative models defined in Eq. \eqref{eq:Generative1}- Eq. \eqref{eq:Generative},
with parameters $({\mathbf{f},\boldsymbol{\lambda},\boldsymbol{\beta}})$. Assume the following holds:
\begin{itemize}
\item [\namedlabel{itm:eps} {(i)}] {$\boldsymbol{\varepsilon}$ is independent noise with probability density function $p_{\boldsymbol{\varepsilon}}(\boldsymbol{\varepsilon})$, which is always finite.}
\item [\namedlabel{itm:bijective}{(ii)}] The function $\mathbf{f}$ in Eq. \eqref{eq:Generative} is invertible and smooth. 
\item [\namedlabel{itm:nu} {(iii)}] There exist $2n+1$ distinct points $\mathbf{u}_{\mathbf{n},0},\mathbf{u}_{\mathbf{n},1},...,\mathbf{u}_{\mathbf{n},2n}$ such
that the matrix
\begin{equation}
    \mathbf{L}_\mathbf{n} = (\boldsymbol{\eta}_\mathbf{n}(\mathbf{u}_{\mathbf{n},1})-\boldsymbol{\eta}_\mathbf{n}(\mathbf{u}_{\mathbf{n},0}),..., \boldsymbol{\eta}_\mathbf{n}(\mathbf{u}_{\mathbf{n},2n})-\boldsymbol{\eta}_\mathbf{n}(\mathbf{u}_{\mathbf{n},0}))
\end{equation}
of size $2n \times 2n$ is invertible.\label{assu3}

\item [\namedlabel{itm:nz} {(iv)}] There exist $k+1$ distinct points $\mathbf{u}_{\mathbf{z},0},\mathbf{u}_{\mathbf{z},1},...,\mathbf{u}_{\mathbf{z},k}$ such
that the matrix
\begin{equation}
    \mathbf{L}_\mathbf{z} = (\boldsymbol{\eta}_\mathbf{z}(\mathbf{u}_{\mathbf{z},1})-\boldsymbol{\eta}_\mathbf{z}(\mathbf{u}_{\mathbf{z},0}),..., \boldsymbol{\eta}_\mathbf{z}(\mathbf{u}_{\mathbf{z},k})-\boldsymbol{\eta}_\mathbf{z}(\mathbf{u}_{\mathbf{z},0}))
\label{assumeiv}
\end{equation}
of size $k \times k$ is invertible.\label{assu4}
\item [\namedlabel{itm:lambda} {(v)}] The function class of $\lambda_{i,j}$ can be expressed by a Taylor
series: for each $\lambda_{i,j}$, $\lambda_{i,j}(\mathbf{u} = \mathbf{0})=0$,
\end{itemize}
{then the true latent causal variables $\mathbf{z}$, which are learned by matching the true marginal data distribution $p(\mathbf{x}|\mathbf{u})$, are related to the estimated latent causal variables ${ \mathbf{\hat z}}$ by the following relationship: $\mathbf{z} = \mathbf{P} \mathbf{\hat z} + \mathbf{c},$ where $\mathbf{P}$ denotes the permutation matrix with scaling, $\mathbf{c}$ denotes a constant vector.}
\label{theory1}
\end{theorem}


\paragraph{Proof sketch} The proof can be done according to the following intuition. With the support of assumptions \ref{itm:eps}-\ref{itm:nu}, one can utilize the identifiability result from nonlinear ICA to identify the latent noise variables $\mathbf{n}$ up to permutation and scaling, \ie, $\mathbf{n}=\mathbf{P}\mathbf{\hat n} + \mathbf{c}_\mathbf{n}$ where $\mathbf{\hat n}$ denotes the recovered latent noise variables obtained by matching the true marginal data distribution, and $\mathbf{c}_\mathbf{n}$ notes a constant vector. Next, using a similar technique, we can utilize assumption \ref{itm:nz} to establish a linear transformation between the true latent causal variables $\mathbf{z}$ and the recovered latent causal variables $\mathbf{\hat z}$, \ie, $\mathbf{z} = \mathbf{A}(\mathbf{\hat z}) + \mathbf{c}_\mathbf{z}$ where $\mathbf{c}_\mathbf{z}$ denotes a constant vector, and $\mathbf{A}$ denotes a liner transformation. Finally, combining the identifiability result for the latent noise variables $\mathbf{n}$ with assumption \ref{itm:lambda}, the linear transformation can be
reduced to permutation and scaling, \ie, $\mathbf{z} = \mathbf{P}(\mathbf{\hat z}) + \mathbf{c}_\mathbf{z}$. See Appendix \ref{prooflinear} for details.

\paragraph{Assumptions \ref{itm:eps}-\ref{itm:nz}} Assumptions \ref{itm:eps}-\ref{itm:nu} are originally developed by nonlinear ICA \citep{hyvarinen2016unsupervised, hyvarinen2019nonlinear,khemakhem2020variational,sorrenson2020disentanglement}. We here consider
unitizes these assumptions considering the following two main reasons. 1) These assumptions have been verified to be practicable in diverse real-world application scenarios \cite{kong2022partial,xie2022multi}. 2) Our result
eliminates the need to know the exact dimensionality of latent causal or noise variables, which is in contrast to existing methods that require prior knowledge of the dimensionality, due to imposing the two-parameter Gaussian noise variables \citep{sorrenson2020disentanglement}. Assumptions \ref{itm:nz}-\ref{itm:lambda} originally introduced by this work. Intuitively, assumption \ref{itm:nz} is similar to assumption \ref{itm:nu} in the hope that the auxiliary variable $\mathbf{u}$ must have a sufficiently strong and diverse effect on the distributions of latent causal variables $\mathbf{z}$, similar to latent noise variables $\mathbf{n}$. 

\paragraph{Assumptions \ref{itm:lambda} and sufficient change} \label{para: sufficient} Among the assumptions above, the only one that cannot be considered weak or natural is assumption \ref{itm:lambda}. Essentially, it is to ensure \textit{sufficient changes} in causal influence. For simplicity, consider two variables, $z_1$ and $z_2$, and $z_2:=\lambda(\mathbf{u})z_1+n_2$, where $\lambda(\mathbf{u})  = \hat \lambda(\mathbf{u}) + b$, with $\lambda$ and $\hat \lambda$ belonging to the same function class. As a consequence, while the causal influence $\lambda(\mathbf{u})$ changes as a whole across $\mathbf{u}$, there always exists a part $bz_1$ that remains unchanged across $\mathbf{u}$. This unchanged term $bz_1$ across $\mathbf{u}$ can be absorbed into $\mathbf{f}$ the mapping from $\mathbf{z}$ to $\mathbf{x}$, resulting in a possible solution $\hat z_2 := \hat \lambda(\mathbf{u})z_1 + n_2$, instead of the groundtruth $(\hat \lambda(\mathbf{u}) + b) z_1 + n_2$, which leads to an unidentifiable outcome. Essentially, this unidentifiable result occurs because a part of a weight remains unchanged across $\mathbf{u}$, a situation we refer to as \textit{insufficient change}. The objective of assumption \ref{itm:lambda} is to prevent such insufficient changes. Specifically, by constraining the function class of $\lambda$ such that  $\lambda(\mathbf{u}=\mathbf{0})=0$ and thus $\hat \lambda(\mathbf{u}=\mathbf{0})=0$, $\lambda(\mathbf{u})$ can not be expressed by $\hat \lambda(\mathbf{u}) + b$ with $b\neq 0$. Intuitively, assumption \ref{itm:lambda} suggests that there is a scenario where an existing edge can be removed at a certain $\mathbf{u}$, \ie, $\lambda(\mathbf{u}=\mathbf{0}) =0$. By removing the edge, we can ensure that no part of the weight corresponding to the edge remains unchanged. It is important to emphasize that assumption \ref{itm:lambda} is to limit the function class, so that $\lambda$ do not include a constant term, such as $b$, by assuming $\lambda(\mathbf{u}=\mathbf{0})=0$. Once samples are drawn from this limited function class, the assumption is satisfied, making the observed data generated by these samples effective for inferring latent causal variables. Therefore, we do not need the samples to necessarily include the specific point $\mathbf{u}=\mathbf{0}$, to generate the corresponding observed data, which is then used for inferring the latent causal variables.

\paragraph{Assumption \ref{itm:lambda} as a Bridge Between Soft and Hard Interventions}
Assumption~\ref{itm:lambda} implies that, although a causal edge may exist between two latent variables, there can be a reference condition under which this influence
effectively vanishes, i.e., $\lambda(\mathbf{u}=\mathbf{0})=0$. This bears a close
connection to hard interventions, which remove causal dependencies by construction. Unlike existing approaches that explicitly rely on hard interventions
\citep{ahuja2023interventional,seigal2022linear,buchholz2023learning,varici2023score}, our formulation characterizes a restricted function class for $\lambda(\mathbf{u})$
that enables identifiability without requiring direct access to such interventions. We view this as a conceptual link between soft and hard interventions, while a systematic unification is beyond the scope of this work.

\paragraph{Potential Application}
Assumption~\ref{itm:lambda} can arise naturally in biological imaging settings, such as cell imaging experiments where samples are exposed to different small-molecule compounds. In this context, latent variables may represent protein groups, and causal edges correspond to protein--protein interactions (PPIs). Certain compounds are known to disrupt or inhibit specific PPIs, effectively suppressing the corresponding causal influences \citep{arkin2004small,lu2020recent}. Such scenarios provide a plausible instance where $\lambda_{i,j}(\mathbf{u}=\mathbf{0})=0$ approximately holds, making the
assumption empirically relevant.

\paragraph{Insights} Intuitively, allowing causal influences among latent causal variables to change leads to varying causal influences. Such variations cannot be 'absorbed' by an invariant nonlinear mapping from $\mathbf{z}$ to $\mathbf{x}$, which breaks the transitivity mentioned in Section \ref{sec:Transitivity}, resulting in identifiability. For example, assume that $z_1:=n_1$, $z_2:= \lambda(\mathbf{u})z_1+n_2$ and $\mathbf{x}:= \mathbf{f}(z_1,z_2)$. If we allow the causal influence $\lambda(\mathbf{u})z_1$ to be 'absorbed' by the mapping $\mathbf{f}$, we would have $z_1:=n_1$, $z_2:= n_2$, and the resulting mapping, $\mathbf{f'} = \mathbf{f} \circ \mathbf{g}$ where $\mathbf{g} (z_1,z_2)=[z_1,\lambda(\mathbf{u})z_1+z_2]$, would need to include the original $\mathbf{f}$ as well as the term $\lambda(\mathbf{u})$ in $\mathbf{g}$, to obtain the same obervation $\mathbf{x}$. This means that the resulting mapping $\mathbf{f'}$ would change across $\mathbf{u}$ due to the term $\lambda(\mathbf{u})$, which violates the fact that the mapping from $\mathbf{z}$ to $\mathbf{x}$ remains unchanged across $\mathbf{u}$ as defined in Eq.~\eqref{eq:Generative}.



\paragraph{Identifiability of Causal Structures among Latent Variables} Theorem \ref{theory1} has shown that latent causal variables can be identified up to trivial permutation and linear scaling. With this result, the identifiability of causal structure in the \textit{latent space} reduces to the identifiability of the causal structure in the \textit{observational space}. Denote by $G_{\mathbf{z}}^u$ the causal graph over latent variables $\mathbf{z}$ when $\mathbf{u}=u$, and let $G_{\mathbf{z}}^{\text{union}} = \cup_u G_{\mathbf{z}}^u$ be the union of causal graphs across different values of $\mathbf{u}$. Similarly, we define $G_{\mathbf{z} \cup \mathbf{u}}^{\text{union}}$ to be the union of causal graphs over $\mathbf{z} \cup \mathbf{u}$.  With the help of recent progress in causal discovery from heterogeneous data \citep{CDNOD_20}, the following corollary shows that the causal structure among latent variables $\mathbf{z}$ is identifiable up to the Markov equivalence class of $G_{\mathbf{z}}^{\text{union}}$. Moreover, although there exists scaling indeterminacy for the recovered latent variables as stated in Theorem \ref{theory1}, it does not affect the identifiability of the causal structure.


\begin{corollary}
  Suppose latent causal variables $\mathbf{z}$ and the observation $\mathbf{x}$ follow the generative models defined in Eq. \eqref{eq:Generative1}- Eq. \eqref{eq:Generative}, and that the conditions in Theorem \ref{theory1} hold. Then, under the assumption that $G_{\mathbf{z}}^{\text{union}}$ is a DAG and that the joint distribution over $\mathbf{z} \cup \mathbf{u}$ is Markov and faithful to $G_{\mathbf{z} \cup \mathbf{u}}^{\text{union}}$, the causal structure among variables $\mathbf{z}$ can be identified up to the Markov equivalence class of $G_{\mathbf{z}}^{\text{union}}$. That is, $\textit{MEC}(G_{\mathbf{z}}^{\text{union}})$ is identifiable, where $\textit{MEC}(\cdot)$ denotes the Markov equivalence class.
  \label{corollary}
\end{corollary}

\paragraph{Proof Sketch} The proof can be done by first establishing that Theorem \ref{theory1} identifies the latent causal variables up to trivial permutation and linear scaling, which reduces the problem to identifying the causal structure in the observed space. Then, by leveraging the results from \citet{CDNOD_20}, which ensure identifiability of the causal structure in the observed space under the Markov condition and faithfulness assumption, we can identify the causal structure up to the Markov equivalence class. Next, we show that the Markov equivalence class over $\mathbf{z}$ remains unchanged after removing the domain variable $\mathbf{u}$ and its edges, preserving the graph's skeleton and directions. Finally, since linear scaling does not affect conditional independence relationships, the identifiability of the causal structure in the latent space is preserved. As a result, the causal structure among the latent variables $\mathbf{z}$ can be identified up to the Markov equivalence class. Please refer to Appendix \ref{corollary:sketch} for more details.

In addition, with the help of the independent causal mechanisms (ICM) principle \citep{ghassami2018multi, CDNOD_20, scholkopf2021toward}, the following corollary shows that the causal structure among latent variables $\mathbf{z}$ is fully identifiable.

\begin{corollary}
  Suppose the conditions in Corollary \ref{corollary} hold. Denote by $\boldsymbol{\theta}_i$ the involved parameters in the causal mechanism of $z_i$, with $\boldsymbol{\theta}_i = (\beta_{i,1}; \beta_{i,2}; \boldsymbol{\lambda}_i)$, and denote by $z_{pa(i)}$ the parents of $z_i$ in $G_{\mathbf{z}}^u$. If $\boldsymbol{\theta}_i$ and $\boldsymbol{\theta}_{pa(i)}$ change independently across different values of $\mathbf{u}$  for any $z_i$, then the acyclic causal structure among latent variables $\mathbf{z}$ can be fully identified. That is, $G_{\mathbf{z}}^{\text{union}}$ is fully identifiable.
  
  \label{corollary1}
\end{corollary}
\paragraph{Proof Sketch} To prove this corollary, we first introduce the ICM principle, which states that in a causally sufficient system, the causal modules and their parameters change independently across different domains. This principle holds in the causal direction, meaning that changes in one module do not affect others. However, this independence generally breaks down in the anti-causal direction. Clearly, The scaling indeterminacy for the recovered latent variables does not affect this principle. Please refer to Appendix \ref{corollary:dag} for more details.

\section{Change of Part of Weights and Partial Identifiability Result}
\label{sec:partial}
The aforementioned theoretical result necessitates that all weights undergo changes across $\mathbf{u}$, as constrained by the assumption \ref{itm:lambda} in Theorem \ref{theory1}. However, in practical applications, this assumption may not hold true. Consequently, two fundamental questions naturally arise: Is this assumption necessary for identifiability in the absence of any supplementary assumptions? Alternatively, can we obtain partial identifiability results if only some of the weights change across $\mathbf{u}$? In fact, when part of weights change, we can still provide partial identifiability results, as outlined below.

\begin{theorem} Suppose latent causal variables $\mathbf{z}$ and the observed variable $\mathbf{x}$ follow the generative models defined in Eq.~\eqref{eq:Generative1}-Eq.~\eqref{eq:Generative}. Under the condition that the assumptions \ref{itm:eps}-\ref{itm:nz} in
Theorem \ref{theory1} are satisfied, for each $z_i$,
\begin{itemize}
     \item [\namedlabel{corollary:a} {(a)}] if $z_i$ is a root node or all weights to $z_i$, $\boldsymbol{\lambda}_i$ in Eq. \eqref{eq:Generative2}, meet assumption \ref{itm:lambda} in Theorem \ref{theory1}, then the true $z_i$ is related to the recovered one $\hat z_j$, obtained by matching the true marginal data distribution $p(\mathbf{x}|\mathbf{u})$, by the following relationship: $z_i=s\hat z_j + c $, where $s$ denotes scaling, $c$ denotes a constant,
     \item [\namedlabel{corollary:b} {(b)}] if there exists an unchanged weight ${\lambda}_{j,i}$ from the parent node $z_j$ to $z_i$, in Eq. \eqref{eq:Generative2}, then $z_i$ is unidentifiable,
     \item [\namedlabel{corollary:c} {(c)}] if there exists an weight that can be expressed as $ \lambda_{j,i}(\mathbf{u}) + b$, where b is a non-zero constant, from the parent node $z_j$ to $z_i$, then $z_i$ is unidentifiable.
 \end{itemize}
\label{theory2}
\end{theorem}
\paragraph{Proof sketch} This can be proved by the fact that regardless of the assumption \ref{itm:lambda} in Theorem \ref{theory1} , two results hold in Theorem \ref{theory1}, i.e., $\mathbf{z} = \mathbf{A}(\mathbf{\hat z}) + \mathbf{c}_\mathbf{z}$, and $\mathbf{n} = \mathbf{P} \mathbf{\hat n} + \mathbf{c}_\mathbf{n}$. Then using the change of all weights in $\boldsymbol{\lambda}_i$, we can prove \ref{corollary:a}. To prove \ref{corollary:b}, we demonstrate that it is always possible to construct an alternative solution $\hat z_i$ by removing the term ${\lambda}_{j,i}z_{j}$, corresponding to unchanged weight, in the true ${z}_i$, which is capable of generating the same observations $\mathbf{x}$. As a result, we can show non-identifiable result. To prove \ref{corollary:c}, similar to \ref{corollary:b}, we can construct an alternative solution $\hat z_i$ by removing the term $bz_{j}$, corresponding to the constant $b$, in true $z_i$. 

\paragraph{Insights} 1) The result (b) of Theorem \ref{theory2} implies the necessity of changing weights, without additional assumptions. This can be proven by the contrapositive of (b), i.e., if $z_i$ is identifiable, then there do not exist unchanged weights from the parent nodes to $z_i$. 2) The result (c) of Theorem \ref{theory2} implies the necessity of eliminating the corner case where $\lambda(\mathbf{u})  = \lambda'(\mathbf{u}) + b$, without additional assumptions. Similarly, this can be proven by the contrapositive of (c), i.e., if $z_i$ is identifiable, then there do not exist a weight that can be expressed as $\lambda(\mathbf{u}) + b$. Essentially, this implies that sufficient changes in weights are required for identifiability. Informally, if $z_i = \boldsymbol{\lambda}_i^{T}\mathbf{z} + n_i$, then sufficient changes mean that $\boldsymbol{\lambda}_i^{T}$ do not include non-zero constant across $\mathbf{u}$. 3) Together with result (a) of Theorem \ref{theory2}, Theorem \ref{theory2} implies that the entire latent space can theoretically be partitioned into two distinct subspaces: one subspace pertains to invariant latent variables, while the other encompasses variant variables. This partitioning may be valuable for applications that focus on learning invariant latent variables to adapt to varying environments, such as domain adaptation or generalization.

\paragraph{Discussion 1: Partial Causal Structure} Partial identifiability of latent causal variables in Theorem \ref{theory2} does not necessarily guarantee the unique recovery of the corresponding partial latent causal graph structure. However, a probable result can still be achieved. Specifically, if there are no interactions (edges) between the two latent subspaces in the ground truth graph structure, it becomes possible to recover the latent causal structure within the latent variant space. When interactions do exist, examining how they affect the recovery of the latent causal graph structure is an intriguing area for further exploration. Additionally, it is valuable to investigate how such partial results influence the outcomes of interventions and counterfactual inference.

\paragraph{Discussion 2: Parent nodes do not impact children} Perhaps counterintuitively, the partial identifiability result in Theorem \ref{theory2} suggests that the identifiability of $z_i$ remains possible even when its parent nodes are unidentifiable. This is primarily due to the identifiability of all latent noise variables $\mathbf{n}$, regardless of changes in the weights. In this context, for latent linear causal models, all necessary information to recover latent causal variables is encapsulated within these identifiable latent noise variables. As a result, it is possible to recover $z_i$, irrespective of the identifiability of its parent nodes.

\paragraph{Discussion 3: Potential Implications} While partial identifiability may be more realistic in practice, since only some causal influences vary across environments, it still poses challenges in determining whether the latent causal variables have been successfully identified or are practically useful, especially in the absence of ground-truth latent variables. Nevertheless, such partial identifiability results may be valuable in downstream applications, particularly in domain shift scenarios such as domain adaptation, domain generalization, and transfer learning \citep{kong2022partial,li2023subspace,liu2025latent}. In these settings, partially identified latent causal variables that exhibit stable causal relationships with the target label across environments may capture invariant mechanisms useful for robust prediction. Their relevance can be empirically assessed through predictive performance, stability under distribution shifts.

\section{Learning Causal Representations with Weight-Variant Linear Models}
\label{sec:suave}
Based on the identifiable results above, in this section, we propose a structural causal variational autoencoder (\FancyName) to learn latent causal representations. We first propose a structural causal model prior relating to the proposed weight-variant linear Gaussian model. We then show how to incorporate the proposed prior into the traditional VAE framework \citep{VAE_Welling13}, together with a variational posterior. 

\paragraph{DAG constraints}
To address the complexity of combinatorial search, recent methods have proposed exact characterizations of DAGs that enable tackling the problem through continuous optimization techniques \citep{zheng2018dags,yu2019dag,zheng2020learning,he2021daring}. However, this approach does not guarantee the absence of cycles at any stage of training, and solutions often require post-processing \citep{zantedeschi2023dag}. By contrast, as mentioned in Section \ref{sec:permutation}, the permutation indeterminacy in latent space allows us to predefine a causal order without specifying semantic information. With the guarantee of identifiability, the nodes in the predefined order are enforced to learn the corresponding semantic information in the true causal order for generating data, ensuring each node learns its designated role. Given this, we can incorporate a predefined causal order in the following prior and posterior models.

\paragraph{Prior Model}
According to the causal model for the latent causal variables, \ie, Eq. \eqref{eq:Generative1} and Eq. \eqref{eq:Generative2}, we construct a linear Gaussian prior with weights and noise depending on $\mathbf{u}$. Since we can pre-define the causal order as mentioned above, let us assume the causal order is $z_1,z_2,..z_n$. Given this, we can create a fully connected causal graph based on the predefined causal order, where each $z_i$ has parent nodes $z_i'$ for $i'<i$. Summarizing the above, we propose the following prior model:
\begin{equation}
p({\bf{z}}|\mathbf{u})=p(z_1|\mathbf{u})\prod \limits_{i=2}^{n}{p({{z_i}}|{\bf z}_{<i},\mathbf{u})}
    =\prod \limits_{i=1}^n{{\cal N}(\mu_{z_i},\delta^2_{z_i})},
    \label{prior}
\end{equation}
where $\mu_{z_i} = \sum\limits_{i' < i}\lambda_{i',i}(\mathbf{u})z_{i'}+\beta_{i,1}(\mathbf{u}),
\delta^2_{z_i} = \beta_{i,2}(\mathbf{u})$, and the corresponding weight matrix $\boldsymbol{\lambda} = [\boldsymbol{\lambda}_{1}(\mathbf{u}),...,\boldsymbol{\lambda}_{n}(\mathbf{u})]$ can be constrained as a upper triangular matrix with zero-value diagonal elements. Again, the proposed prior naturally ensures a directed acyclic graph estimation, avoiding traditional DAG constraints.

\paragraph{Variational Posterior and Objective}

The nature of the proposed prior in Eq. \eqref{prior} gives rise to the following variational posterior:
\begin{equation}
q({\bf{z}}|\mathbf{x,u})=q(z_1|\mathbf{x,u})\prod \limits_{i=2}^{n}{q({{z_i}}|{\bf z}_{<i},\mathbf{x,u})}
    =\prod \limits_{i=1}^n{{\cal N}(\mu'_{z_i},\delta^{'2}_{z_i})},
    \label{poster}
\end{equation}
where $\mu'_{z_i} =\sum\limits_{i' < i}\lambda^{'}_{i',i}(\mathbf{u})z_{i'} +\beta^{'}_{i,1}(\mathbf{x,u}),
\delta^{'2}_{z_i} = \beta^{'}_{i,2}(\mathbf{x,u}).$ Therefore, we can arrive at the following evidence lower bound (ELBO):
\begin{equation}
\label{obj}
\mathop { }\mathbb{E}_{q(\mathbf{z}|\mathbf{x,u})}(\log p(\mathbf {x}|\mathbf{z},\mathbf{u})) - {D_{KL}}({q }(\mathbf{z|x,u})||p(\mathbf{z}|\mathbf{u})),
\end{equation}
where ${D_\text{KL}}$ denotes Kullback–Leibler divergence, a measure of how one probability distribution ${q}(\mathbf{z|x,u})$ is different from a second, reference probability distribution $p(\mathbf{z}|\mathbf{u})$. We implement assumption (v) and incorporate it into the evidence lower bound by the following objective:
\begin{equation}
\label{obj1}
\mathop { }\mathbb{E}_{q(\mathbf{z}|\mathbf{x,u})}(\log p(\mathbf {x}|\mathbf{z},\mathbf{u})) - {D_\text{KL}}({q}(\mathbf{z|x,u})||p(\mathbf{z}|\mathbf{u})) - \gamma_1 \sum \limits_{i,j}  ||\lambda_{i,j}(\mathbf{u}=\mathbf{0})||_{1} - \gamma_2 \sum\limits_{i,j} ||\lambda'_{i,j}(\mathbf{u}=\mathbf{0})||_{1}.
\end{equation}

\begin{figure} 
\centering
\includegraphics[width=0.7\textwidth]{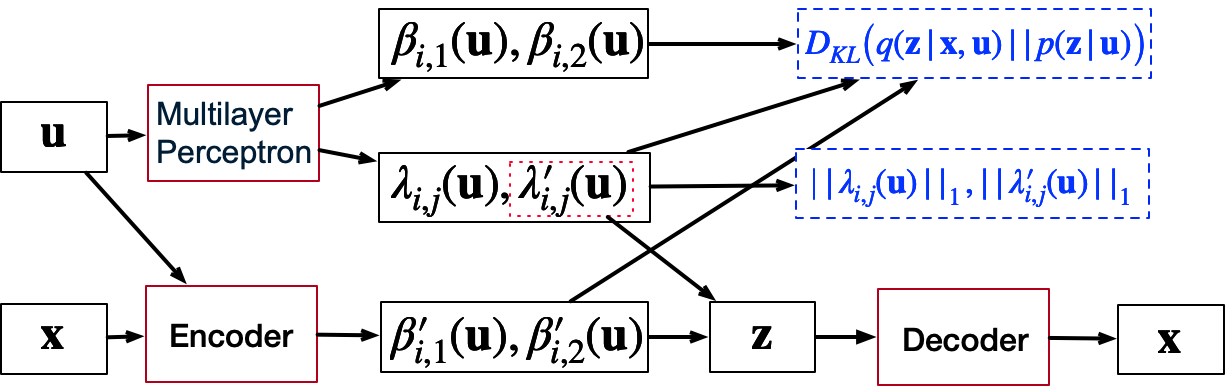}
  \caption{Implementation framework to learn linear Gaussian causal representations.}
  \label{fig:alg}
\end{figure}

In our implementation, the observed variable $\mathbf{u}$ is represented using one-hot encoding, so that in the experiments $\mathbf{u}$ takes values in a finite discrete set. In this discrete setting, Assumption~(v) can in principle be satisfied exactly by fixing
$\lambda_{i,j}(\mathbf{u}=0)=0$ by construction and learning $\lambda_{i,j}(\mathbf{u})$ only for $\mathbf{u}\neq 0$. In our implementation, we instead adopt a simple penalty that encourages $\lambda_{i,j}(\mathbf{u}=0)$ to be close to zero, which serves as a convenient implementation choice. This penalty is included in the last two terms of Eq.~\eqref{obj1}, where $\gamma_1$ and $\gamma_2$ are hyperparameters that balance the penalty with the ELBO.
Figure~\ref{fig:alg} illustrates the overall implementation of the proposed method.

\paragraph{Discussion: Relating to hierarchical VAE} Beyond standard VAEs, recent advanced works aim to explore hierarchical VAEs with the hope that they could improve the evidence lower bound in standard VAE and decrease reconstruction error compared to standard VAEs. More importantly, the stack of latent variables in hierarchical VAEs might learn a feature hierarchy, where different layers capture information at varying levels of detail and abstraction \citep{{kingma2016improved,sonderby2016ladder,maaloe2019biva,vahdat2020nvae}}. To this end, hierarchical VAE models often employ the following generative model: $z_1\rightarrow z_2\rightarrow...\rightarrow z_n \rightarrow \mathbf{x}$. Interestingly, hierarchical VAE models can be regarded as a special case of the proposed prior model in Eq. \eqref{prior}, where the fully-connected graph is reduced to the simpler connections used in the generative model of hierarchical VAEs. Essentially, the proposed \FancyName also aims to learn hierarchical latent variables. Due to permutation indeterminacy, we can model the prior or posterior as hierarchical structures. The difference is that the edges among latent variables of the proposed \FancyName are learned adaptively from data, unlike the simple connections in hierarchical VAEs. Importantly, the proposed \FancyName has rigorous theoretical justification for learning hierarchical latent variables and the structures among them, whereas hierarchical VAEs lack such support without further assumptions. Further exploring the relationship between hierarchical VAEs and latent causal models is both interesting and worthwhile.

\section{Experiments} 
\label{sec:exp}
\subsection{Synthetic Data} 
We first conduct experiments on synthetic data, generated by the following process: we divide the latent noise variables into $M$ segments, where each segment corresponds to one conditional variable $\mathbf{u}$ as the segment label. Within each segment, we first sample the mean ${\beta}_{i,1}$ and variance ${\beta}_{i,2}$ from uniform distributions $[-2,2]$ and $[0.01,3]$, respectively. Subsequently, we sample the weights ${\lambda}_{i,j}$ from a uniform distribution over $[0.1, 2]$. Then for each segment, we generate the latent causal samples according to the generative model in Eq. \eqref{eq:Generative2}. Finally, we obtain the observed data samples $\mathbf{x}$ by an invertible nonlinear mapping. More details can be found in Appendix.

\paragraph{Comparison} We compare the proposed method with identifiable VAE (iVAE) \citep{khemakhem2020variational}, $\beta$-VAE \citep{higgins2016beta}, CausalVAE \citep{yang2020causalvae}, and vanilla VAE \citep{kingma2013auto}. Among them, iVAE has been proven to be identifiable so that it is able to learn the true independent noise variables with certain assumptions. While $\beta$-VAE has no theoretical support, it has been widely used in various disentanglement tasks. Note that both methods assume that the latent variables are independent, and thus they cannot model the relationships among latent variables. To make a fair comparison in the unsupervised setting, we implement an unsupervised version of CausalVAE, which is not identifiable.
\begin{figure}[!htp]
  \centering
{\includegraphics[width=0.5\textwidth]{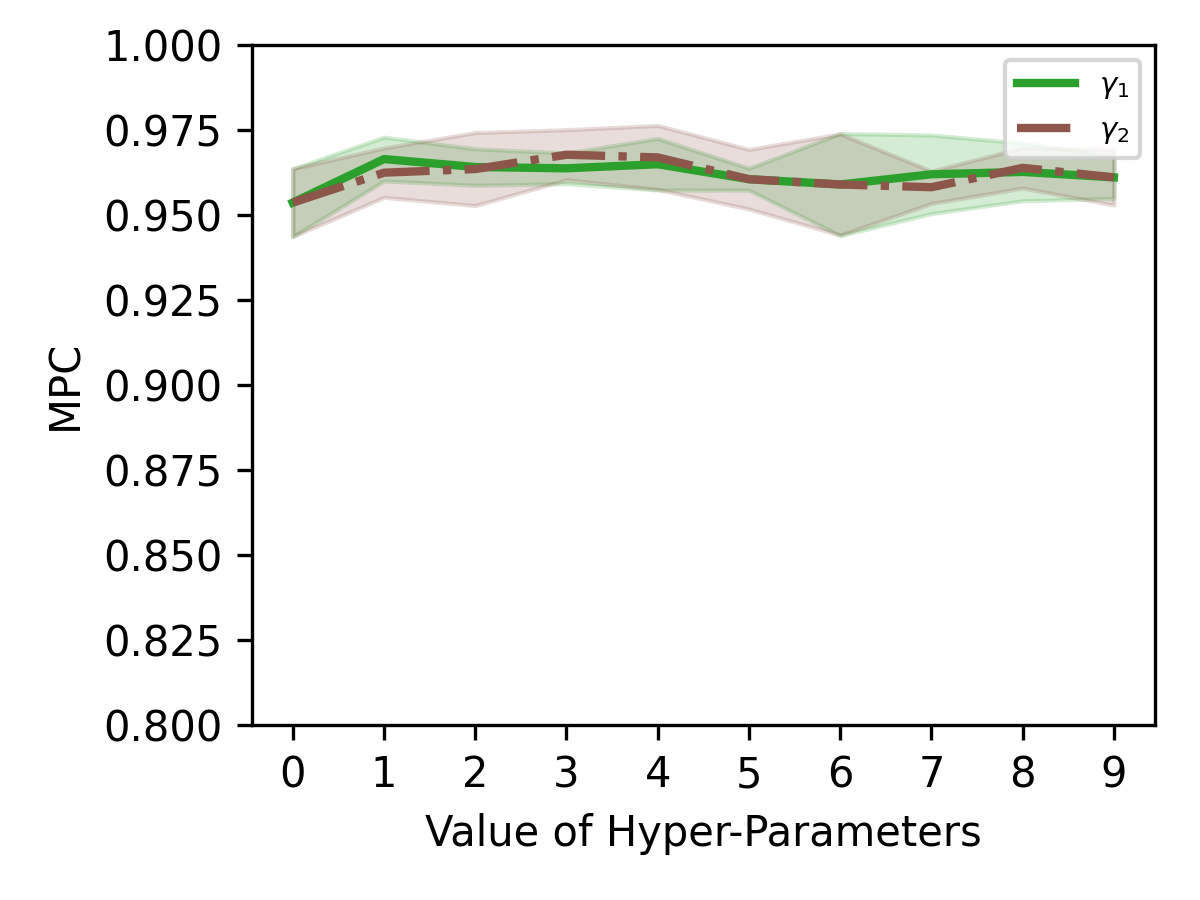}}\
\caption{Performances of the proposed \FancyName with different values of hyper-parameters $\gamma_1$ and $\gamma_2$. Implementation details can be found in Appendix \ref{expbc}. } 
  \label{fig:synthetic:gamma}
\end{figure}

\paragraph{Performance metric} Since the proposed method can recover the latent causal variables up to trivial permutation and linear scaling, we compute the mean of the Pearson correlation coefficient (MPC) to evaluate the performance of our proposed method. The Pearson correlation coefficient is a measure of linear correlation between the true latent causal variables and the recovered latent causal variables. Note that the Pearson coefficient is suitable for iVAE, since it has been shown that iVAE can also recover the latent noise variables up to linear scaling under the setting where the mean and variance of the latent noise variables are changed by $\mathbf{u}$ \citep{sorrenson2020disentanglement}. To remove the permutation effect, following \citet{khemakhem2020variational}, we first calculate all pairs of correlation and then solve a linear sum assignment problem to obtain the final results. A high correlation coefficient means that we successfully identified the true parameters and recovered the true variables, up to component-wise linear transformations. 

\paragraph{Ablation Study} 
We first conduct experiments on the proposed method, varying the hyper-parameters $\gamma_1$ and $\gamma_2$. Figure \ref{fig:synthetic:gamma}
shows the results of performance of the proposed \FancyName with different values of hyper-parameters. The performance remains relatively stable on synthetic data. This suggests that on randomly generated synthetic data, the variation in weights generally satisfies assumption (v), which requires avoiding insufficient change—specifically, the scenario where a portion of the weights remains unchanged across $\mathbf{u}$. This stability occurs because the weights are independently and randomly generated without any inherent patterns, making it unlikely that they would fit any specific nonlinear/linear function with a constant term. Consequently, the proposed method naturally avoids the special case where some weights remain unchanged. Therefore, in the subsequent experiments, we set the values of the hyper-parameters $\gamma_1$ and $\gamma_2$ to zero. 

\begin{figure}[h]
  \centering
{\includegraphics[width=0.5\textwidth]{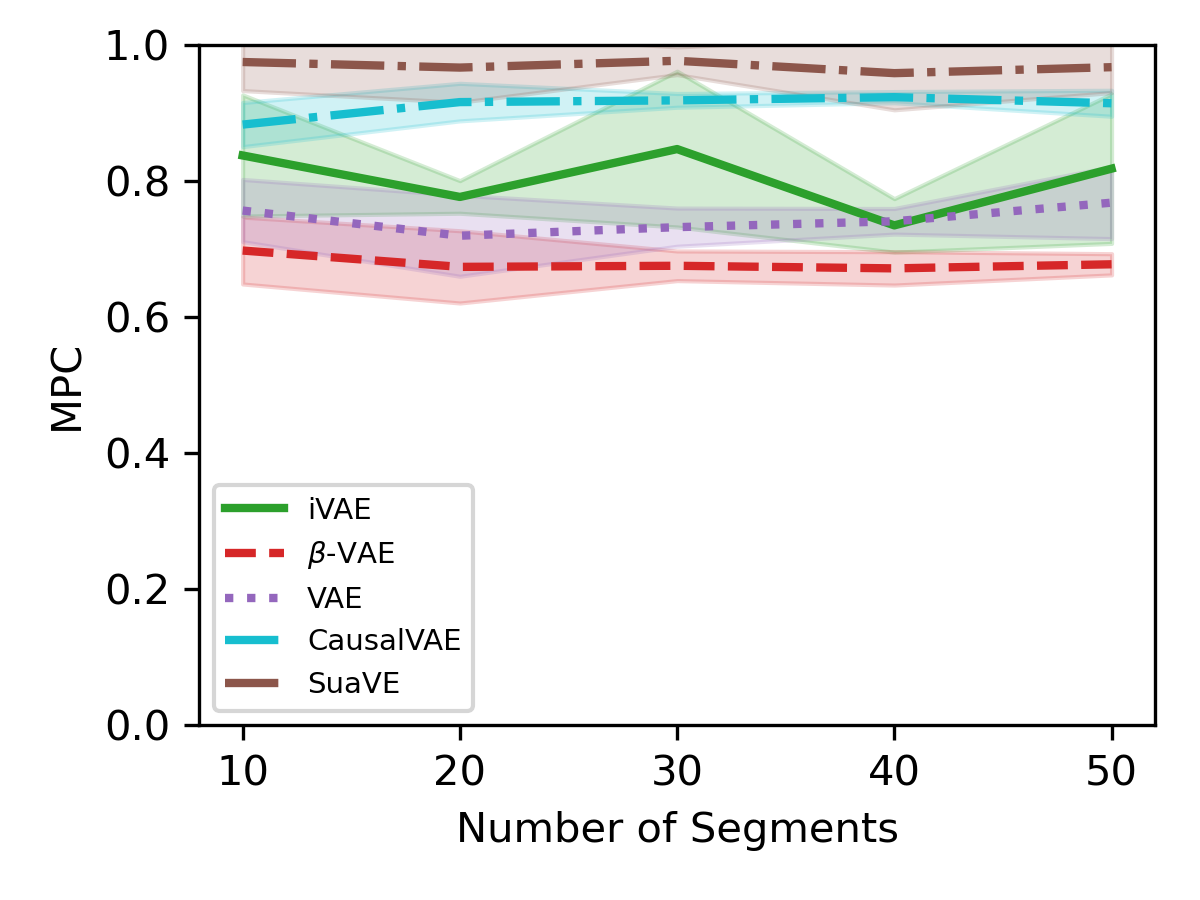}}\
\caption{Performances of the proposed \FancyName in comparison to iVAE, $\beta$-VAE, VAE and CausalVAE in recovering the latent causal variables on synthetic data with different numbers of segments. Implementation details can be found in Appendix \ref{appendix: details1}.} 
  \label{fig:synthetic}
\end{figure}

\begin{figure}[h]
  \centering
{\includegraphics[width=0.5\textwidth]{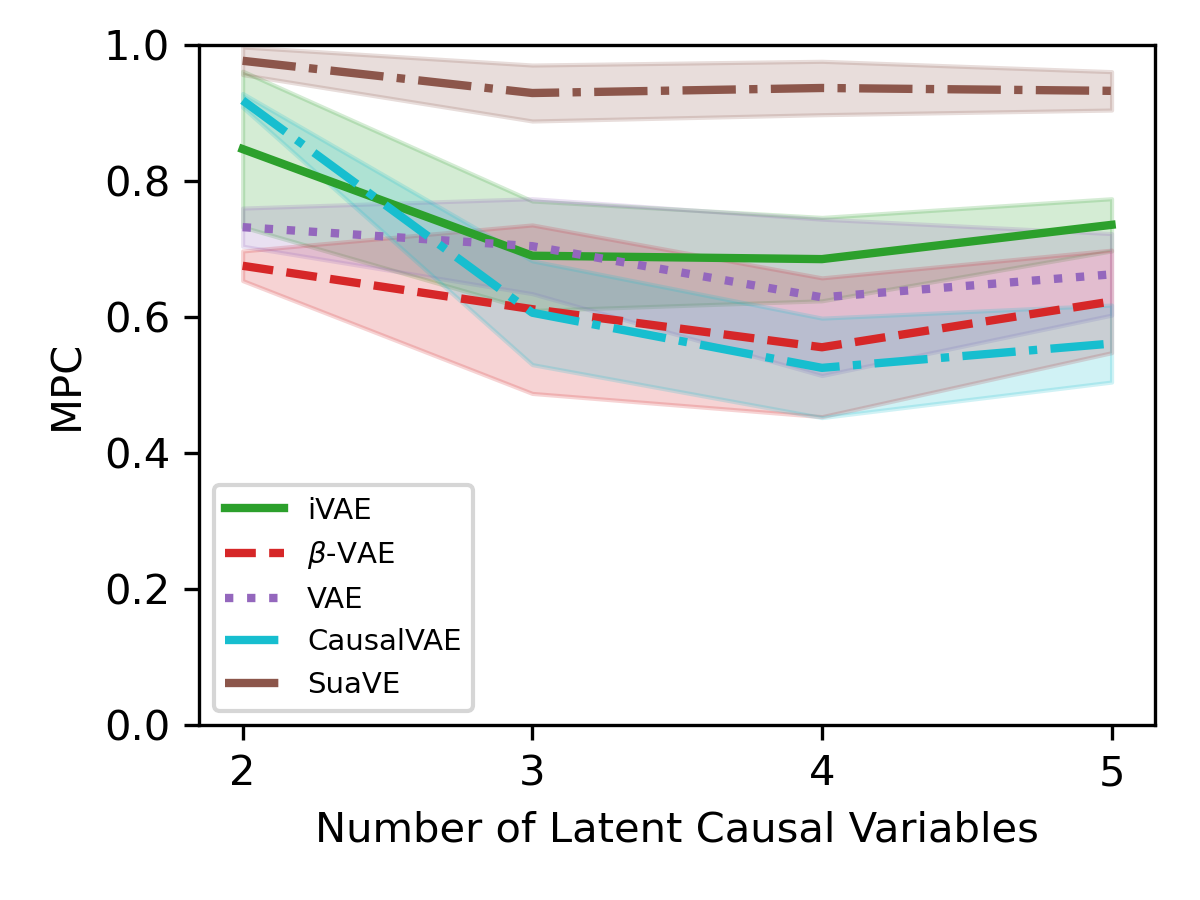}}
  \caption{Performances of the proposed \FancyName in comparison to iVAE, $\beta$-VAE, VAE and CausalVAE in recovering the latent causal variables on synthetic data with different numbers of the latent causal variables. Details can be found in Appendix \ref{appendix: details}.} 
  \label{fig:synthetic:b}
\end{figure}

\begin{figure}[h]
  \centering
{\includegraphics[width=0.5\textwidth]{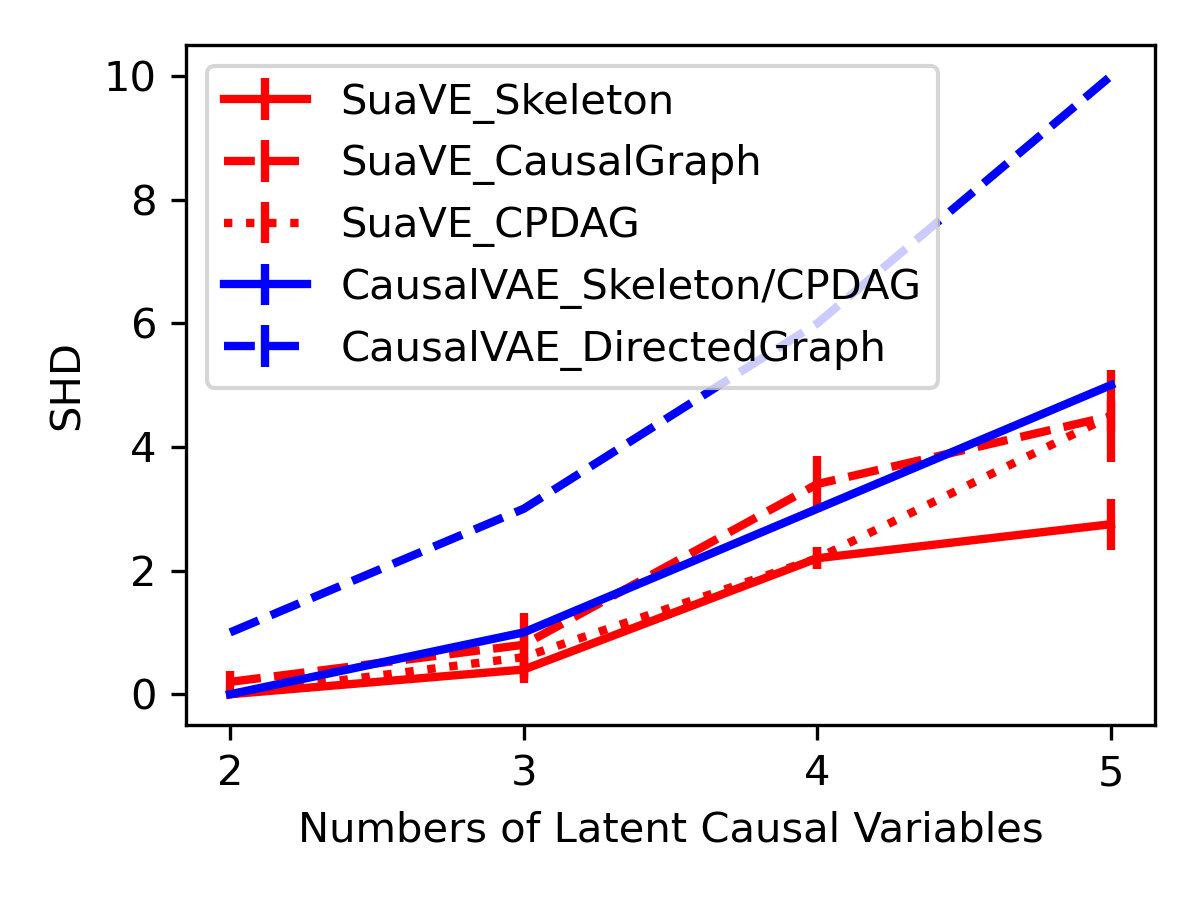}}
  \caption{The SHD obtained by the proposed \FancyName and CausalVAE. Since CausalVAE is not identifiable, it obtains fully connected graphs, the SHD of the recovered CPDAG is the same as one of the recovered skeletons. Implementation details can be found in Appendix \ref{appendix: details}.} 
  \label{fig:synthetic:c}
\end{figure}

\paragraph{Analysis of Comparison Results} We compared the performance of the proposed \FancyName to some variants of VAE mentioned above. We used the same network architecture for encoder and decoder parts in all these models. In particular, we add a sub-network to model the conditional prior in both iVAE and the proposed \FancyName. We further assign a linear SCM sub-network to model the relations among latent causal variables in the proposed \FancyName. We trained these 5 models on the dataset described above, with different numbers of segments and different numbers of latent causal variables. For each method, we use 5 different random seeds for data sampling. Figure \ref{fig:synthetic} shows the performance on two latent causal variables with different numbers of segment. The proposed \FancyName obtains the score 0.96 approximately for all the different numbers of segment. In contrast, $\beta$-VAE, VAE and CausalVAE fail to achieve a good estimation of the true latent variables, since they are not identifiable. iVAE obtains unsatisfactory results, since its theory only holds for i.i.d. latent variables. Figure \ref{fig:synthetic:b} shows the performance in recovering latent causal variables on synthetic data with different numbers of the latent causal variables. Figure \ref{fig:synthetic:c} shows the structural Hamming distance (SHD), a standard metric for comparing graphs via their adjacency matrices, for the recovered skeletons (the undirected version of the graph with all directed edges replaced by undirected edges), causal graphs, and completed partial directed acyclic graphs (CPDAGs, which capture the set of all DAGs that are Markov equivalent), by \FancyName and CausalVAE. Since CausalVAE is not identifiable, it obtains fully connected graphs, the SHD of the recovered CPDAG is the same as one of the recovered skeletons.

\subsubsection{Experiments on Changes of Part of Weights} \label{localchanges}
\begin{figure}
  \centering
  \includegraphics[width=0.5\linewidth]{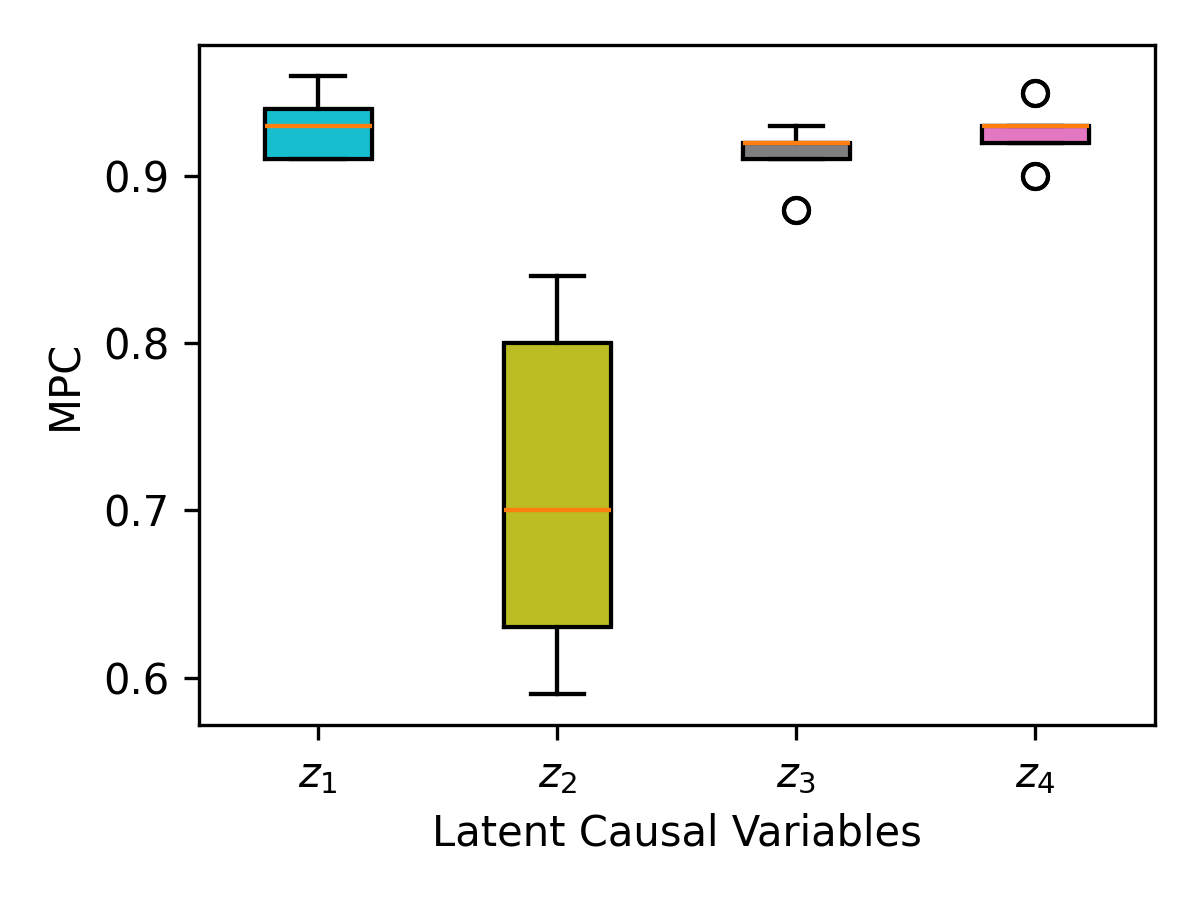}~\\
  \includegraphics[width=0.5\linewidth]{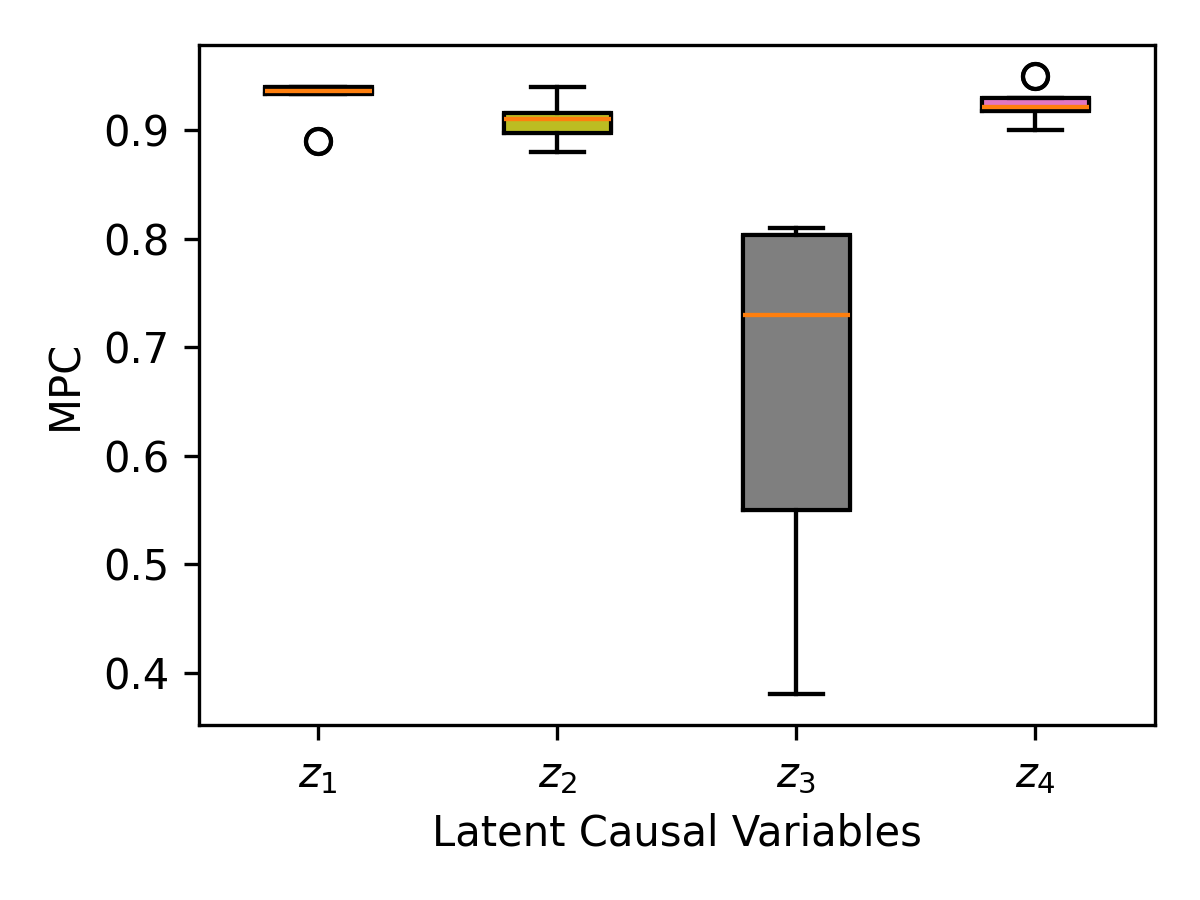}~\\
  \includegraphics[width=0.5\linewidth]{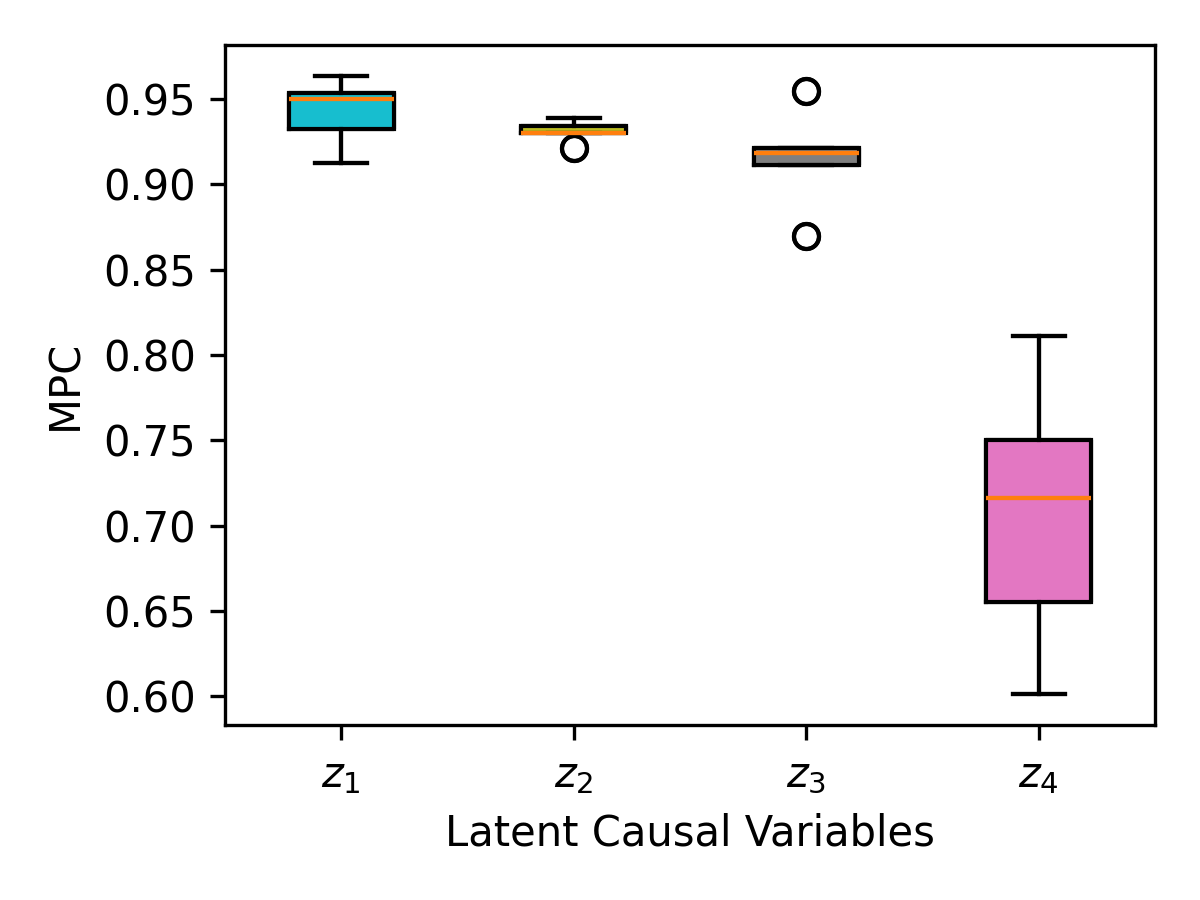}\\
  \caption{\textit{Case 1:} Performances of the proposed method with the change of part of weights. The ground truth of the causal graph is $z_1\rightarrow z_2\rightarrow z_3\rightarrow z_4$. From top to bottom: keeping weight on $z_1\rightarrow z_2$ (and accordingly, $z_2\rightarrow z_3$, and $z_3\rightarrow z_4$) unchanged. Those results are consistent with the analysis of partial identifiability results in Theorem \ref{theory2}.}
  \label{fig:partialc}
\end{figure}

\begin{figure}
  \centering
  \includegraphics[width=0.5\linewidth]{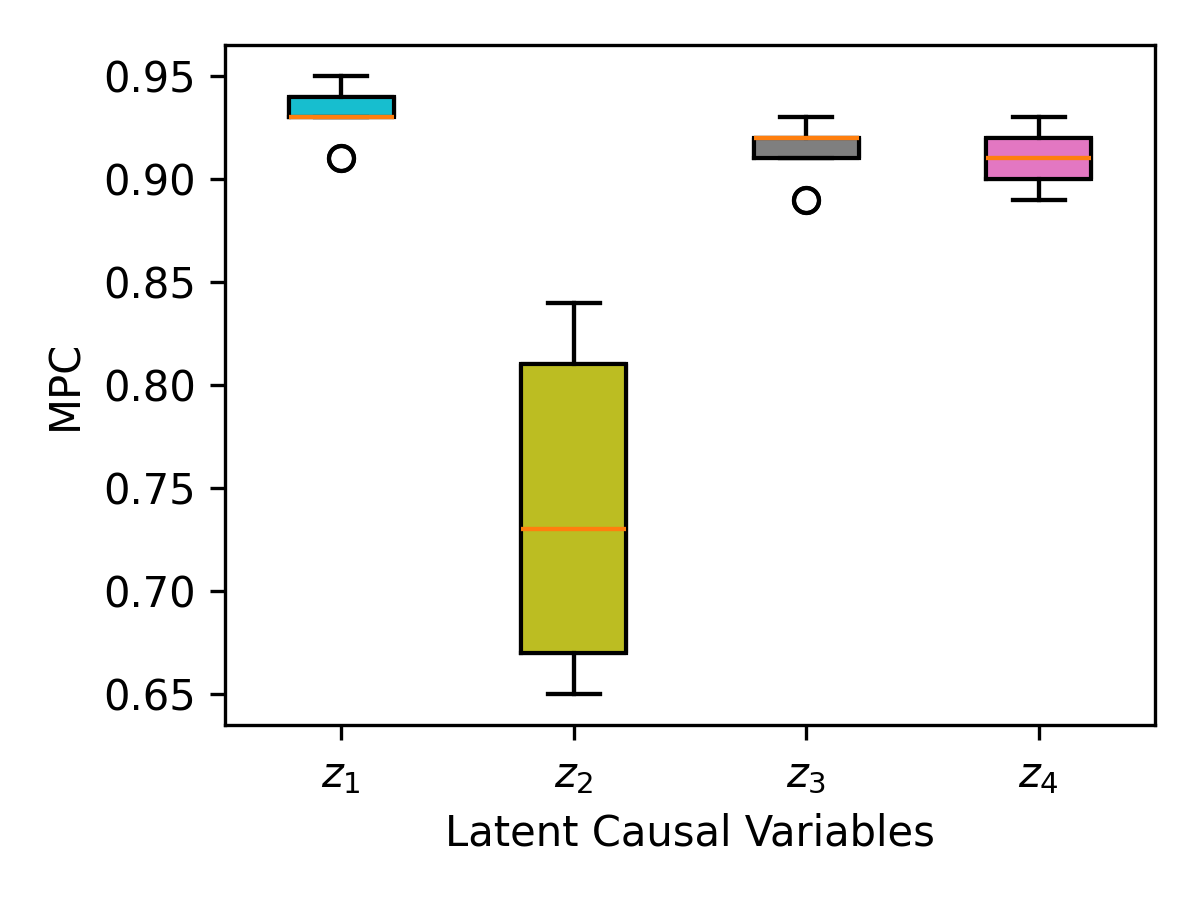}~\\
  \includegraphics[width=0.5\linewidth]{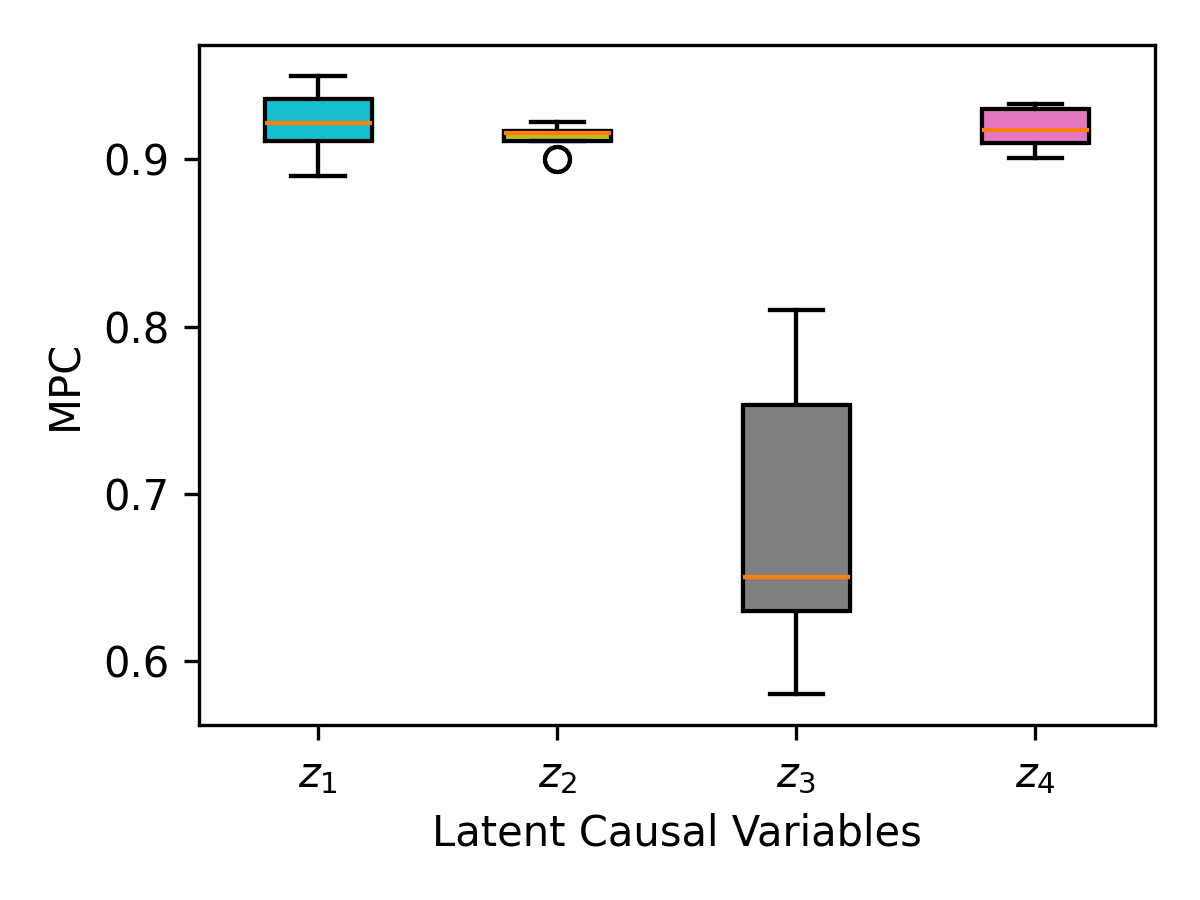}~\\
  \includegraphics[width=0.5\linewidth]{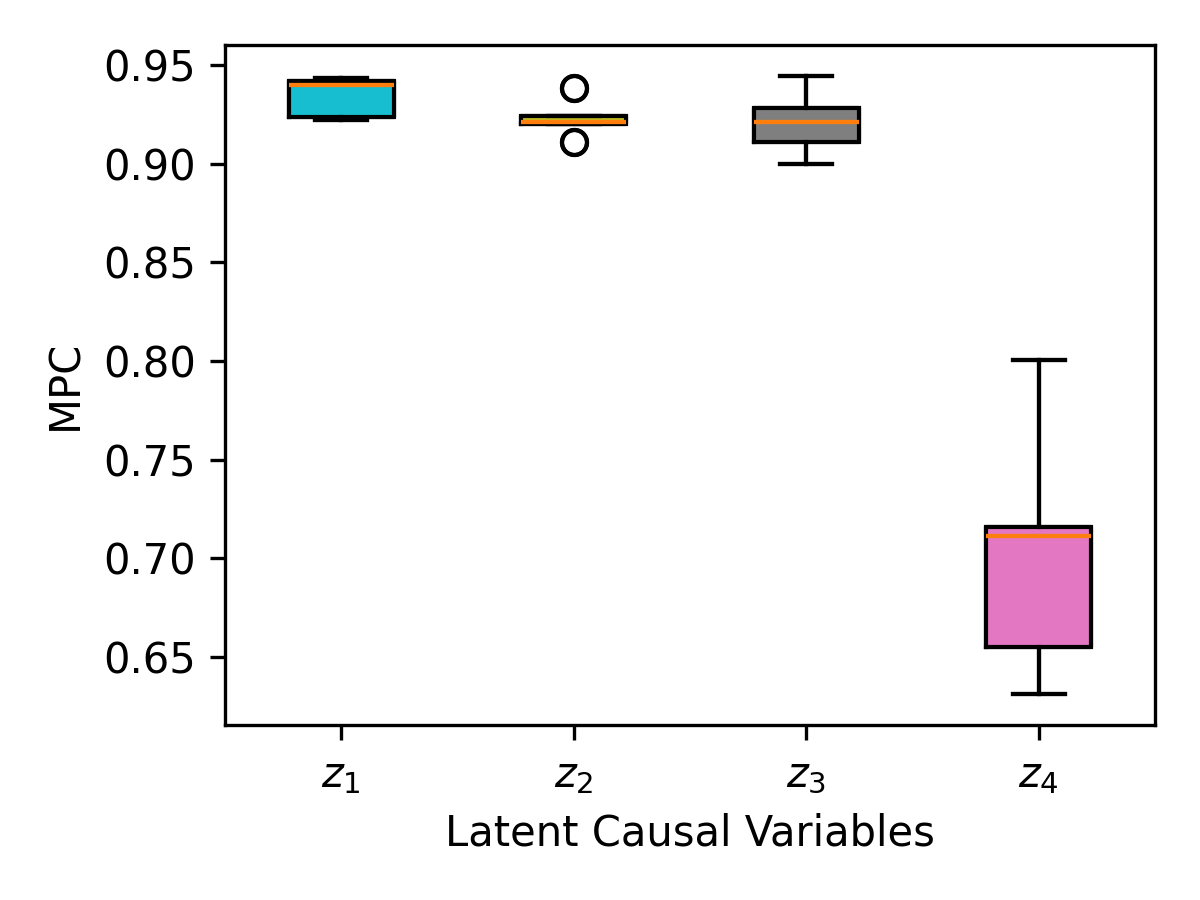}\\
  \caption{\textit{Case 2:} Performances of the proposed method with the change of part of weights. The ground truth of the causal graph is $z_1\rightarrow z_2\rightarrow z_3\rightarrow z_4$. From top to bottom: weight on $z_1\rightarrow z_2$ (and accordingly, $z_2\rightarrow z_3$, and $z_3\rightarrow z_4$) includes a non-zero constant part and changing part, \ie, $\lambda(\mathbf{u}) + b$. Those results are consistent with the analysis of partial identifiability results in Theorem \ref{theory2}.}
  \label{fig:partialc2}
\end{figure}

In this part, we conduct experiments on changes of part of weighs, to verify the results in Theorem \ref{theory2}. To this end, we consider two cases on the following causal graph: $z_1 \rightarrow z_2\rightarrow z_3\rightarrow z4$. The data details are similar to those of the 4-dimensional case mentioned in Appendix \ref{expbc}, except that $z_3$ is a parent node of $z_4$. \textit{Case 1:} the weight on the edge of $z_1 \rightarrow z_2$ (and accordingly, $z_2 \rightarrow z_3$, $z_3 \rightarrow z_4$) remains unchanged across $\mathbf{u}$, while the weights on the remaining edges change across $\mathbf{u}$, \textit{Case 2:} the weight on the edge of $z_1 \rightarrow z_2$ (and accordingly, $z_2 \rightarrow z_3$, $z_3 \rightarrow z_4$) includes a constant part (\ie, $\lambda_{1,2}(\mathbf{u}) + b$, $b$ is a non-zero constant.) across $\mathbf{u}$, while the weights on the remaining edges change across $\mathbf{u}$. Again, the network architecture, optimization, and hyper-parameters are the same as used for experiments of Figure \ref{fig:synthetic} (a). Figures \ref{fig:partialc} and \ref{fig:partialc2} show the performance of the proposed method in the settings corresponding to \textit{Case 1} and \textit{Case 2} mentioned above. According to MPC, we observe that unchanged weights in \textit{Case 1} or weights involving a constant term in \textit{Case 2} lead to non-identifiability. Conversely, changing weights contribute to the identifiability of the corresponding nodes. These empirical results align with the partial identifiability results in Theorem \ref{theory2}.

\subsubsection{Experiments on i.i.d. $z_i$}
\begin{figure}[!htp]
  \centering
{\includegraphics[width=0.5\textwidth]{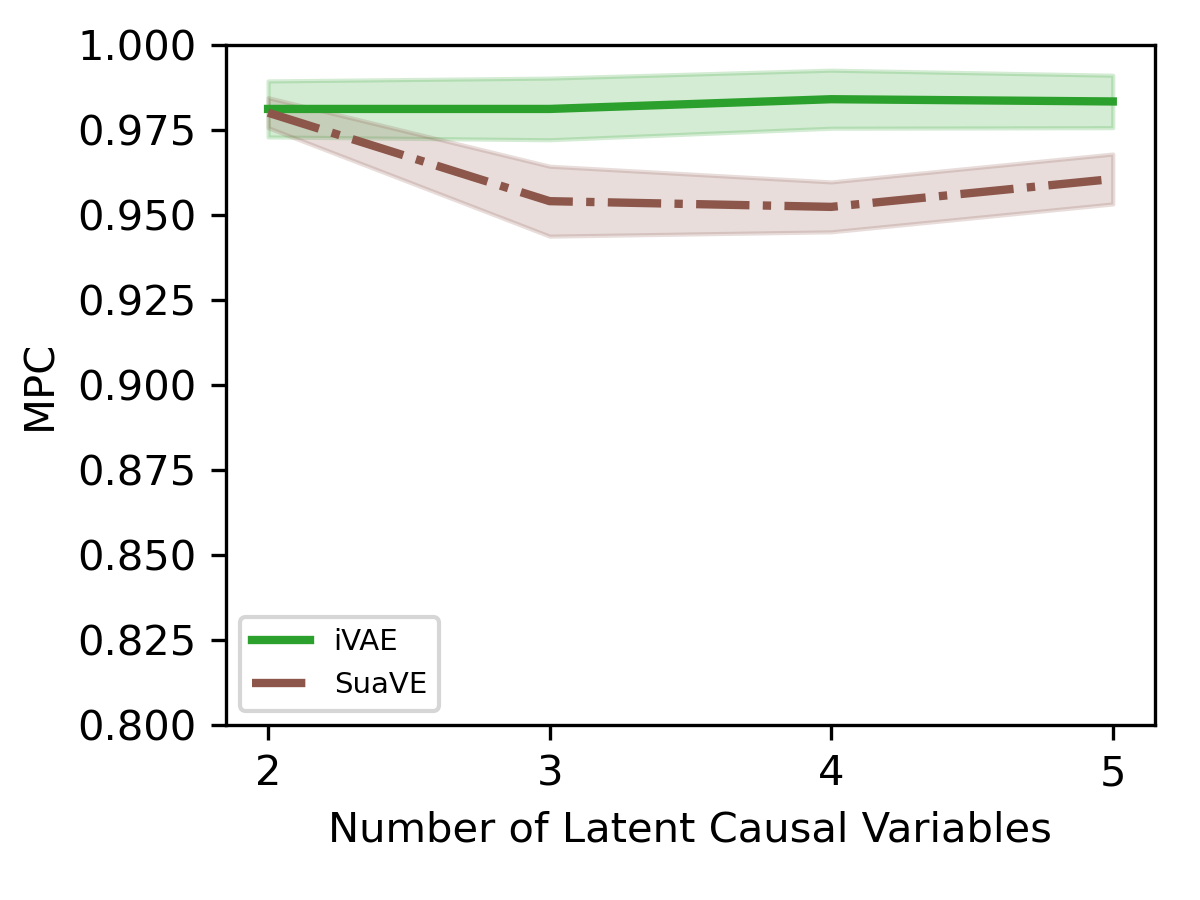}}\
  \caption{Performances of the proposed \FancyName and iVAE in recovering the independent latent causal variables conditional $\mathbf{u}$.} 
  \label{fig:syniidcase}
\end{figure}

Our identifiability result includes nonlinear ICA as a special case, in which the latent causal variables do not exhibit any causal relationships among themselves, effectively collapsing the causal structure into a fully disconnected graph. To verify this point, we conduct experiments on the case where any two latent causal variables $z_i$ are independent given $\mathbf{u}$. Data details are the same as mentioned in section \ref{expbc}, but we here enforce $\lambda_{i,j}(\mathbf{u})=0$ for each $i,j$. Again, the network architecture, optimization and hyper-parameters are the same as section \ref{expbc}, except for learning rate (here we set $1r=1e-2$). Figure \ref{fig:syniidcase} shows the performances of the proposed \FancyName and iVAE. We can see that iVAE is slightly better than the proposed \FancyName, both obtain satisfying results in terms of MPC.

\subsubsection{Performance for violated assumption on distribution} To further understand the assumptions of our identifiability result, we conduct experiments in a setting where the assumption on distribution is violated. Table \ref{table} shows the performance when the distribution assumption is violated. We observe that violating the distribution assumption leads to unsatisfactory performance in terms of MPC. The possible reasons for the results with Uniform (Laplace, Gamma) Noise may be: 1) the proposed method enforces Gaussian prior and posterior distributions to approximate non-Gaussian noise, and 2) the settings with Uniform (Laplace, Gamma) noise may be inherently non-identifiable.

\begin{table}[h]
\caption{Performance for violated assumptions. Uniform (Laplace, Gamma) Noise: the distribution of latent noise variables. 
Matching Assumptions: a setting that matches our theoretical assumptions.}
\label{table}
\vskip 0.15in
\begin{center}
\begin{small}
\begin{sc}
\begin{tabular}{lcr}
\toprule
Generative process & MPC(Mean $\pm$ Std) \\
\midrule
Uniform Noise     & 0.83 $\pm$ 0.09 \\
Laplace Noise     &  0.83 $\pm$ 0.03  \\
Gamma Noise       & 0.82 $\pm$ 0.05 \\
Matching Assumptions& 0.95 $\pm$ 0.01 \\
\bottomrule
\end{tabular}
\end{sc}
\end{small}
\end{center}
\end{table}

\subsection{Image Data} 
\begin{figure}[!htp]
  \centering
\includegraphics[width=0.95\textwidth]{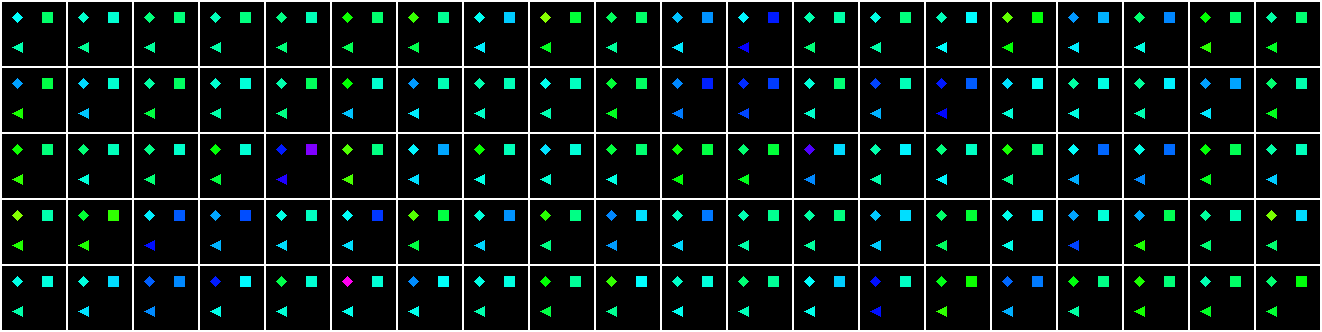}
  \caption{Demonstration of the chemistry environment. The colors (states) of the objects change according to the causal graph: the 'diamond' causes the ‘triangle’, and the ‘triangle’ causes the 'square'.} 
  \label{fig:rlsamples}
\end{figure}

We further verify our identifiability results and the proposed method on high-dimensional image data from the chemistry dataset proposed in \citet{ke2021systematic}, which is corresponding to simple chemical reactions where the state of an element can cause changes to another variable’s state. The environment consists of a number of objects whose positions are kept fixed and thus, uniquely identifiable, while the colors (states) of the objects affected by this variable (according to the causal graph) can change. We use a weight-variant linear Gaussian model as mentioned in section \ref{expbc} for a 3-dimensional case, to generate the latent variables, and obtain the corresponding observational images. A visualization of the environment can be found in Figure \ref{fig:rlsamples}.

\begin{figure}
  \centering
\includegraphics[width=0.3\linewidth]{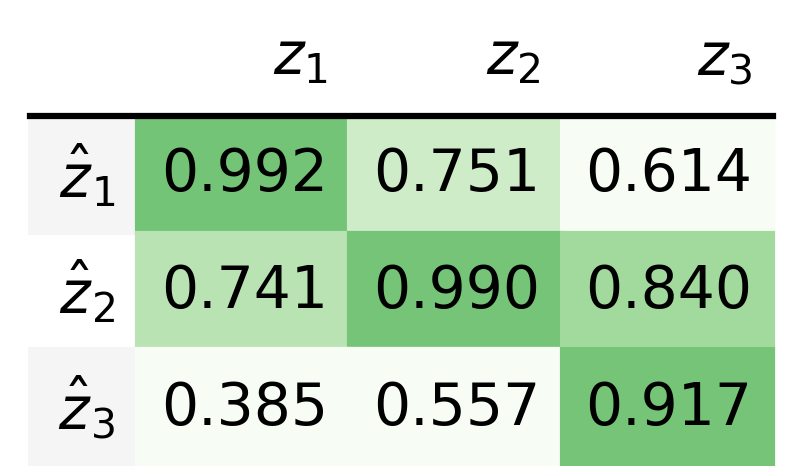}~
\includegraphics[width=0.3\linewidth]{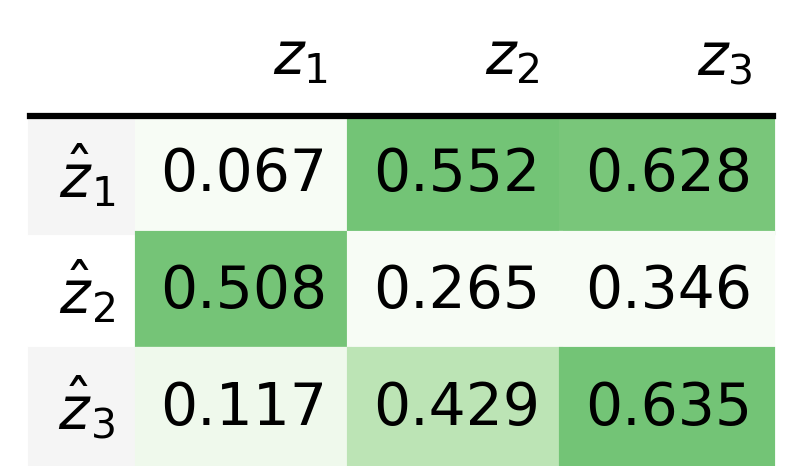}~
\includegraphics[width=0.3\linewidth]{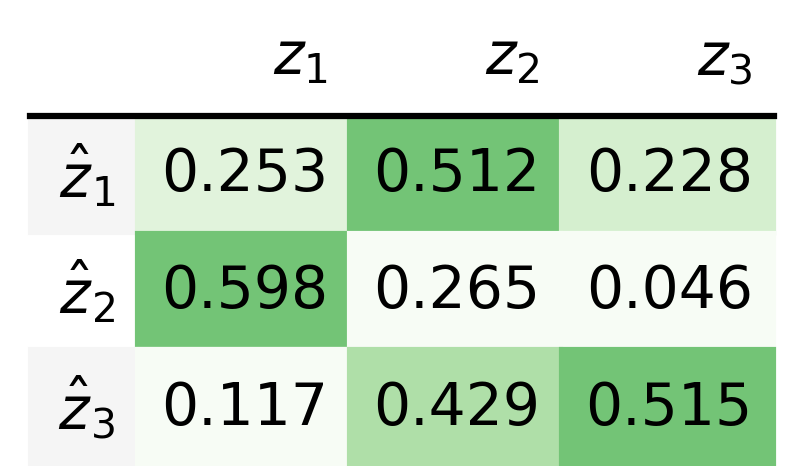}\\
\begin{minipage}{0.3\linewidth}
    \centering 
    Ours
  \end{minipage}
  \hfill 
  \begin{minipage}{0.3\linewidth}
    \centering 
    $\beta$-VAE
  \end{minipage}
  \hfill
  \begin{minipage}{0.3\linewidth}
    \centering 
    CausalVAE
  \end{minipage}
  \caption{MPC obtained by different methods on the image dataset. Supported by our identifiability analysis, the proposed method achieves satisfactory MPC, indicating its success in recovering latent causal representations. }\label{fig:interv_mpc}
\end{figure}

\begin{figure}[!htp]
  \centering
\includegraphics[width=0.3\textwidth]{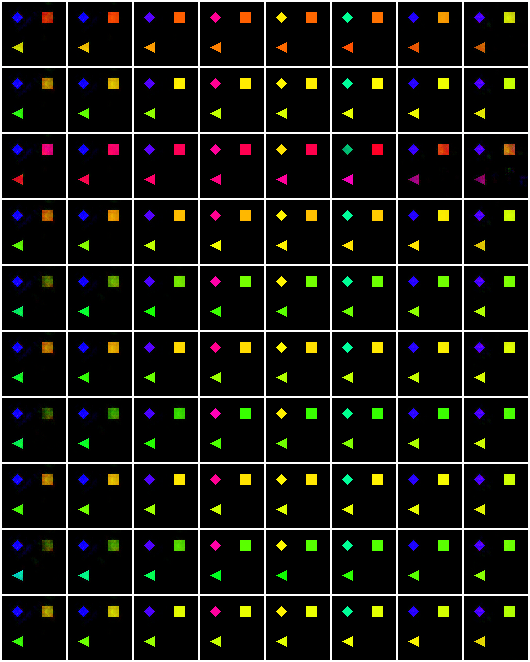}
\includegraphics[width=0.3\textwidth]{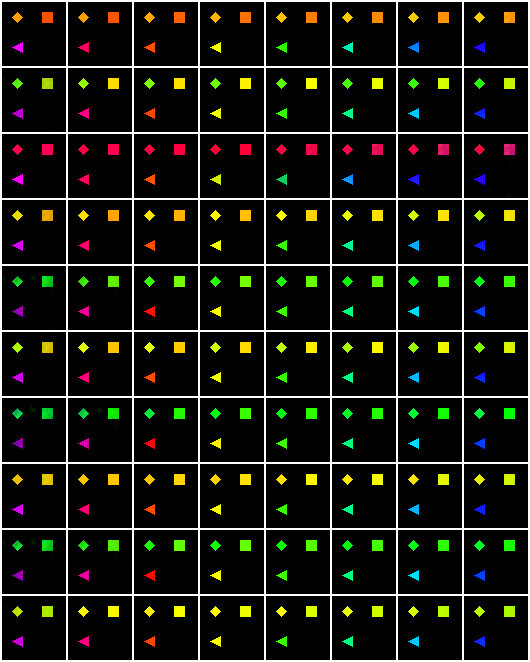}
\includegraphics[width=0.3\textwidth]{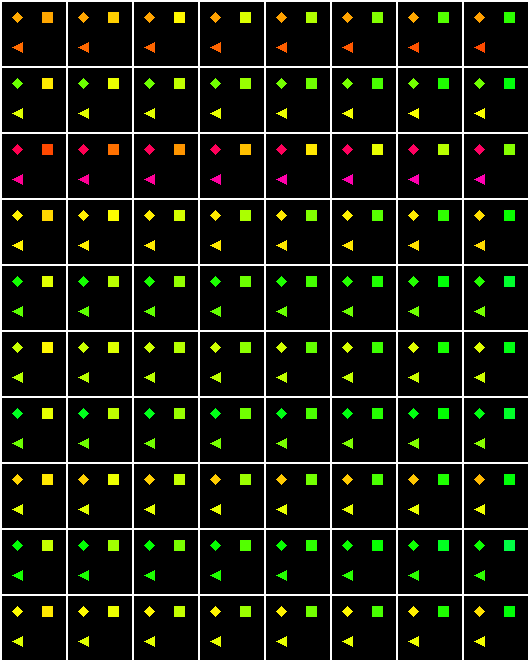}
  \caption{Intervention results on the learned latent variables by the proposed method \FancyName. From left to right: intervention on the learned $z_1,z_2,z_3$, respectively. The vertical axis denotes different samples, The horizontal axis denotes enforcing different values on the learned causal representation. We can see that: 1) (from the left) the changes of $z_1$ (the 'diamond') cause the change of both $z_2$ and $z_3$ (the 'triangle' and the 'square'). 2) (from the middle) the changes of $z_2$ (the 'triangle') only cause the change of $z_3$ (the 'square'). 3) (from the right) the changes of $z_3$ (the 'square') do not cause the change of $z_1$ and $z_2$ (the 'diamond' and the 'triangle').} 
  \label{fig:rlinter}
\end{figure}

\begin{figure}[!htp]
  \centering
\includegraphics[width=0.3\textwidth]{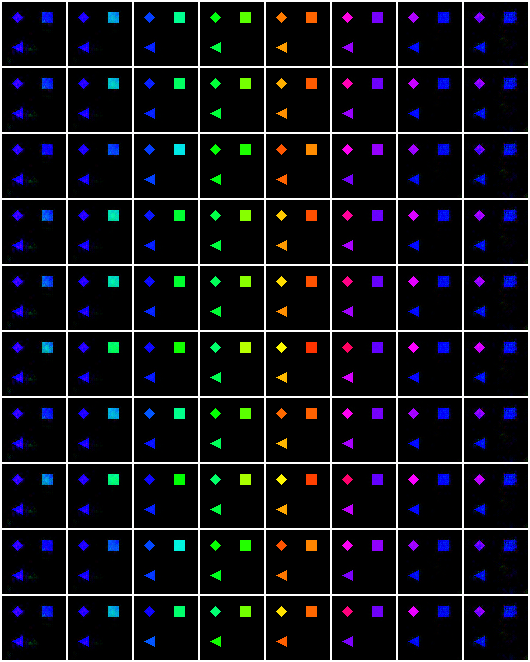}
\includegraphics[width=0.3\textwidth]{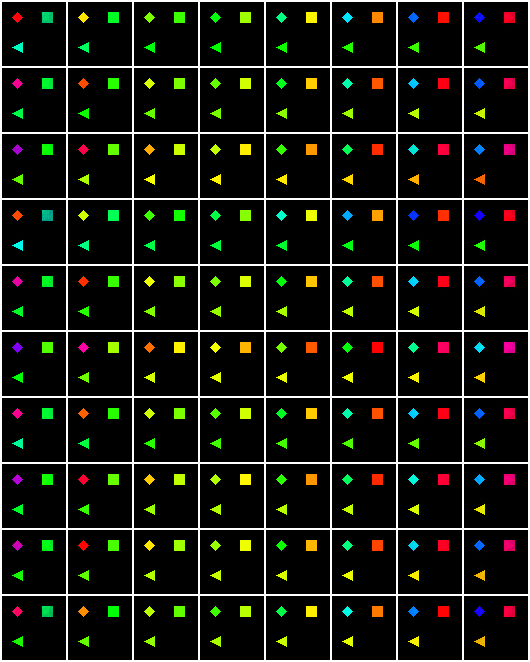}
\includegraphics[width=0.3\textwidth]{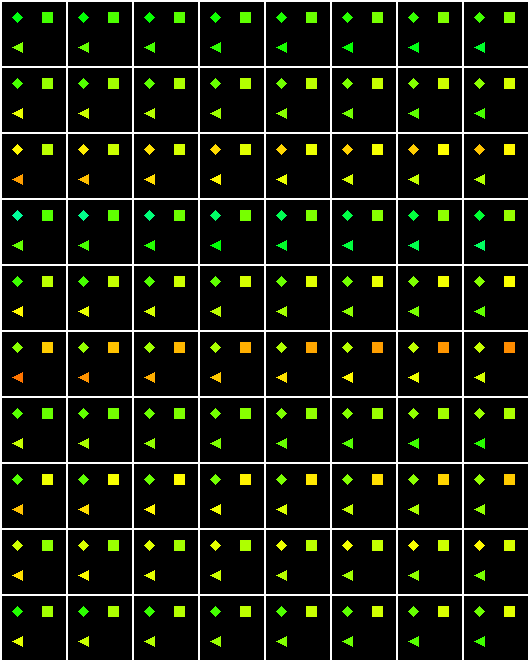}
\caption{Traversal results on the learned latent variables by $\beta$-VAE \citep{higgins2016beta}. From left to right: Traversal on the learned $z_1,z_2,z_3$, respectively. The vertical axis denotes different samples, The horizontal axis denotes enforcing different values on the learned causal representation. We can see that from the left to the right with changes of anyone $z_i$, the remaining two variables always changes, which means that each learned variable by $\beta$-VAE \citep{higgins2016beta} tend to be mixtures of the true underlying latent causal variables.} 
\label{fig:rlinter_beta}
\end{figure}

\begin{figure}[!htp]
  \centering
\includegraphics[width=0.3\textwidth]{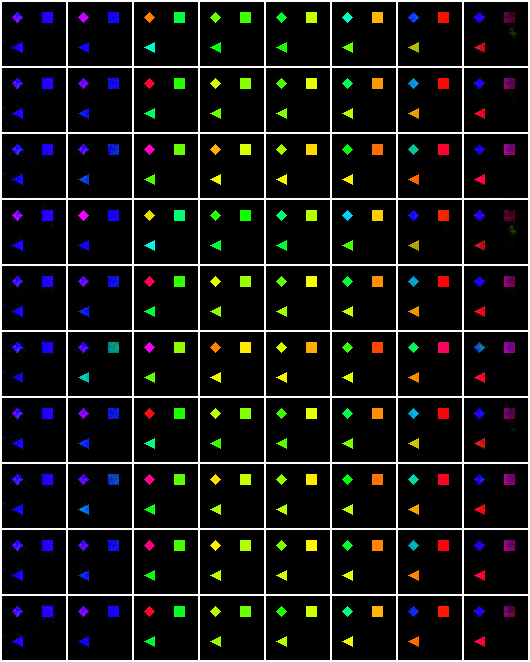}
\includegraphics[width=0.3\textwidth]{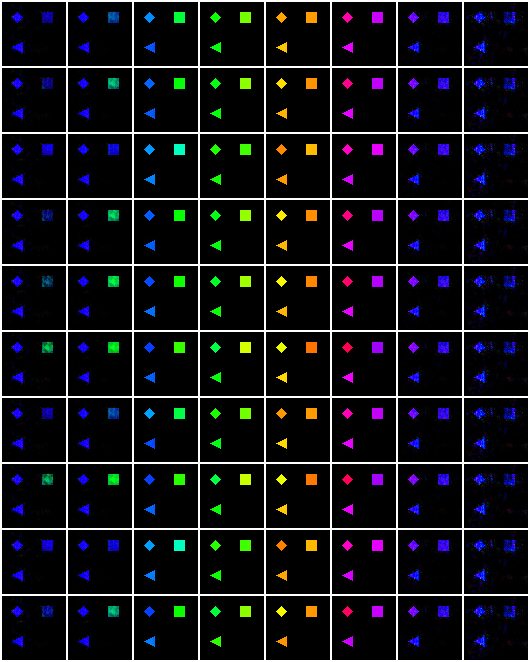}
\includegraphics[width=0.3\textwidth]{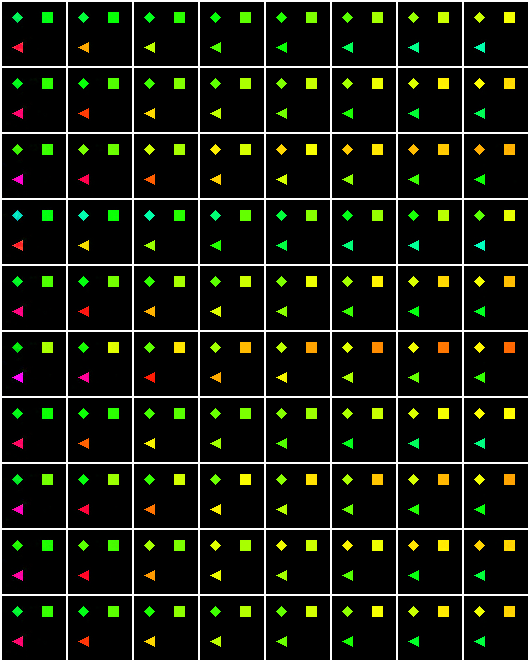}
\caption{Traversal results on the learned latent variables by CausalVAE \citep{yang2020causalvae}. From left to right: Traversal on the learned $z_1,z_2,z_3$, respectively. The vertical axis denotes different samples, The horizontal axis denotes enforcing different values on the learned causal representation. Again, we can see that from the left to the right with changes of anyone $z_i$, the remaining two latent variables also change. These results mean that each learned variable tend to be mixtures of the true underlying latent causal variables, and CausalVAE \citep{yang2020causalvae} fails to recovering latent causal structure.} 
\label{fig:rlinter_cvae}
\end{figure}

Figure \ref{fig:interv_mpc} shows the MPC obtained by different methods. The proposed method performs better than $\beta$-VAE \citep{higgins2016beta} and CausalVAE \citep{yang2020causalvae}. In term of MPC, the proposed method successfully learns causal representations, while the others fails to learn latent causal variables. To further verify whether the learned causal structure by the proposed method is consistent with the true underlying latent structure, we investigate the results of intervening on each learned variable. Figure \ref{fig:rlinter} shows the intervention results on each learned latent variable by the proposed \FancyName. We can see from Figure \ref{fig:rlinter} that:
\begin{itemize}
    \item Intervention on $z_1$ (the 'diamond') cause the change of both $z_2$ and $z_3$ (the colors of 'triangle' and 'square'), as depicted in the left in Figure \ref{fig:rlinter}.
    \item Intervention on $z_2$ (the 'triangle') only cause the change of $z_3$ (the color of 'square'), in the medium in Figure \ref{fig:rlinter}.
    \item Intervention on the changes of $z_3$ (the 'square') \textit{do not} cause the change of $z_1$ and $z_2$ (the 'diamond' and the 'triangle'), in the right in Figure \ref{fig:rlinter}
\end{itemize}  
All these observations are consistent with the ground truth latent causal structure, \ie, $z_1 \rightarrow z_2 \rightarrow z_3$, together with results in Figure \ref{fig:interv_mpc}, demonstrating that the proposed method successfully learns latent causal structures. Similar, we inversigate the change in reconstructions while traversing each learned latent variable obtained by $\beta$-VAE \citep{higgins2016beta} and CausalVAE \citep{yang2020causalvae}. Figure \ref{fig:rlinter_beta} and \ref{fig:rlinter_cvae} show traversal results on the learned representations by $\beta$-VAE \citep{higgins2016beta} and CausalVAE \citep{yang2020causalvae}, respectively. we observed that changing each learned variable results in changes to the colors of all objects. Together with the results in Figure \ref{fig:interv_mpc}, these suggests that $\beta$-VAE \citep{higgins2016beta} and CausalVAE \citep{yang2020causalvae} tend to learn representations that entangle the true underlying latent causal variables and fail to recover the true underlying latent causal structure.

\subsection{fMRI Data} Following \citet{ghassami2018multi}, we applied the proposed method to fMRI hippocampus dataset \citep{linkLP}, which contains signals from six separate brain regions: perirhinal cortex (PRC), parahippocampal cortex (PHC), entorhinal cortex (ERC), subiculum (Sub), CA1, and CA3/Dentate Gyrus (DG) in the resting states on the same person in 84 successive days. Each day is considered as different $\mathbf{u}$, thus $\mathbf{u}$ is a 84-dimensional vector. Since we are interested in recovering latent causal variables, we treat the six signals as latent causal variables by applying a random nonlinear mapping on them to obtain observed data. We then apply various methods on the observed data to recover the six signals. It is worth noting that, with respect to Assumption (v), one of the key conditions in our theory, the context of fMRI might be interpreted as providing an approximate rather than a literal reference condition. For example, different cognitive tasks or resting states are known to substantially suppress or activate specific effective connections between brain regions. While such modulations may not constitute perfect hard interventions, they may weaken certain causal influences, serving as empirical analogues of the reference condition
assumed in our theory. Our empirical results suggest that such approximate satisfaction of the assumption might already be sufficient in practice.

\begin{figure}[h]
\centering
{\includegraphics[width=0.36\textwidth]{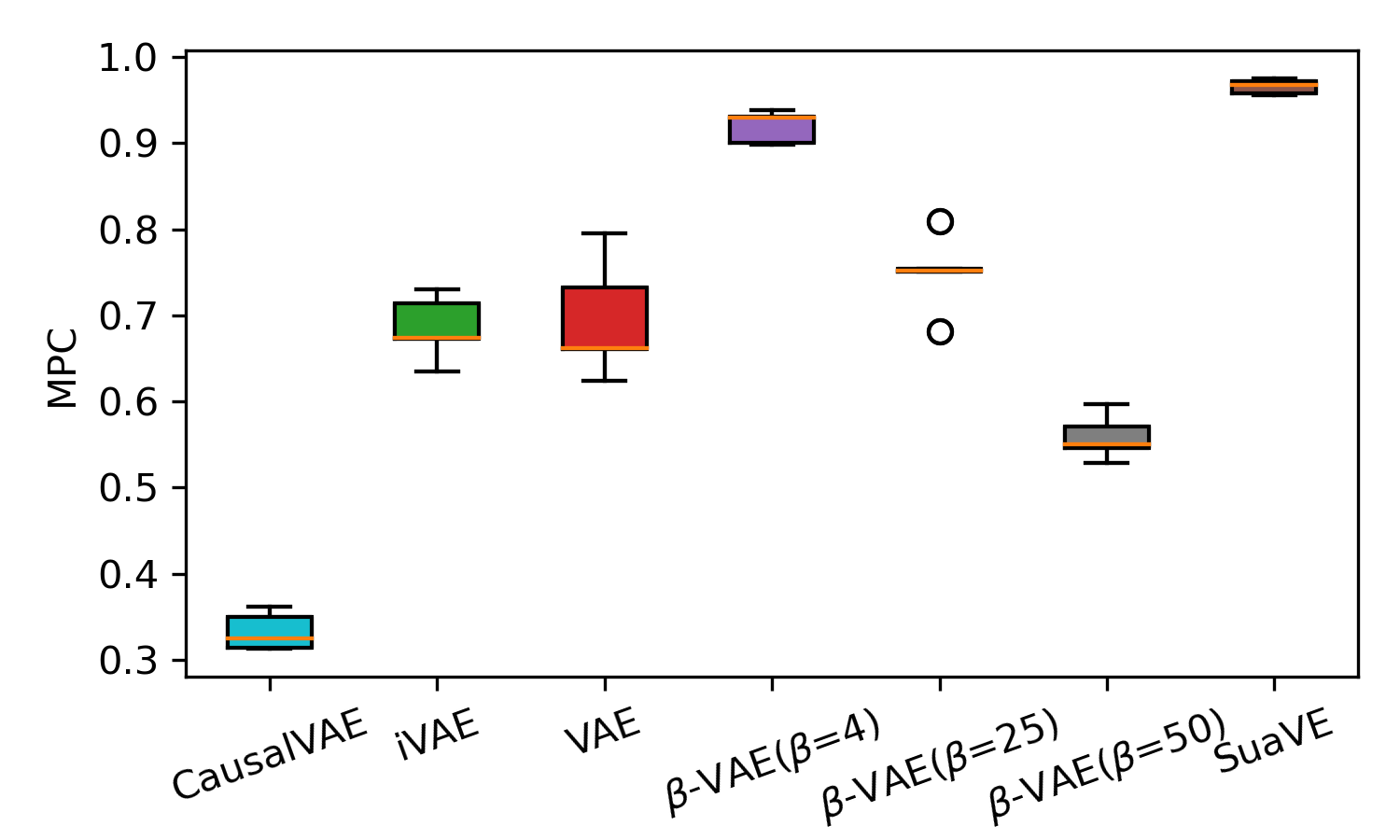}}\  
{\includegraphics[width=0.25\textwidth]{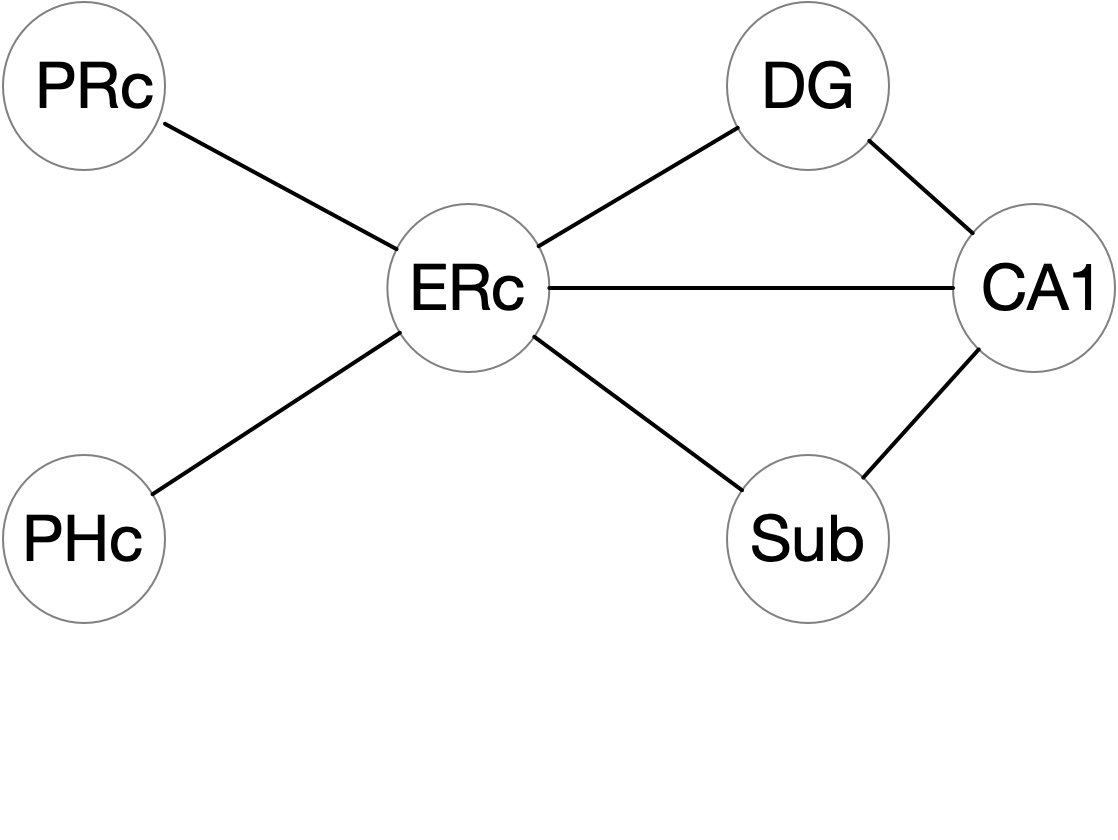}}\ 
{\includegraphics[width=0.25\textwidth]{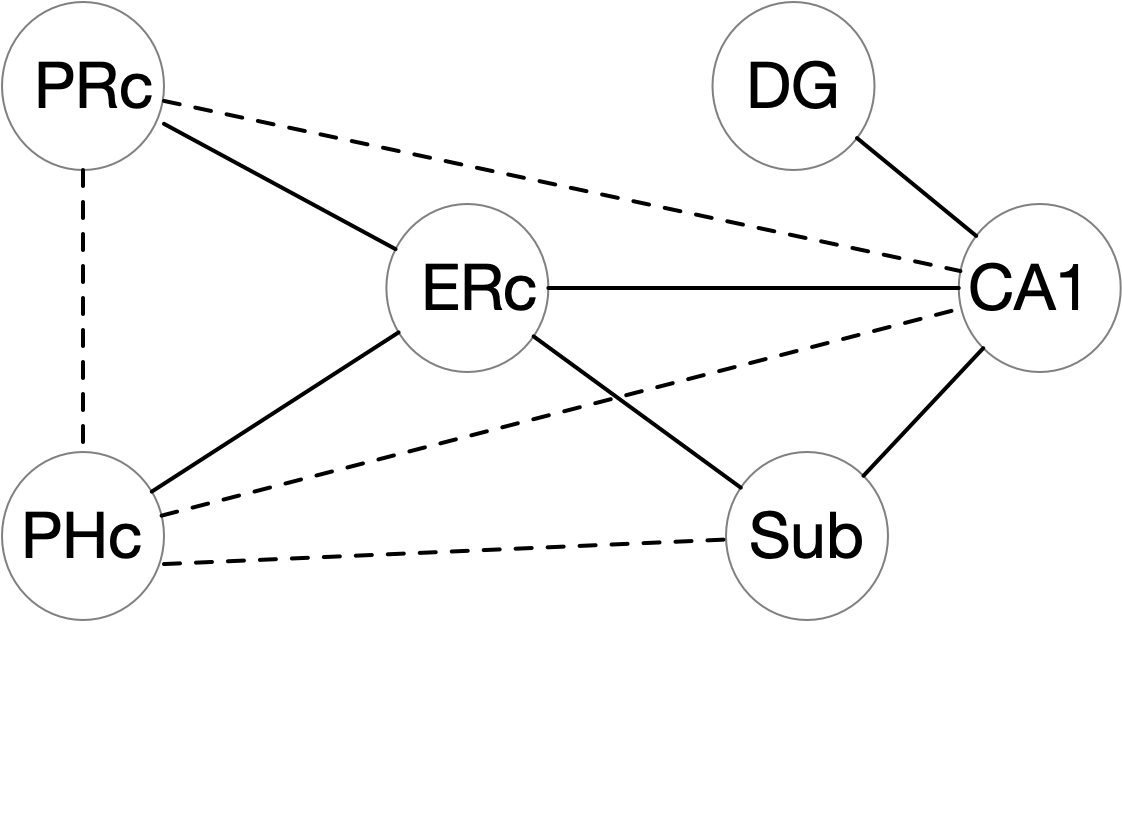}}
  \caption{Performance on fMRI Data. (a) the performance of the proposed \FancyName in comparison to iVAE, $\beta$-VAE, VAE and CausalVAE in recovering the latent six signals. (b) The skeleton of the anatomical connections given in \citet{bird2008hippocampus}. (c) The recovered skeleton by the proposed \FancyName, where the dashed lines indicate errors.}
  \label{fig:fmri}
\end{figure}

Figure \ref{fig:fmri} (a) shows the performance of the proposed \FancyName in comparison to iVAE, $\beta$-VAE, VAE and CausalVAE in recovering the latent six signals. $\beta$-VAE aims to recover independent latent variables, and it obtains an interesting result: enforcing independence (\eg, $\beta = 25, 50$) leads to worse MPC, and relaxing it (though contracting to its own independence assumption) improves the result (\eg, $\beta = 4$). This is because the latent variables given the time index are not independent in this dataset.

We further verify the ability of the proposed method to recover causal skeletons. We use the anatomical connections \citep{bird2008hippocampus} as a reference as shown in Figure \ref{fig:fmri} (b), since causal connections among the six regions should not exist if there is no anatomical connection. 
To obtain the final estimated graph, note that in the computing MPC, we have solved a linear sum assignment problem, which matches the semantic information of the estimated variables with the one of the true underlying latent variables. After matching all semantic information, we can use a threshold (0.1) to removed a edge separately across $\mathbf{u}$, and once an edge appears in more than $70\%$ of all $\mathbf{u}$, we keep this edge in the final graph. Figure \ref{fig:fmri} (c) shows an example of the estimated causal skeleton by the proposed \FancyName. Averaging over 5 different random seeds for the used nonlinear mapping from latent space to observed space, structural Hamming distance (SHD) obtained by the proposed \FancyName is \textbf{5.0 $\pm$ 0.28}. In contrast, iVAE, $\beta$-VAE and VAE assume latent variables to be independent, and thus can not obtain the causal skeleton. We further analyse the result of $\beta$-VAE with $\beta=4$ by using the PC algorithm \citep{Spirtes00} to discovery skeleton on the recovered latent variables, and obtain \textbf{SHD = 5.8 $\pm$ 0.91}. This means that: 1) $\beta$-VAE with $\beta=4$ does not ensure the strong independence among the recover latent variables as it is expected, 2) the proposed \FancyName is an effective one-step method to discovery the skeleton. The unsupervised version of CausalVAE is not identifiable, and can also not ensure the relations among the recover variables to be causal skeleton in principle. In experiments we found that CausalVAE always obtains fully-connected graphs for the 5 different random seeds, for which SHD is \textbf{9.0 $\pm$ 0}.

\begin{figure}[!htp]
  \centering
{\includegraphics[width=0.8\textwidth]{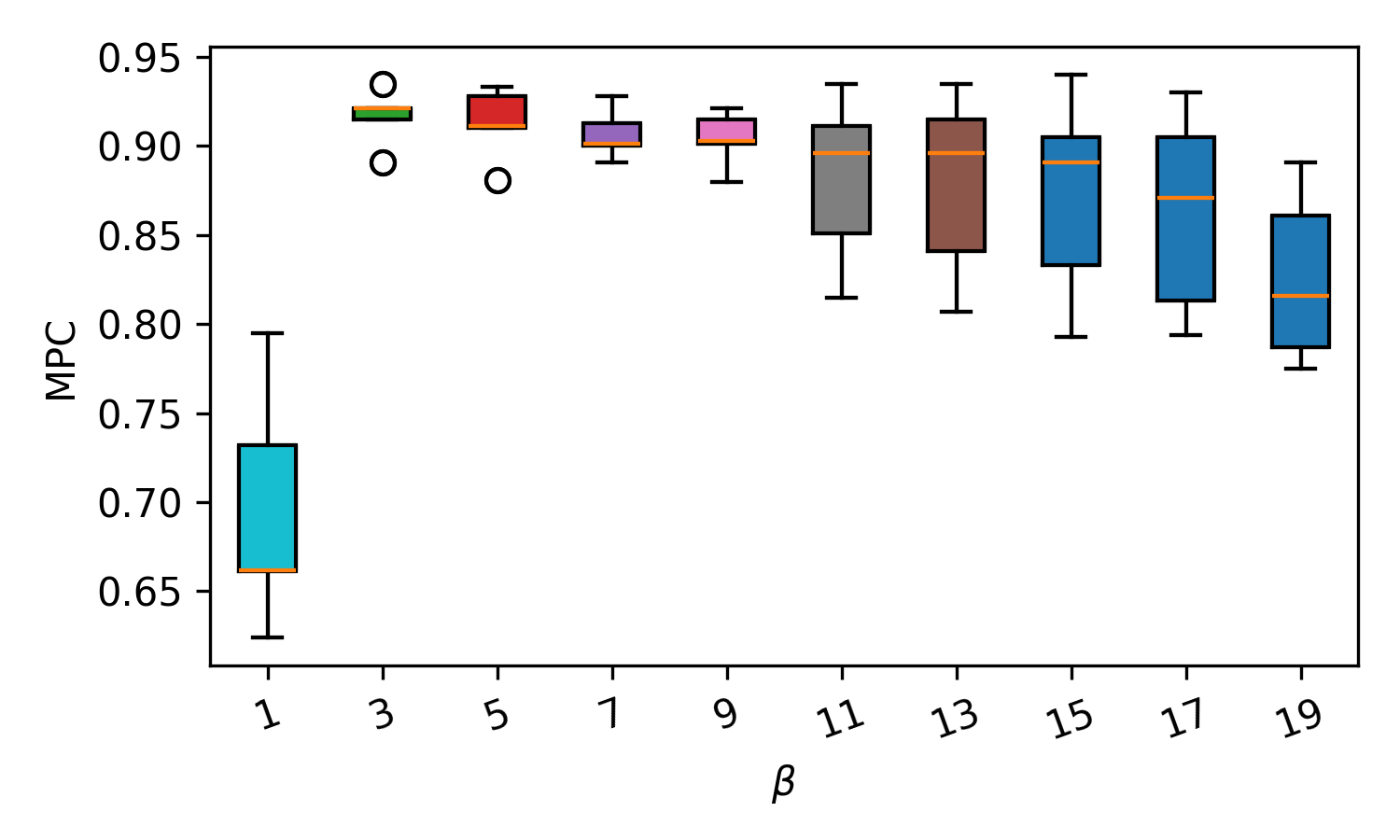}}\
  \caption{Performances of $\beta$-VAE with different $\beta$ in recovering fMRI data.} 
  \label{fig:betafMRI}
\end{figure}

We further investigate the performance of $\beta$-VAE across different values of $\beta$. Figure \ref{fig:betafMRI} illustrates the performance trend as $\beta$ varies. $\beta$-VAE suffers from two critical limitations: In practical applications, determining an appropriate value for $\beta$ is challenging due to the inherent uncertainty in the underlying latent variables. This lack of guidance can lead to suboptimal model performance if $\beta$ is not carefully tuned. For $\beta \geq 1$, as the value of $\beta$ increases, there is an inevitable trade-off between disentanglement and reconstruction quality. Specifically, improving disentanglement typically comes at the cost of degraded reconstruction, as compared to a standard VAE. This trade-off is particularly problematic in scenarios where preserving reconstruction quality is crucial, making it difficult to balance these competing objectives in real-world tasks. These challenges highlight the practical difficulties in applying $\beta$-VAE effectively, especially when considering applications where the precise nature of the latent structure is unknown and both disentanglement and reconstruction are essential.

\begin{figure}[h]
\centering
\includegraphics[width=0.3\textwidth]{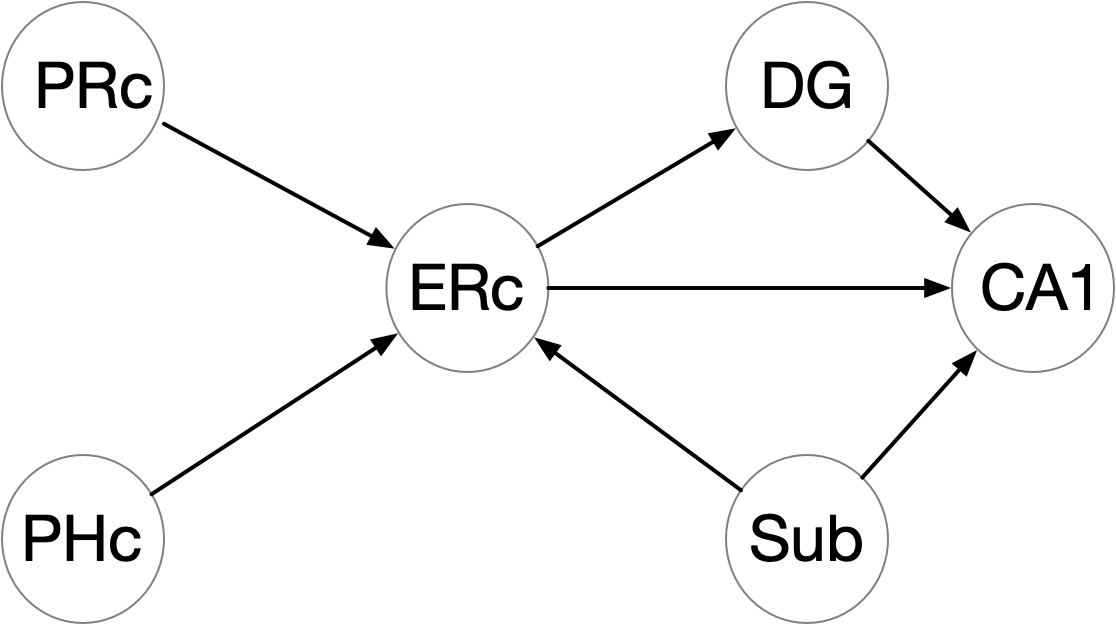}\  
\includegraphics[width=0.3\textwidth]{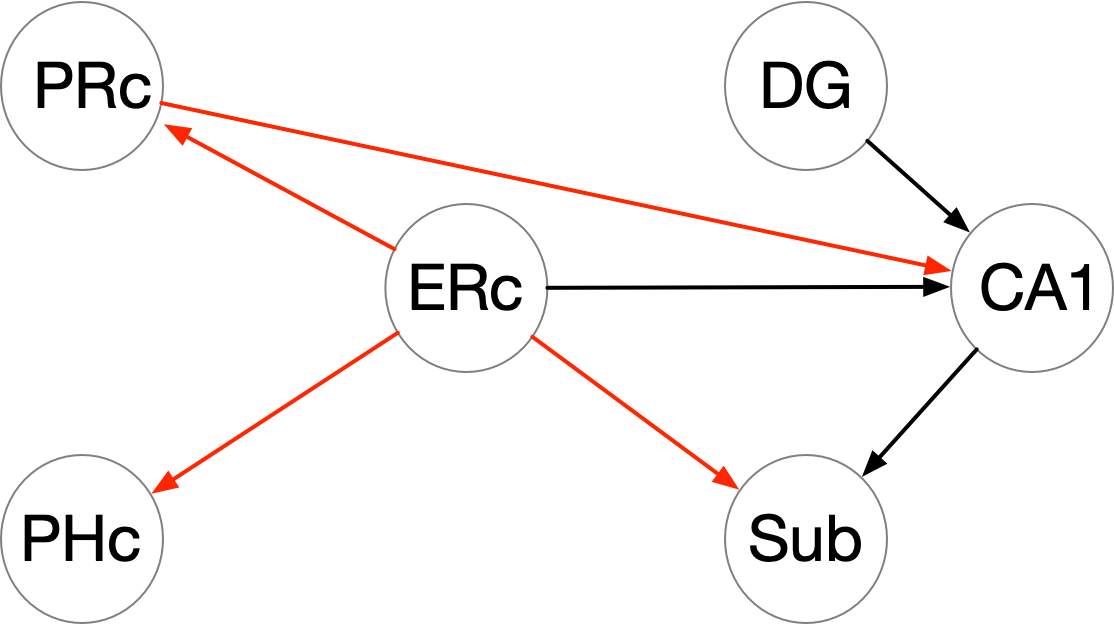}\ 
\includegraphics[width=0.3\textwidth]{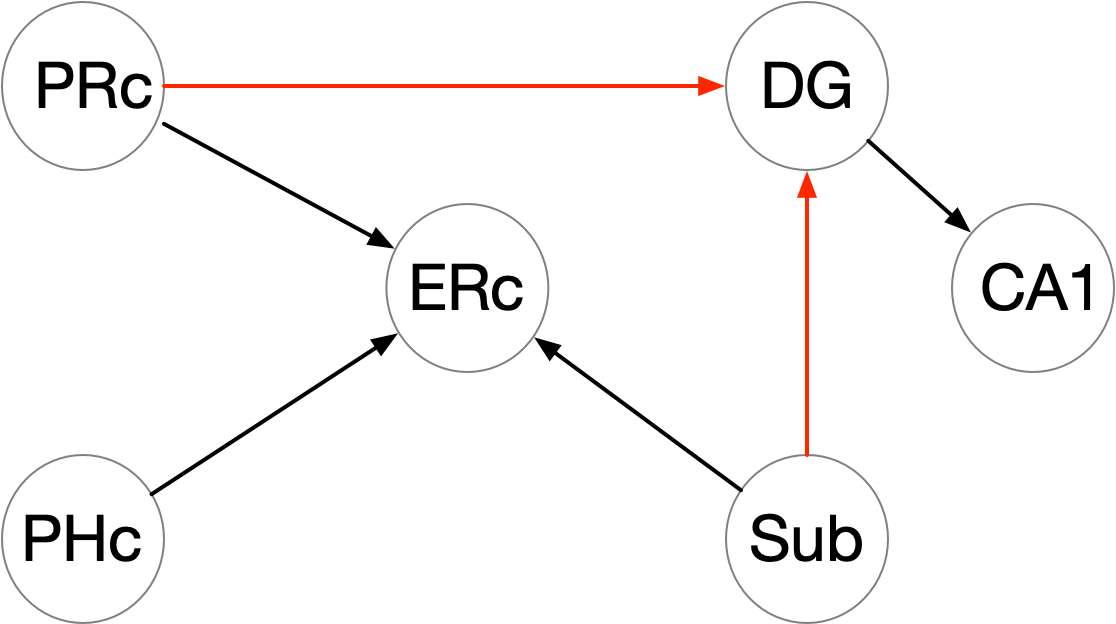}
  \caption{Recovered graphs on fMRI Data. (a) the anatomical connections given in \cite{bird2008hippocampus}. (b) The recovered causal graph by the proposed \FancyName. (c) The recovered causal graph reported in \citep{ghassami2018multi}. Note that the setting in (b) treats fMRI data as latent variables, while the setting in (c) treats fMRI data as observed variables.}
  \label{fig:fmridag}
\end{figure}

Figure \ref{fig:fmridag} (a) show the edges in the anatomical ground truth reported in \cite{ghassami2018multi}. Figure \ref{fig:fmridag} (b) show the latent recovered edges by the proposed \FancyName, in the setting where we treat fMRI data as latent causal variables. For comparison, Figure \ref{fig:fmridag} (c) shows the recovered causal graph reported in \citep{ghassami2018multi}, in the setting where fMRI data are treated as observed variables. That is, the setting in (b) is more challenging than the setting in (c).

\section{Conclusion}
Identifying latent causal variables from observed data is generally impossible without additional assumptions. Motivated by recent advances in the identifiability of nonlinear ICA, we investigate three fundamental sources of indeterminacy in the latent space, \ie, transitivity, permutation, and scaling, which offer key insights into the challenges and opportunities in identifying latent causal variables. Among these, we show transitivity as the primary obstacle to identifiability and propose a new research direction that leverages changes in causal influences between latent variables to overcome it. Under weight-variant linear Gaussian models and mild conditions inspired by nonlinear ICA, we prove that latent causal variables can be identified up to permutation and scaling. Importantly, we show that partial identifiability remains possible even when only a subset of the causal influences varies, making our results more applicable in practice. Building on these theoretical insights, we introduce Structural caUsAl Variational autoEncoder (SuaVE), a novel method for jointly learning latent causal variables and their underlying causal graph. Empirical results on both synthetic and fMRI data validate our identifiability theory and demonstrate the practical effectiveness of the proposed method.



\acks{This project was partially funded by the Responsible AI Research Centre (Yuhang Liu, Zhen Zhang, Anton van den Hengel, and Javen Qinfeng Shi). Dong Gong was partially supported by the Australian Government through the Australian Research Council Discovery Early Career Researcher Award (DECRA)(DE230101591). Mingming Gong was partially supported by the Australian Government through the Australian Research Council Discovery Projects (DP240102088). The authors also thank the anonymous reviewers for their constructive feedback. }


\newpage

\appendix
\section{}
\subsection{The proof of Theorem \ref{theory1}}
\label{prooflinear}
 The proof of Theorem \ref{theory1} is done in three steps. Step I is to show the identifiability result of nonlinear ICA holds in our setting where there is an additional mapping from $\mathbf{n}$ to $\mathbf{z}$, and the mapping changes across $\mathbf{u}$. Step II consists of building a linear transformation between the true latent variables $\mathbf{z}$ and the estimated ones $\mathbf{\hat z}$. Step III shows that the linear transformation in Step II can be reduced to permutation transformation by using the result of Step I and the assumption \ref{itm:lambda}.

For convenience, let us discuss the property of the mapping from $\mathbf{n}$ to $\mathbf{z}$ first, which plays a key role in the proof. Let function $\mathbf{g}_{\mathbf{u}}$ denote the mapping from $\mathbf{n}$ to $\mathbf{z}$. That is:
\begin{equation}
    \mathbf{z} = \mathbf{g}_{\mathbf{u}}(\mathbf{n}) = (\mathbf{I}-\boldsymbol{\lambda}^{T}(
    \mathbf{u}
    ))^{-1}\mathbf{n},
\label{appenx:n2z}
\end{equation}
where $\mathbf{I}$ denotes the identity matrix. Due to DAG constraints in a causal system and the assumption of linear models on $\mathbf{z}$, we have 1) the determinant of the Jacobian matrix is equal to 1, \ie, $|\det{\mathbf{J}}_{\mathbf{g_{u}}^{-1}}(\mathbf{z})|=|\det{\mathbf{J}}_{\mathbf{g_{u}}}(\mathbf{u})| = 1$, 2) the mapping $\mathbf{g}_{\mathbf{n}}$ is invertible.

{\bf{Step I:}}
Suppose we have two sets of parameters $\theta = ({\mathbf{f},\mathbf{T}_{\mathbf{n}},\boldsymbol{\lambda},\boldsymbol{\beta}},\boldsymbol{\varepsilon})$ and $\hat{\theta} = ({\mathbf{\hat {f}},\mathbf{\hat T}_{\mathbf{n}},\boldsymbol{\hat \lambda},\boldsymbol{\hat \beta}})$ such that $p_{({\mathbf{f},\mathbf{T}_{\mathbf{n}},\boldsymbol{\lambda},\boldsymbol{\beta}},\boldsymbol{\varepsilon})}(\mathbf{x}|\mathbf u) = p_{({\mathbf{\hat {f}},\mathbf{\hat T}_{\mathbf{n}},\boldsymbol{\hat \lambda},\boldsymbol{\hat \beta}})}(\mathbf x|\mathbf u)$ for all pairs $(\mathbf{x},\mathbf{u})$. Due to the assumptions \ref{itm:eps} and \ref{itm:bijective}, by expanding these expressions via the change of variables formula and
taking the logarithm we find:

\begin{align}
\log |{\mathbf{J}_{{\mathbf{f}}^{-1}}(\mathbf{x})}| +  \log p_{\boldsymbol{\varepsilon}}(\boldsymbol{\varepsilon}) + & \log |{\mathbf{J}_{\mathbf{g}^{-1}_{\mathbf{u}}}(\mathbf{z})}| + \log p_{\mathbf{(\mathbf{T}_\mathbf{n},\boldsymbol{ \beta})}}({\mathbf{n}|\mathbf u})  \notag \\ &=\log {|\mathbf{J}_{\mathbf{\hat f} ^{-1}}(\mathbf{x})|} + \log {|\mathbf{J}_{(\mathbf{\hat g}_{\mathbf{u}}) ^{-1}}(\mathbf{\hat z})|} + \log p_{\mathbf{(\mathbf{\hat T}_\mathbf{ n},\boldsymbol{ \hat \beta})}}({\mathbf{\hat n}|\mathbf u}),\label{proofn}
\end{align}
Using the exponential family Eq. \eqref{nef} to replace $p_{\mathbf{(\mathbf{T}_\mathbf{n},\boldsymbol{ \beta})}}({\mathbf{n}|\mathbf u})$, we have:

\begin{align}
&\log |{\mathbf{J}_{{\mathbf{f}}^{-1}}(\mathbf{x})}| +  \log p_{\boldsymbol{\varepsilon}}(\boldsymbol{\varepsilon}) + \log |{\mathbf{J}_{\mathbf{g}^{-1}_{\mathbf{u}}}(\mathbf{z})}| + \mathbf{T}^{T}_{\mathbf{n}}\big(\mathbf{n}\big)\boldsymbol{\eta}_{\mathbf{n}}(\mathbf u) -\log {Z_{\mathbf{n}}(\boldsymbol{\beta},\mathbf{u})}
= \label{proofn21} \\ 
&\log {|\mathbf{J}_{\mathbf{\hat f} ^{-1}}(\mathbf{x})|} + \log {|\mathbf{J}_{(\mathbf{\hat g}_{\mathbf{u}}) ^{-1}}(\mathbf{\hat z})|} + \mathbf{\hat T}_{\mathbf{n}}^{T}\big(\mathbf{\hat n}\big)\boldsymbol{ \hat \eta}_{\mathbf{n}}(\mathbf u) -\log {\hat Z_{\mathbf{ n}}(\boldsymbol{\hat \beta},\mathbf{u})},\label{proofn22}
\end{align}
then by using 
\begin{align}
|\det{\mathbf{J}}_{\mathbf{g_{u}}^{-1}}(\mathbf{z})|=1, |\det{\mathbf{J}}_{\mathbf{\hat g_{u}}^{-1}}(\mathbf{z})| =1.
\end{align}
here the latter holds since $\hat g_{u}$ must be the same function class as $g_{u}$, Eqs. \ref{proofn21} and \ref{proofn22} can be rewriteed as:
\begin{align}
&\log { |\mathbf{J}_{\mathbf{f}^{-1}}(\mathbf{x})| } + \log p_{\boldsymbol{\varepsilon}}(\boldsymbol{\varepsilon}) + \mathbf{T}^{T}_{\mathbf{n}}\big(\mathbf{n}\big)\boldsymbol{\eta}_{\mathbf{n}}(\mathbf u) -\log {Z_{\mathbf{n}}(\boldsymbol{\beta},\mathbf{u})}
= \label{proofn211} \\ 
&\log {|\mathbf{J}_{\mathbf{
\hat f}^{-1}}(\mathbf{x})|} + \mathbf{\hat T}_{\mathbf{n}}^{T}\big(\mathbf{\hat n}\big)\boldsymbol{ \hat \eta}_{\mathbf{n}}(\mathbf u) -\log {\hat Z_{\mathbf{ n}}(\boldsymbol{\hat \beta},\mathbf{u})}.\label{proofn221}
\end{align}
The following proof is similar to the proof of \citet{sorrenson2020disentanglement}. For completeness, we present a slightly simplified proof. With different points in assumption \ref{itm:nu}, we can subtract this expression with $\mathbf{u}_{\mathbf{n},0}$ by the expression with $l_{\mathbf{n}}$, $l_{\mathbf{n}}=1,...,2n$. Since the Jacobian and noise terms do not depend on $\mathbf{u}$, the Jacobian and noise terms will be canceled out:
\begin{align}
&\log \frac{Z_{\mathbf{n}}(\boldsymbol{\beta},\mathbf{u}_{\mathbf{n},0})}{Z_{\mathbf{n}}(\boldsymbol{\beta},\mathbf{u}_{\mathbf{n},l_{\mathbf{n}}})}
+\mathbf{T}^{T}_{\mathbf{n}}\big(\mathbf{n}\big)\big(\boldsymbol{\eta}_{\mathbf{n}}(\mathbf{u}_{\mathbf{n},l_{\mathbf{n}}})-\boldsymbol{\eta}_{\mathbf{n}}(\mathbf{u}_{\mathbf{n},0})\big)
= \\ 
& \log \frac{\hat Z_{\mathbf{ n}}(\boldsymbol{\hat \beta},\mathbf{u}_{\mathbf{n},0})} {\hat Z_{\mathbf{ n}}(\boldsymbol{\hat \beta},\mathbf{u}_{\mathbf{n},l_{\mathbf{n}}})}+\mathbf{\hat T}_{\mathbf{n}}^{T}\big(\mathbf{\hat n}\big)\big(\boldsymbol{\hat \eta}_{\mathbf{n}}(\mathbf{u}_{\mathbf{n},l_{\mathbf{n}}})-\boldsymbol{\hat \eta}_{\mathbf{n}}(\mathbf{u}_{\mathbf{n},0})\big).
\end{align}
Combining the $2n$ expressions into a single matrix equation we can write it in terms of $\mathbf{L}_{\mathbf{n}}$ from assumption (iii):
\begin{equation}
\mathbf{L}^{T}_{\mathbf{n}} \mathbf{T_{\mathbf{n}}\big(\mathbf{n}\big)}= \mathbf{\hat{L}}^{T}_{\mathbf{ n}} \mathbf{\hat{T}_{\mathbf{ n}}\big(\mathbf{\hat n}\big)} + \mathbf{b}_{\mathbf{n}},
\label{eqLn}
\end{equation}
where $\mathbf{L}_{\mathbf{n}}$ is defined as the assumption \ref{itm:nu}, $\mathbf{b}_{\mathbf{n},l_{\mathbf{n}}}=\log \frac{\hat Z_{\mathbf{ n}}(\boldsymbol{\hat \beta},\mathbf{u}_{\mathbf{n},0})Z_{\mathbf{n}}(\boldsymbol{\beta},\mathbf{u}_{\mathbf{n},l_{\mathbf{n}}})} {\hat Z_{\mathbf{ n}}(\boldsymbol{\hat \beta},\mathbf{u}_{\mathbf{n},l_{\mathbf{n}}})Z_{\mathbf{n}}(\boldsymbol{\beta},\mathbf{u}_{\mathbf{n},0})}$. Since $\mathbf{L}^{T}_{\mathbf{n}}$ is invertible by the assumption \ref{itm:nu}, we can multiply this expression by its inverse:
\begin{equation}
\mathbf{T_{\mathbf{n}}\big(\mathbf{n}\big)}= \mathbf{A}_{\mathbf{n}} \mathbf{\hat{T}_{\mathbf{ n}}\big(\mathbf{\hat n}\big)} + \mathbf{c}_{\mathbf{n}},
\label{eqAn}
\end{equation}
According to lemma 3 in \citet{khemakhem2020variational} and Proof for Theorem 1 in \citet{sorrenson2020disentanglement} we can show that $\mathbf{A}_{\mathbf{n}}$ has full rank. Since we assume the noise to be Gaussian, Eq. \eqref{eqAn} can be re-expressed as:
\begin{equation}
\left(                
  \begin{array}{c} 
    \mathbf{n} \\  
    \mathbf{n}^{2} \\ 
  \end{array}
\right)= \mathbf{A}_{\mathbf{n}} \left(    
  \begin{array}{c} 
    \mathbf{\hat n} \\  
    \mathbf{\hat{n}}^{2} \\ 
  \end{array}
\right) + \mathbf{c}_{\mathbf{n}},
\label{eqAn1n2}
\end{equation}
Then, by the contradiction between 1) for every $n_i$, there exists a polynomial with a degree at most 2, and 2) for every $n^2_i$, there also exists a polynomial with a degree at most 2 \citep{sorrenson2020disentanglement}, we can obtain: $n_i = A_{i,j}\hat n_j + c_i$. For simplicity, in the following we neglect the noise term $\boldsymbol{\varepsilon}$ in Eq \eqref{eq:Generative}. As a result, we can express the result in vector form as:
\begin{align}      
\mathbf{n} = \mathbf{P} \mathbf{\hat n} + \mathbf{c}_\mathbf{n}, 
\end{align}
where $\mathbf{P}$ denote the permutation matrix with scaling. More details can be found in \citet{sorrenson2020disentanglement}. Note that this simplification is for convenience only, the identifiability result still holds even when the noise term is included.

{\bf{Step II:}} Again, suppose we have two sets of parameters $\theta = ({\mathbf{f},\mathbf{T}_{\mathbf{z}},\boldsymbol{\lambda},\boldsymbol{\beta}})$ and $\hat{\theta} = ({\mathbf{\hat {f}},\mathbf{\hat T}_{\mathbf{z}},\boldsymbol{\hat \lambda},\boldsymbol{\hat \beta}})$ such that $p_{({\mathbf{f},\mathbf{T}_{\mathbf{z}},\boldsymbol{\lambda},\boldsymbol{\beta}})}(\mathbf{x}|\mathbf u) = p_{({\mathbf{\hat {f}},\mathbf{\hat T}_{\mathbf{z}},\boldsymbol{\hat \lambda},\boldsymbol{\hat \beta}})}(\mathbf x|\mathbf u)$ for all pairs $(\mathbf{x},\mathbf{u})$. Due to the assumptions \ref{itm:eps} and \ref{itm:bijective}, by expanding these expressions via the change of variables formula and taking the logarithm we have:
\begin{align}
\log {|\det \mathbf{J}_{\mathbf{f}^{-1}}(\mathbf{x})|} + \log p_{\mathbf{(\mathbf{T}_\mathbf{z},\boldsymbol{\lambda},\boldsymbol{\beta})}}({\mathbf{z}|\mathbf u})+\log p_{\boldsymbol{\varepsilon}}(\boldsymbol{\varepsilon}) =\log {|\det \mathbf{J}_{\mathbf{\hat f}^{-1}}(\mathbf{x})|} + \log p_{{(\mathbf{\hat T}_\mathbf{z},\boldsymbol{\hat \lambda},\boldsymbol{\hat \beta})}}({\mathbf{\hat z}|\mathbf u}),\label{proofz}
\end{align}
Using the exponential family Eq. \eqref{mgef} to replace $p_{\mathbf{(\mathbf{T}_\mathbf{z},\boldsymbol{\lambda},\boldsymbol{\beta})}}({\mathbf{z}|\mathbf u})$:
\begin{align}
&\log {|\det \mathbf{J}_{\mathbf{f}^{-1}}(\mathbf{x})|} + \mathbf{T}^{T}_\mathbf{z}(\mathbf{z})\boldsymbol{\eta}_\mathbf{z}(\mathbf u) -\log {Z_\mathbf{z}(\boldsymbol{\lambda},\boldsymbol{\beta},\mathbf{u})} + \log p_{\boldsymbol{\varepsilon}}(\boldsymbol{\varepsilon})
=\\
&\log {|\det \mathbf{J}_{\mathbf{\hat f}^{-1}}(\mathbf{x})|} + \mathbf{\hat T}^{T}_\mathbf{ z}(\mathbf{\hat z})\boldsymbol{ \hat \eta}_\mathbf{ z}(\mathbf u) -\log {\hat Z_\mathbf{ z}(\boldsymbol{\hat \lambda},\boldsymbol{\hat \beta},\mathbf{u})}.\label{prooftbeta2}
\end{align}
Let $\mathbf{u}_{\mathbf{z},0},\mathbf{u}_{\mathbf{z},1},...,\mathbf{u}_{\mathbf{z},k}$ be the points provided by the assumption \ref{itm:nz}. We define $\bar{\boldsymbol{\eta}}_{\mathbf{z}}(\mathbf{u}_{\mathbf{z},l_\mathbf{z}})={ \boldsymbol{\eta}}_{\mathbf{z}}(\mathbf{u}_{\mathbf{z},l_\mathbf{z}})-{\boldsymbol{\eta}_{\mathbf{z}}}(\mathbf{u}_{\mathbf{z},0})$. We plug each of those $\mathbf{u}_{\mathbf{z},l_\mathbf{z}}, l_\mathbf{z} = 1,..., k$ in Eq. \eqref{prooftbeta2} to obtain $k + 1$ equations. We subtract the first equation for $\mathbf{u}_{\mathbf{z},0}$ from the
remaining $k$ equations. For example, for $\mathbf{u}_{\mathbf{z},0}$ and $\mathbf{u}_{\mathbf{z},l_\mathbf{z}}$ we have two equations:
\begin{subequations}
\begin{equation}
\begin{aligned}
&\log {|\det \mathbf{J}_{\mathbf{ f}^{-1}}(\mathbf{x})|} +\mathbf{T}^{T}_{\mathbf{z}}(\mathbf{z})\boldsymbol{\eta}_{\mathbf{z}}(\mathbf{u}_{\mathbf{z},0}) -\log {Z_\mathbf{z}(\boldsymbol{\lambda},\boldsymbol{\beta},\mathbf{u}_{\mathbf{z},0})} + \log p_{\boldsymbol{\varepsilon}}(\boldsymbol{\varepsilon})
=\\
&\log {|\det \mathbf{J}_{\mathbf{\hat f}^{-1}}(\mathbf{x})|} + \mathbf{\hat T}^{T}_{\mathbf{z}}(\mathbf{\hat z})\boldsymbol{ \hat \eta}_{\mathbf{ z}}(\mathbf{u}_{\mathbf{z},0}) -\log {\hat Z_\mathbf{ z}(\boldsymbol{\hat \lambda},\boldsymbol{\hat \beta},\mathbf{u}_{\mathbf{z},0})}, \label{subtract1}
\end{aligned}
\end{equation}
\begin{equation}
\begin{aligned}
&\log {|\det \mathbf{J}_{\mathbf{ f}^{-1}}(\mathbf{x})|} +\mathbf{T}^{T}_{\mathbf{z}}(\mathbf{z})\boldsymbol{\eta}_{\mathbf{z}}(\mathbf{u}_{\mathbf{z},l_\mathbf{z}}) -\log {Z_\mathbf{z}(\boldsymbol{\lambda},\boldsymbol{\beta},\mathbf{u}_{\mathbf{z},l_\mathbf{z}})} + \log p_{\boldsymbol{\varepsilon}}(\boldsymbol{\varepsilon})
=\\
&\log {|\det \mathbf{J}_{\mathbf{\hat f}^{-1}}(\mathbf{x})|} + \mathbf{\hat T}^{T}_{\mathbf{z}}(\mathbf{\hat z})\boldsymbol{ \hat \eta}_{\mathbf{ z}}(\mathbf{u}_{\mathbf{z},l_\mathbf{z}}) -\log {\hat Z_\mathbf{ z}(\boldsymbol{\hat \lambda},\boldsymbol{\hat \beta},\mathbf{u}_{\mathbf{z},l_\mathbf{z}})}, \label{subtract2}
\end{aligned}
\end{equation}
\end{subequations}
Using Eq. \eqref{subtract2} subtracts Eq. \eqref{subtract1}, canceling the terms that do not include $\mathbf{u}$, we have:
\begin{equation}
\begin{aligned}
&\log\frac{Z_\mathbf{z}(\boldsymbol{\lambda},\boldsymbol{\beta},\mathbf{u}_{\mathbf{z},0})}{Z_\mathbf{z}(\boldsymbol{\lambda},\boldsymbol{\beta},\mathbf{u}_{\mathbf{z},l_\mathbf{z}})} +  \mathbf{T}_{\mathbf{z}}{(\mathbf{z})} \big(\boldsymbol{\eta}_{\mathbf{z}}(\mathbf{u}_{\mathbf{z},l_{\mathbf{n}}})-\boldsymbol{\eta}_{\mathbf{z}}(\mathbf{u}_{\mathbf{z},0})\big)  \\
& \qquad \qquad \qquad = \log\frac{\hat Z_\mathbf{ z}(\boldsymbol{\hat \lambda},\boldsymbol{\hat \beta},\mathbf{u}_{\mathbf{z},0})}{\hat Z_\mathbf{ z}(\boldsymbol{\hat \lambda},\boldsymbol{\hat \beta},\mathbf{u}_{\mathbf{z},l_\mathbf{z}})} +  \mathbf{\hat T}_{\mathbf{ z}}{(\mathbf{\hat z})} \big(\boldsymbol{\hat \eta}_{\mathbf{ z}}(\mathbf{u}_{\mathbf{z},l_{\mathbf{n}}})-\boldsymbol{\hat \eta}_{\mathbf{ z}}(\mathbf{u}_{\mathbf{z},0})\big)
\label{equtbeta}.
\end{aligned}
\end{equation}
Let $\mathbf{L}_\mathbf{z}$ be the matrix defined in Eq. \eqref{assumeiv} in assumption \ref{itm:nz}, and $\mathbf{\hat{L}}_\mathbf{\hat z}$ similarly ($\mathbf{\hat{L}}_\mathbf{\hat z}$ is not necessarily invertible). Define $b_{\mathbf{z},l_\mathbf{z}}={\log{\frac{\hat Z_{\mathbf{ z}}(\boldsymbol{\hat \lambda},\boldsymbol{\hat \beta}, \mathbf{u}_{\mathbf{z},0}) Z_{\mathbf{ z}}(\boldsymbol{\lambda},\boldsymbol{\beta}, \mathbf{u}_{\mathbf{z},l_\mathbf{z}})}{Z_{\mathbf{ z}}( \boldsymbol{\lambda},\boldsymbol{\beta},\mathbf{u}_{\mathbf{z},0}) \hat Z_{\mathbf{ z}}( \boldsymbol{\hat \lambda},\boldsymbol{\hat \beta},\mathbf{u}_{\mathbf{z},l_\mathbf{z}})}}}$. Expressing Eq. \eqref{equtbeta} for all points in matrix form, we get:
\begin{equation}
\mathbf{L}_{\mathbf{z}}^{T} \mathbf{T_{\mathbf{z}}(\mathbf{z})}= \mathbf{\hat{L}}_{\mathbf{ z}}^{T} \mathbf{\hat{T}_{\mathbf{z}}(\mathbf{\hat z})} + \mathbf{b}_{\mathbf{z}} .
\label{equLT}
\end{equation}
We multiply both sides of Eq. \eqref{equLT} by the inverse
of $\mathbf{L}^{T}_{\mathbf{z}}$ (by assumption \ref{itm:nz}) to find:
\begin{equation}
\mathbf{T_{\mathbf{z}}(\mathbf{z})}= \mathbf{A}_{\mathbf{z}} \mathbf{\hat{T}_{\mathbf{ z}}(\mathbf{\hat z})} + \mathbf{c}_{\mathbf{z}}, 
\label{equTA}
\end{equation}
where $\mathbf{A}_{\mathbf{z}} =(\mathbf{L}_{\mathbf{z}}^{T})^{-1}\mathbf{\hat{L}}_{\mathbf{\hat z}}^{T}$ and $\mathbf{c}_{\mathbf{z}} = (\mathbf{L}_{\mathbf{z}}^{T})^{-1} \mathbf{b}_{\mathbf{z}}$. As mentioned in Eq. \eqref{mgef}, the sufficient statistic $\mathbf{T}_{\mathbf{z}}(\mathbf{z})=[\mathbf{z};\vect(\mathbf{z}\mathbf{z}^{T})]$. In this case, the relationship Eq. \eqref{equTA} becomes:

\begin{equation}      
\left(                
  \begin{array}{c} 
    \mathbf{z} \\  
    \mathbf{z}^{2} \\ 
    \mathbf{z}_{i \neq j} \\
  \end{array}
\right) =
\mathbf{A}_{\mathbf{z}}\left(                
  \begin{array}{c} 
    \mathbf{\hat z} \\  
    \mathbf{\hat{z}}^{2} \\ 
     \mathbf{\hat z}_{i \neq j} \\
  \end{array}
\right)
+\mathbf{c}_{\mathbf{z}},
\label{equZA}
\end{equation}
where $\mathbf{z}$ denotes $[z_1,...,z_i]$, $\mathbf{z}^2$ denotes $[z^2_1,...,z^2_i]$, $\mathbf{z}_{i \neq j}$ denotes the vector whose elements are $ {z}_{i}{z}_{j}$ for all $i \neq j$, $\mathbf A_{\mathbf{z}}$ in block matrix can be rewritten as:
\begin{equation}      
\mathbf{A}_{\mathbf{z}} =
\left(                
  \begin{array}{ccc} 
    \mathbf{ A}^{(1)} &\mathbf{ A}^{(2)} &\mathbf{ A}^{(3)}\\  
    \mathbf{ A}^{(4)} &\mathbf{ A}^{(5)} &\mathbf{ A}^{(6)}\\
    \mathbf{ A}^{(7)} &\mathbf{ A}^{(8)} &\mathbf{ A}^{(9)}\\
  \end{array}
\right)
\label{equABlock}
\end{equation}
and $\mathbf{c}_{\mathbf{z}}$ as:
\begin{equation}      
\mathbf{c}_{\mathbf{z}} =
\left(                
  \begin{array}{c} 
    \mathbf{ c}^{(1)}\\  
    \mathbf{ c}^{(2)} \\ 
    \mathbf{ c}^{(3)} \\ 
  \end{array}
\right).
\label{equCBlock}
\end{equation}
Then, we have:
\begin{align}
\mathbf{z} = \mathbf{ A}^{(1)} \mathbf{\hat z} + \mathbf{ A}^{(2)} \mathbf{\hat{z}}^{2} + \mathbf{ A}^{(3)} \mathbf{\hat z}_{i \neq j} + \mathbf{ c}^{(1)}, \label{zAhatz1} \\
\mathbf{z}^{2} = \mathbf{ A}^{(4)} \mathbf{\hat z} + \mathbf{ A}^{(5)} \mathbf{\hat{z}}^{2} + \mathbf{ A}^{(6)} \mathbf{\hat{z}}_{i \neq j} + \mathbf{ c}^{(2)}. \label{zAhatz2}
\end{align}

So we can write for each ${z}_i$:
\begin{align}      
{z_i} = \sum_{j}({A}_{i,j}^{(1)} {\hat{z}}_{j}) + \sum_{j}({ A}_{i,j}^{(2)} {\hat{z}^{2}}_{j}) + ({\mathbf  A}_{i,:}^{(3)} \hat{\mathbf z}_{i \neq j})+ { c}_{i}^{(1)}, \label{eachzi1}\\
z_i^{2} = \sum_{j}({A}_{i,j}^{(4)} {\hat{z}}_{j}) + \sum_{j}({ A}_{i,j}^{(5)} {\hat{z}^{2}}_{j}) + ({ \mathbf A}_{i,:}^{(6)} \hat{\mathbf z}_{i \neq j}) + { c}_{i}^{(2)}. \label{eachzi2}
\end{align}

Squaring Eq. \eqref{eachzi1}, we have:
\begin{equation}      
\begin{aligned}
{z}_{i}^{2} = &\underbrace{(\sum_{j}({ A}_{i,j}^{(2)} {\hat{z}^{2}}_{j}))^{2}}_{\mathbf{(a)}} + \underbrace{(\sum_{j}({A}_{i,j}^{(1)} {\hat{z}}_{j}))^{2}}_{\text{(b)}} + \underbrace{ ({\mathbf A}_{i,:}^{(3)} \hat{\mathbf z}_{i \neq j})^2}_{\text{(c)}} +  ({ c}_{i}^{(1)})^{2} +.....
\label{squaringz1}
\end{aligned}
\end{equation}

It is notable that Eq. \eqref{eachzi2} and Eq. \eqref{squaringz1} are derived from Eq. \eqref{equTA} which holds for arbitrary $\mathbf{x}$, then by the assumption (ii) in Theorem \ref{theory1} that $\mathbf{f}$ is a bijective mapping from $\mathbf{z}$ to $\mathbf{x}$, and thus Eq. \eqref{eachzi2} and Eq. \eqref{squaringz1} must holds for $\mathbf{z}$ everywhere. Then by the fact that the right sides of the Eq. \eqref{eachzi2} and Eq. \eqref{squaringz1} are both polynomials with finite degree, we have each coefficients of the two polynomials must be equal. In more detail, for the term (a) in Eq. \eqref{squaringz1}:
\begin{equation}      
{(\sum_{j}({ A}_{i,j}^{(2)} {\hat{z}^{2}}_{j}))^{2}}= \sum_{j}({A}_{i,j}^{(2)})^2 {\hat{z}^{4}}_{j}+ \sum_{j\neq j'}( 2{A}_{i,j}^{(2)}{A}_{i,j'}^{(2)} {\hat{z}^{2}}_{j}{\hat{z}^{2}}_{j'}).
\end{equation}
Compared with Eq. \eqref{eachzi2}, since there is no term ${\hat{z}^{4}}_{j}$ in Eq. \eqref{eachzi2}, we must have that:
\begin{equation}
   \mathbf{A}^{(2)} = 0.
\end{equation}
For the term (c) in Eq. \eqref{squaringz1}:
\begin{equation}      
 { ({\mathbf A}_{i,:}^{(3)} \hat{\mathbf z}_{i \neq j})^2} = { \sum_{j^{'},i \neq j}({A}_{i,j'}^{(3)})^2( \hat{z}_{i} \hat{z}_{j})^2} + ...
\end{equation}
Compared with Eq. \eqref{eachzi2}, since there is no term $( \hat{z}_{i} \hat{z}_{j})^2$ in Eq. \eqref{eachzi2}, we must have that:
\begin{equation}
\mathbf{A}^{(3)}  = 0
\end{equation}
{
As a result, Eq. \eqref{zAhatz1} becomes:
\begin{align}      
\mathbf{z} = \mathbf{ A}^{(1)} \mathbf{\hat z} + \mathbf{ c}^{(1)}.
\label{lineartrans}
\end{align}
The above equation indicates the latent causal variables can be recovered up to linear transformation.}

{{\bf{Step III}} We then show that the linear transformation matrix $\mathbf{ A}^{(1)}$ in Eq. \eqref{lineartrans} must be a permutation matrix. As we mentioned in step I, for the Gaussian noise variables $\mathbf{n}$, we have:
\begin{align}      
\mathbf{n} = \mathbf{P} \mathbf{\hat n} + \mathbf{ c}_n,\label{nhatn}
\end{align}
where $\mathbf{P}$ denote the permutation matrix with scaling. With Eq. \eqref{nhatn} and Eq. \eqref{appenx:n2z}, the Eq. \eqref{lineartrans} can be rewritten as follows:
\begin{align}      
&\mathbf{B}(\mathbf{P} \mathbf{\hat n} + \mathbf{ c}_n) = \mathbf{ A}^{(1)} (\mathbf{\hat B}  \mathbf{\hat n}) + \mathbf{ c}^{(1)}. \label{lineartransplusn1}\\
\Rightarrow &(\mathbf{B}\mathbf{P}-\mathbf{ A}^{(1)} \mathbf{\hat B})\mathbf{\hat n} = \mathbf{ c}^{(1)}-\mathbf{B}\mathbf{ c}_n,
\label{lineartransplusn}
\end{align}
where $\mathbf{B} =(\mathbf{I}-\boldsymbol{\lambda}^{T}(
    \mathbf{u}
    ))^{-1} $ and $\mathbf{\hat B}= (\mathbf{I}-\boldsymbol{\hat \lambda}^{T}(
    \mathbf{u}
    ))^{-1}$. By differentiating Eq. \eqref{lineartransplusn} with respect to $\mathbf{\hat n}$, we have:
\begin{align}      
&(\mathbf{B}\mathbf{P}-\mathbf{ A}^{(1)} \mathbf{\hat B})\mathbf{I}=\mathbf{0},
\end{align}
where $\mathbf{I}$ denote the identity matrix, which implies that:
\begin{align}      
&\mathbf{B}\mathbf{P}=\mathbf{ A}^{(1)} \mathbf{\hat B},
\label{bpab}
\end{align}
that is,
\begin{align}      
(\mathbf{I}-\boldsymbol{\lambda}^{T}(
    \mathbf{u}
    ))^{-1}\mathbf{P}=\mathbf{ A} (\mathbf{I}-\boldsymbol{\hat \lambda}^{T}(
    \mathbf{u}
    ))^{-1},
\end{align}
for convenience, we use $\mathbf{ A}$ to replace $\mathbf{ A}^{(1)} $ here and in the following. By inverting both sides of the equation, we have
\begin{align}      
\mathbf{P}^{-1}(\mathbf{I}-\boldsymbol{\lambda}^{T}(
    \mathbf{u}
    ))=(\mathbf{I}-\boldsymbol{\hat \lambda}^{T}(
    \mathbf{u}
    ))\mathbf{ A}^{-1}.
\label{bpab1}
\end{align}
Without loss of generality, suppose that $\boldsymbol{\lambda}^{T}$ and $\boldsymbol{\hat \lambda}^{T}$ denote lower triangular matrices, respectively. 
\paragraph{Elements above the diagonal} We can see from Eq. \eqref{bpab1} that since the left is a lower triangular matrix, and $(\mathbf{I}-\boldsymbol{\hat \lambda}^{T}(
    \mathbf{u}
    ))$ in the right is also a lower triangular matrix, $\mathbf{ A}^{-1}$ must be a lower triangular matrix. That is, elements above the diagonal in matrix $\mathbf{ A}^{-1}$ are equal to 0.

\paragraph{The diagonal elements}  Assume $\mathbf{P}^{-1}$ to be diagonal with elements $s_{1,1},s_{2,2},s_{3,3},...$. By expanding the both sides of Eq. \eqref{bpab1}, we have:
\begin{equation}
\begin{aligned}
 &\left(                
  \begin{array}{cccc} 
    s_{1,1}&0&0&...\\  
    -s_{2,2}\lambda_{1,2}(\mathbf{u})&s_{2,2}&0&...\\ 
    -s_{3,3}\lambda_{1,3}(\mathbf{u})&-s_{3,3}\lambda_{2,3}(\mathbf{u})&s_{3,3}&...\\ 
    .&.&.&...\\
  \end{array}
\right) \\ & = 
 \left(                
  \begin{array}{cccc} 
    A_{1,1}&0&0&...\\  
    A_{2,1}-A_{1,1}\hat \lambda_{1,2}(\mathbf{u})&A_{2,2}&0&...\\ 
    A_{3,1}-\sum_{i=1}^{2}A_{i,1}\hat\lambda_{i,3}(\mathbf{u})&A_{3,2}-A_{2,2}\hat\lambda_{2,3}(\mathbf{u})&A_{3,3}&...\\ 
    .&.&.&...\\
  \end{array}
\right)
.\label{bpexpand}
\end{aligned}
\end{equation}
Comparing the left and the right of Eq. \eqref{bpexpand}, we can see that the diagonal elements of $\mathbf{ A}^{-1}$ must be $s_{1,1},s_{2,2},s_{3,3},...$.

 \paragraph{Elements below the diagonal} By comparison between the left and right of \eqref{bpexpand}, for each element ${ A}_{i,j}$ where $i > j$ in the right of Eq. \eqref{bpexpand}, we have: ${ A}_{i,j} - \sum_{i'<i}{\mathbf{ A}_{i',j} {\hat \lambda_{i',j}}(\mathbf{u})} = s_{j,j}\lambda_{j,i}(\mathbf{u})$. Then by assumption \ref{itm:lambda}, we have ${ A}_{i,j} = 0$.
 
As a result, the matrix $\mathbf{A}^{-1}$ is the same as the matrix $\mathbf{P}^{-1}$. That is,
\begin{equation}      
\mathbf{z} = \mathbf{P} \mathbf{\hat z} + \mathbf{c}_\mathbf{z}.
\end{equation}
}


\subsection{Understanding Assumptions in Theorem 4.1} \label{underassu}

\paragraph{Gaussian assumption on the latent noise variables} One of our model assumption is enforcing Gaussian distribution on the latent noise variables. Note that the assumptions of nonlinear ICA \citep{khemakhem2020variational} on the noise could be broad exponential family distribution, \eg, Gaussian distributions, Laplace, Gamma distribution, and so on. This work considers Gaussian distribution, mainly because it can be straightforwardly implemented for the re-parameterization trick in VAE. 2) It is convenient for simplifying the proof process. It is worthwhile and promising to extend the Gaussian assumption to exponential family for future work.
\paragraph{Linear model assumption for the latent causal variables} As the first work to discuss the relation between the change of causal influences and identifiability of latent causal model, this work simply considers linear models for the latent causal variables, because it can be directly parameterized, which helps us to analyze the challenge of identifiability and how to handle it. In addition, together with Gaussian noise, we arrive at the same probabilistic model as linear Gaussian Bayesian networks, which is a well-studied model as mentioned in Eq. \eqref{lineargaussian}. We expect that the proposed weights-variant linear models can motivate  more general functional classes in future work, \eg, nonlinear additive noise models.
\paragraph{Changes of weights} We allow changes of weights (\ie, causal influences) among latent causal variables across $\mathbf{u}$ to handle the transitivity problem. We argue that this may be not a necessary condition for identifiability (Existing methods, including sparse graph structure and temporal information, could be regarded as feasible
ways to handle the transitivity.), but it is a sufficient condition to handle the transitivity, and provide a new research line for causal representation learning. In fact, exploring the change of causal influences is not new in observed space \citep{ghassami2018multi, CDNOD_20}. In addition, identifying latent causal variables by randomly chosen unknown hard intervention can also be regarded as a special change of causal influences \citep{brehmer2022weakly}. From this viewpoint, we argue that exploring the change of causal influences for identifying latent causal representation may be a promising way and not too restricted in reality.

We emphasize that changes of weights among latent variables across different values of $\mathbf{u}$ are essential for achieving full identifiability. However, this does not mean that all weights must vary continuously or at every point in $\mathbf{u}$. Some weights may remain unchanged at certain values of $\mathbf{u}$, as long as they exhibit variation at other values of $\mathbf{u}$. This flexibility is both theoretically sufficient and practically relevant. See Section~\ref{localchanges} for experimental evidence and further discussion on this point.

\subsection{The Proof of Corollary \ref{corollary}}
\label{corollary:sketch}
Theorem \ref{theory1} has shown that the latent causal variables $\mathbf{z}$ can be identified up to trivial permutation and linear scaling. Hence, the identifiability of the causal structure in latent space can be reduced to the identifiability of the causal structure in observed space. Moreover, we need to show that the linear scaling does not affect theoretical identifiability of the causal structure. 

For the identifiability of the causal structure in observed space from heterogeneous data, fortunately, we can leverage the results from \citet{CDNOD_20}. Corollary 4.2 relies on the Markov condition and faithfulness assumption and the assumption that the latent change factor (i.e., causal strength in the linear case) can be represented as a function of the domain index $\mathbf{u}$. Hence, it relies on the same assumptions as that in \citet{CDNOD_20}.

It has been shown in \citet{CDNOD_20} that if the joint distribution over $\mathbf{z}$ and $\mathbf{u}$ ($\mathbf{z}$ and $\mathbf{u}$ are observed variables here) are Markov and faithful to the augmented graph, then the causal structure over $\mathbf{z}\cup \mathbf{u}$ can be identified up to the Markov equivalence class, by making use of the conditional independence relationships.

Next, we show that the Markov equivalence class over $\mathbf{z}$ is also identifiable. Denote by $M$ the Markov equivalence class over $\mathbf{z} \cup \mathbf{u}$, and by $M_z$ the Markov equivalence class over $\mathbf{z}$. Then after removing variable $\mathbf{u}$ in $M$ and its edges, the resulting graph (denoted by $M’_z$) is the same as $M_z$. This is because of the following reasons. First, it is obvious that $M’_z$ and $M_z$ have the same skeleton. Second, in this paper, $\mathbf{u}$ has an edge over every $z_i$  when considering $\mathbf{n}$ as latent noise variables, because all causal strength and noise distributions change with $\mathbf{u}$. Hence, there is no v-structure over $\mathbf{u}$, $z_i$, and $z_j$, so it is not possible to have more oriented edges in $M’_z$. Therefore, $M’_z$ and $M_z$ have the same skeleton and the same directions. 

Moreover, the conditional independence relationships will not be affected by the linear scaling of the variables, so the conditional independence relationships still hold in the identified latent variables. Furthermore, for linear-Gaussian models, independence equivalence is the same as distributional equivalence. This is because of the following reasons. First, since we are concerned with linear Gaussian models over the latent variables, the identifiability up to equivalence class also holds for score-based methods that use the likelihood of data as objective functions. This is because independence equivalence, i.e., two DAGs have the same conditional independence relations, is the same as distributional equivalence for linear-Gaussian models, i.e., two Bayesian networks corresponding to the two DAGs can define the same probability distribution. That is to say, score-based methods also find the structure based on independence relations implicitly. Because the scaling does not change the independence relations, it will also not affect the identifiability of the graph structure.

Therefore, the causal structure among latent variables $\mathbf{z}$ can be identified up to the Markov equivalence class.

\subsection{The Proof of Corollary \ref{corollary1}}
\label{corollary:dag}
To prove this corollary, let us introduce ICM first.
\begin{definition}
(Independent Causal Mechanisms). In a causally sufficient system, the causal modules, as well as their included parameters, change independently across domains.
\end{definition}
Here we follow \citet{ghassami2018multi} to define ICM principle. Another slightly different definition appears in \citet{peters2017elements,scholkopf2021toward}. Both two definitions imply the same principle that the causal modules change independently across $\mathbf{u}$. According to ICM principle, in the causal direction, the causal modules, as well as their included parameters, change independently across $\mathbf{u}$. However, such independence generally does not hold in the anti-causal direction \citep{ghassami2018multi, CDNOD_20}. For example, consider $z_1$ and $z_2$ in Figure \ref{fig:problem}. According to model definition in Eq. \eqref{eq:Generative1}- Eq. \eqref{eq:Generative}, for causal direction $z_1 \rightarrow z_2$, we have:
\begin{equation}
 n_i: \sim {\cal N}( {\beta_{i,1}(\mathbf{u})}, {\beta_{i,2}(\mathbf{u})}), \qquad z_1: = n_1,  \qquad   z_2: = \lambda_{1,2}({\mathbf{u}})z_1+n_2.
\end{equation}
For the reverse direction, we have
\begin{equation}
\begin{split}
   & z_1 = \lambda'_1(\mathbf{u}) z_2 + n_1', \qquad \qquad \qquad \qquad \ z_2 = n_2' \\
  &  \lambda'_1(\mathbf{u}) = \frac{1}{\lambda_{1,2}(\mathbf{u})}, \quad  n_1' = -\frac{1}{\lambda_{1,2}(\mathbf{u})}n_2 \qquad
    n_2' = \lambda_{1,2}({\mathbf{u}})z_1+n_2
\end{split} 
\end{equation}
In this case, the module $p(z_2)$ is dependent of the module $p(z_1|z_2)$, \ie, both modules depend on the same parameter $\lambda_{1,2}({\mathbf{u}})$, which violates ICM principle. Clearly, the scaling indeterminacy for the recovered latent variables does not affect such dependence. As a result, regardless of the scaling, with the help of ICM, we can fully identify the acyclic causal structure among latent variables $\mathbf{z}$.

\subsection{The proof of Theorem \ref{theory2}}
Throughout the proof process in Section \ref{prooflinear}, Eq. \eqref{bpexpand} holds without the need for assumption \ref{itm:lambda}. Note that this also implies that $\mathbf{ A}^{-1}$ is a lower triangular matrix. Then consider the following two cases. 
\begin{itemize}
    \item For the case where $z_i$ is a root node or all weights on all paths from parent nodes of $z_i$ to $z_i$ meet assumption \ref{itm:lambda}, by using assumption \ref{itm:lambda} and by comparison both sides of Eq. \eqref{bpexpand}, we have: for all $j<i$, ${A}_{{i,j}}=0$, which implies that we can obtain that $A^{-1}_{i,i}z_i=\hat z_i + c'_i$.
    \item If there exists an unchanged weight on all paths from parent nodes of $z_i$ to $z_i$ across $\mathbf{u}$, we demonstrate that it is always possible to construct an alternative solution, which is different from the true $\mathbf{z}$, but capable of generating the same observations $\mathbf{x}$. Suppose that for $z_i$, there is an unchanged weight $\lambda_{j,i}$ across $\mathbf{u}$, related to the parent node $z_{j}$. Then, we can always construct new latent variables $\mathbf{z'}$ as: for all $k \neq i$, ${z'}_k=z_k$, and $z'_i=z_i-\lambda_{j,i}z_j$. Given this, we can construct a matrix $\mathbf{M}$, so that $\mathbf{M}\mathbf{z'}=\mathbf{z}$.
It is clear the matrix $\mathbf{M}$ is invertible. In addition, all the weights of the matrix $\mathbf{M}$ are constant and thus do not depend on $\mathbf{u}$. As a result, we can construct a mapping from $\mathbf{z}'$ to $\mathbf{x}$, \eg, $\mathbf{f}(\mathbf{M}\mathbf{z}')$, which is invertible and do not depend on $\mathbf{u}$, and can create the same data $\mathbf{x}$ generated by $\mathbf{f}(\mathbf{z})$. Therefore, the alternative solution $\mathbf{z}'$ can lead to a non-identifiability result.
\item For case that there exists an weight $\lambda_{j,i}(\mathbf{u}) + b$, we can always construct new latent variables $\mathbf{z'}$ as: for all $k \neq i$, ${z'}_k=z_k$, and $z'_i=z_i-bz_j$. Given this, we can construct a matrix $\mathbf{M}$, so that $\mathbf{M}\mathbf{z'}=\mathbf{z}$.
It is clear the matrix $\mathbf{M}$ is invertible. In addition, all the weights of the matrix $\mathbf{M}$ are constant and thus do not depend on $\mathbf{u}$. Finally, we can construct a mapping from $\mathbf{z}'$ to $\mathbf{x}$, \eg, $\mathbf{f}(\mathbf{M}\mathbf{z}')$, which do not depend on $\mathbf{u}$, and can create the same data $\mathbf{x}$.
\end{itemize}


\subsection{Implementation Details}
\label{appendix: details}

\subsubsection{Experiments for Figure \ref{fig:synthetic}} 
\label{appendix: details1}
\paragraph{Data} For experimental results of Figure \ref{fig:synthetic} the number of segments (\ie, $\mathbf{u}$) $M$ is 10, 20, 30, 40 and 50 respectively. For each segment, the number of latent causal variables is 2 and the sample size is 1000. We consider the following structural causal model
\begin{align}
&n_1 := {\cal N}({\beta}_{1,1}(\mathbf{u}), {\beta}_{1,2}(\mathbf{u})), 
\qquad n_2 := {\cal N}({\beta}_{2,1}(\mathbf{u}), {\beta}_{2,2}(\mathbf{u})), \label{datagene1}\\
&{z_1} := n_1, \qquad\qquad \qquad \qquad \quad \ \ {z_2} := \lambda_{1,2}(\mathbf{u}) {z_1} +n_2,\label{datagene2}
\end{align}
where we sample the mean ${\beta}_{i,1}(\mathbf{u})$ and variance ${\beta}_{i,2}(\mathbf{u})$ from uniform distributions $[-2,2]$ and $[0.01,3]$, respectively. We sample the weights $\lambda_{1,2}(\mathbf{u})$ from uniform distributions $[0.1, 2]$. We sample latent variable z according to Eqs. \ref{datagene1} and \ref{datagene2}, and then mix them using a 2-layer multi-layer perceptron (MLP) to generate observed $\mathbf{x}$.
\paragraph{Network and Optimization} For all methods, we used a encoder, \ie, 3-layer fully connected network with 30 hidden nodes and Leaky-ReLU activation functions, and decoder, \ie, 3-layer fully connected network with 30 hidden nodes and Leaky-ReLU activation functions. We also use 3-layer fully connected network with 30 hidden nodes and Leaky-ReLU activation functions. For optimization, we use Adam optimizer with learning rate $1e-3$. For hyper-parameters, we set $\beta=4$ for the $\beta$-VAE. For CausalVAE, we set the hyper-parameters as suggested in \citet{yang2020causalvae}.
\subsubsection{Experiments for Figures \ref{fig:synthetic:gamma} and \ref{fig:synthetic:b}} \label{expbc}
\paragraph{Data} For experimental results of Figures \ref{fig:synthetic:gamma} and \ref{fig:synthetic:b}, the number of segments is $30$ and sample size is 1000, while the number of latent causal variables is 2,3,4,5 respectively. For 2-dimensional latent causal variables, the setting is the same as experiments for Figure \ref{fig:synthetic} (a). For 3-dimensional case, we consider the following structural causal model:
\begin{align}
n_1 &:= {\cal N}({\beta}_{1,1}(\mathbf{u}), {\beta}_{1,2}(\mathbf{u})), 
\quad n_2 := {\cal N}({\beta}_{2,1}(\mathbf{u}), {\beta}_{2,2}(\mathbf{u})), \quad n_3 := {\cal N}({\beta}_{3,1}(\mathbf{u}), {\beta}_{3,2}(\mathbf{u})), \label{datagene31}\\
{z_1} &:= n_1, \quad\qquad \qquad \qquad \quad \  {z_2} := \lambda_{1,2}(\mathbf{u}) {z_1} +n_2,   \qquad \quad \  {z_3} := \lambda_{2,3}(\mathbf{u}) {z_2} +n_3. \label{datagene32}
\end{align}
For 4-dimensional case, we consider the following structural causal model:
\begin{align}
&n_1 := {\cal N}({\beta}_{1,1}(\mathbf{u}), {\beta}_{1,2}(\mathbf{u})), 
\qquad n_2 := {\cal N}({\beta}_{2,1}(\mathbf{u}), {\beta}_{2,2}(\mathbf{u})), \\ 
&n_3 := {\cal N}({\beta}_{3,1}(\mathbf{u}), {\beta}_{3,2}(\mathbf{u})), 
\qquad  n_4 := {\cal N}({\beta}_{4,1}(\mathbf{u}), {\beta}_{4,2}(\mathbf{u})), \\
&{z_1} := n_1, \quad\qquad \qquad \qquad \quad \quad \ \ {z_2} := \lambda_{1,2}(\mathbf{u}) {z_1} +n_2,  \\
& {z_3} := \lambda_{2,3}(\mathbf{u}) {z_2} +n_3, \qquad \ \qquad
 {z_4} := \lambda_{2,4}(\mathbf{u}) {z_2} +n_4.
\end{align}
For 5-dimensional case, we consider the following structural causal model:
\begin{align}
&n_1 := {\cal N}({\beta}_{1,1}(\mathbf{u}), {\beta}_{1,2}(\mathbf{u})), 
\quad n_2 := {\cal N}({\beta}_{2,1}(\mathbf{u}), {\beta}_{2,2}(\mathbf{u})), \\ 
&n_3 := {\cal N}({\beta}_{3,1}(\mathbf{u}), {\beta}_{3,2}(\mathbf{u})), 
\quad  n_4 := {\cal N}({\beta}_{4,1}(\mathbf{u}), {\beta}_{4,2}(\mathbf{u})), \quad n_5 := {\cal N}({\beta}_{5,1}(\mathbf{u}), {\beta}_{5,2}(\mathbf{u})), \\
&{z_1} := n_1, \quad\quad \qquad \qquad \quad \quad \ \ {z_2} := \lambda_{1,2}(\mathbf{u}) {z_1} +n_2,  \\
& {z_3} := \lambda_{2,3}(\mathbf{u}) {z_2} +n_3, \qquad \ \quad{z_4} := \lambda_{2,4}(\mathbf{u}) {z_2} +n_4, 
\qquad \quad \ {z_5} := \lambda_{4,5}(\mathbf{u}) {z_4} +n_5.\label{data5}
\end{align}
Again, for all these Eqs \ref{datagene31}-\ref{data5}, we sample the mean ${\beta}(\mathbf{u})_{i,1}$ and variance ${\beta}(\mathbf{u})_{i,2}$ from uniform distributions $[-2,2]$ and $[0.01,3]$, respectively. We sample the weights $\lambda_{i,j}(\mathbf{u})$ from uniform distributions $[0.1, 2]$. We sample latent variable $\mathbf{z}$ according to these equations, and then mix them using a 2-layer MLP to generate observed $\mathbf{x}$.
\paragraph{Network and Optimization} The network architecture, optimization and hyper-parameters are the same as used for experiments of Figure \ref{fig:synthetic} (a), except for the 5-dimensional case where we use 40 hidden nodes for each linear layer.

\subsubsection{Experiments for Table \ref{table}}
\label{appendix: Table}
For experimental results of Table \ref{table}, the number of segments is $30$ and the sample size is 1000, and the number of latent causal variables is 3. For the case of the matching assumptions, we consider the structural causal model as mentioned in Eqs. \eqref{datagene31}-\eqref{datagene32}. For the cases of uniform, Laplace and Gamma noise, we replace the original Gaussian noise by using uniform noise, Laplace and Gamma noise, respectively, while the remaining parts are the same as Eqs. \eqref{datagene31}-\eqref{datagene32}. The network architecture, optimization, and hyper-parameters are the same as used for experiments of Figure \ref{fig:synthetic} (a).

\subsubsection{Implementation of Enforcing Assumption (v)} 

To enforce assumption (v), which requires $\lambda_{i,j}(\mathbf{u} = \mathbf{0}) = 0$ in the prior model and $\lambda'_{i,j}(\mathbf{u} = \mathbf{0}) = 0$ in the posterior model, we augment the training process with \emph{synthetic pairs} that regularize model behavior at $\mathbf{u} = \mathbf{0}$, in the absence of observed data at this point. We construct synthetic pairs of the form: $\text{(input: } \mathbf{u} = \mathbf{0}, \quad \text{target: } 0\text{)}$, where the target value ``0'' is not associated with the observed variable $\mathbf{x}$, but instead reflects the expected output of a specific model component evaluated at $\mathbf{u} = \mathbf{0}$. Specifically, for the {prior model} Eq. \eqref{prior}, this corresponds to enforcing $\lambda_{i,j}(\mathbf{0}) = 0$. For the {inference model} Eq. \eqref{poster}, it enforces $\lambda'_{i,j}(\mathbf{0}) = 0$. These synthetic pairs are therefore not part of the data distribution $(\mathbf{u}, \mathbf{x})$, but rather encode theoretical constraints derived from assumption (v). We integrate the synthetic pairs into the training objective alongside observed data $(\mathbf{u}, \mathbf{x})$. For each pair, we apply an $L_1$ penalty on the model output at $\mathbf{u} = \mathbf{0}$: $\lVert \lambda_{i,j}(\mathbf{0}) \rVert_1$ for the prior model, $\lVert \lambda'_{i,j}(\mathbf{0}) \rVert_1$ for the inference model. These terms are added to the ELBO loss as shown in Eq.~\eqref{obj1}, enabling the model to conform to assumption (v).

\subsection{Detailed Results on fMRI}

\begin{figure}[h]
  \centering
\includegraphics[width=0.8\textwidth]{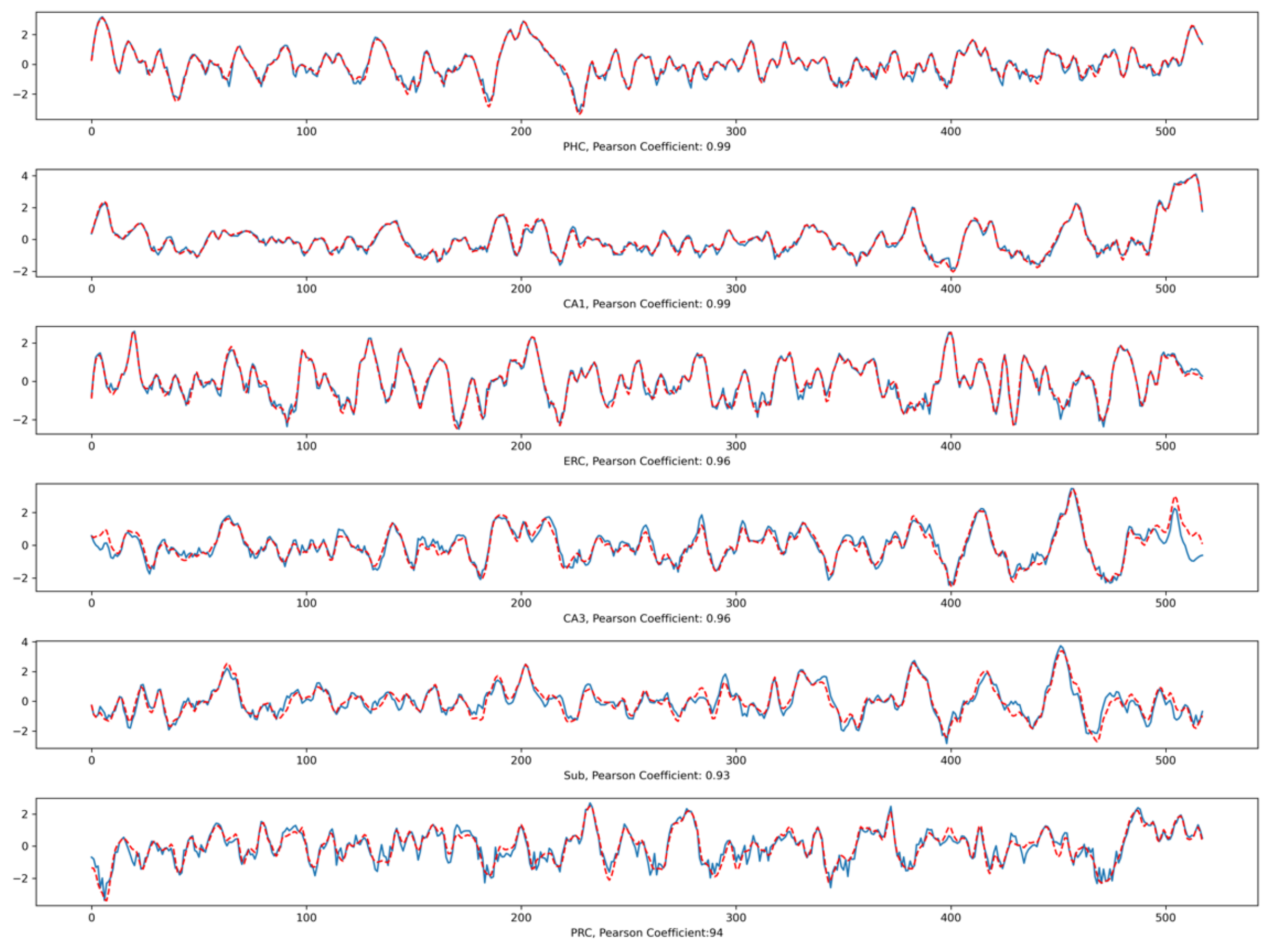}
  \caption{\small Recovered six signals (Blue) and the true ones (Red) within one day by \FancyName.}
  \label{fig:fmridata}
\end{figure}

\begin{figure}[h]
  \centering
\includegraphics[width=0.8\textwidth]{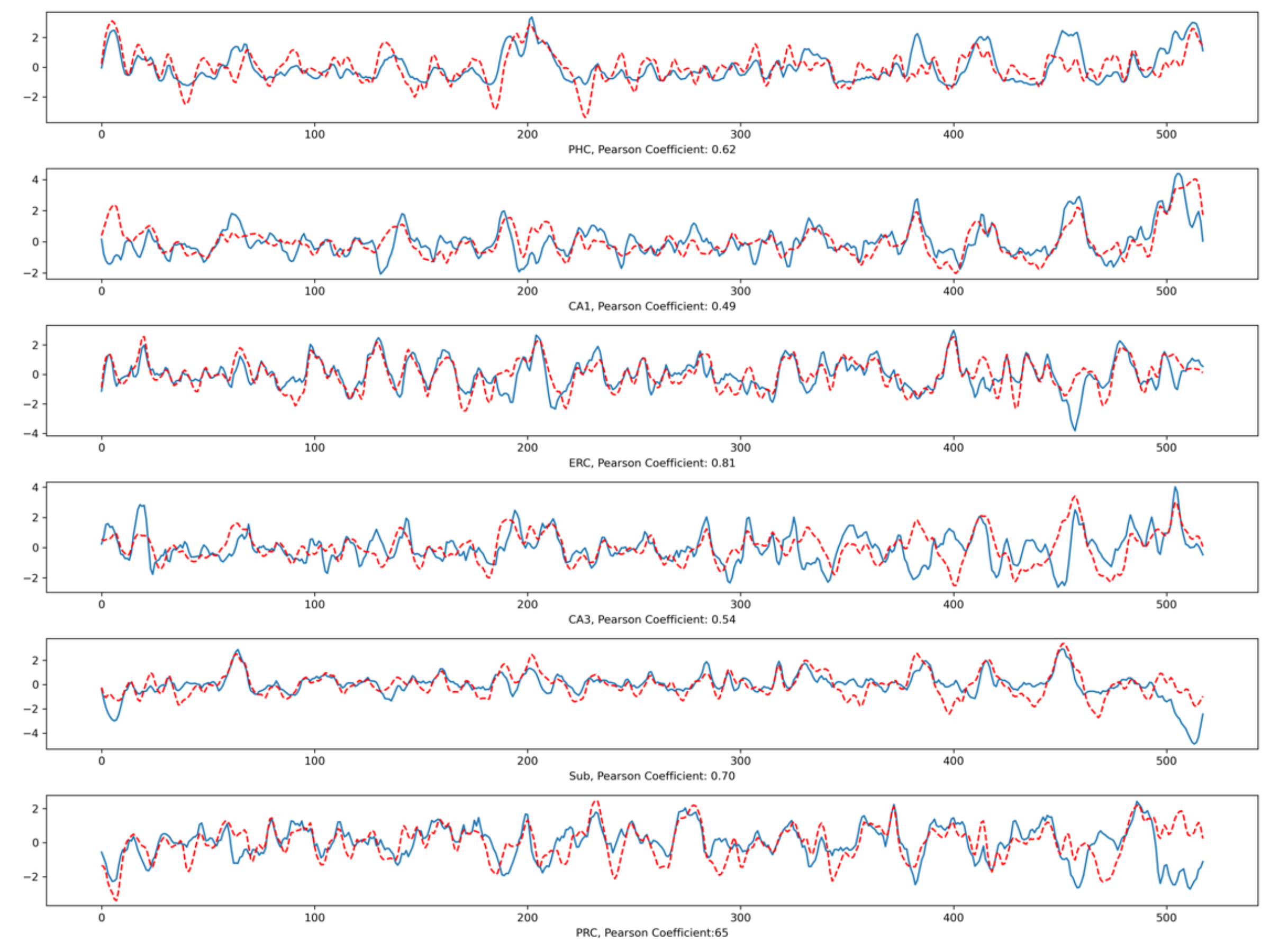}
  \caption{Recovered six signals (Blue) and the true ones (Red) within one day by iVAE.}
  \label{fig:ivaefmridata}
\end{figure}

\begin{figure}[h]
  \centering
\includegraphics[width=0.8\textwidth]{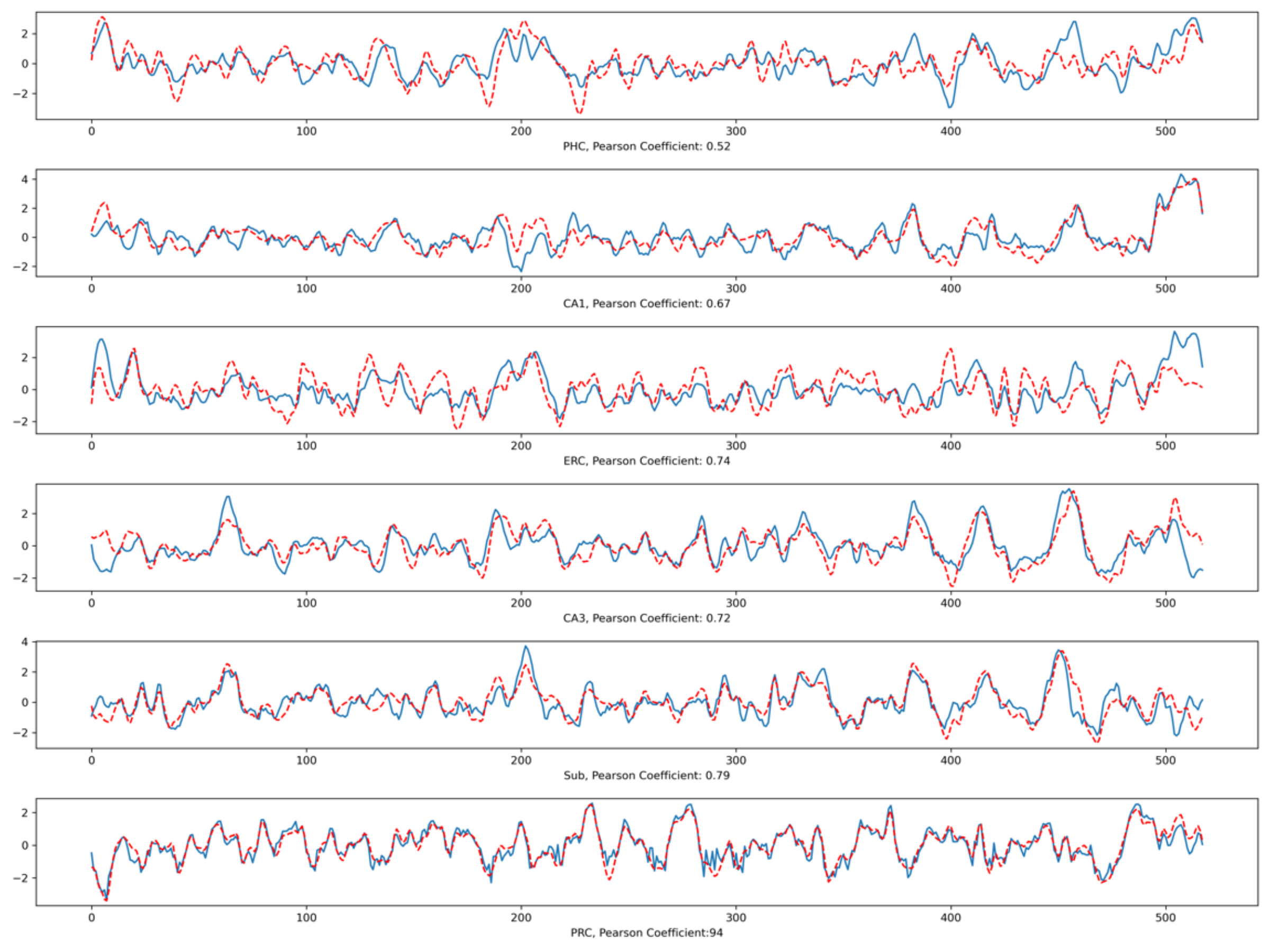}
  \caption{Recovered six signals (Blue) and the true ones (Red) within one day by VAE.}
  \label{fig:vaefmridata}
\end{figure}

\begin{figure}[h]
  \centering
\includegraphics[width=0.8\textwidth]{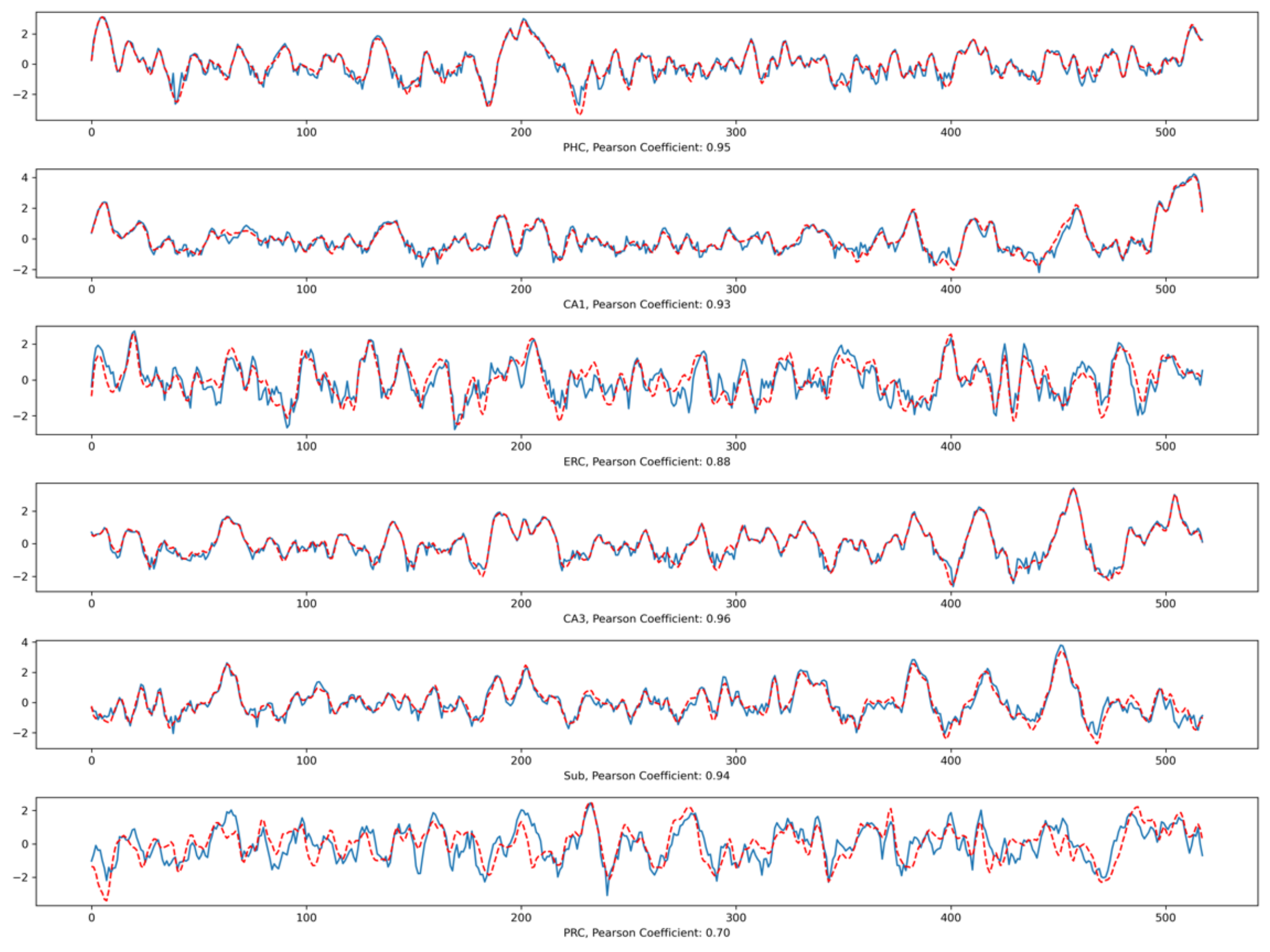}
  \caption{Recovered six signals (Blue) and the true ones (Red) within one day by $\beta$-VAE.}
  \label{fig:betavaefmridata}
\end{figure}

\begin{figure}[h]
  \centering
\includegraphics[width=0.8\textwidth]{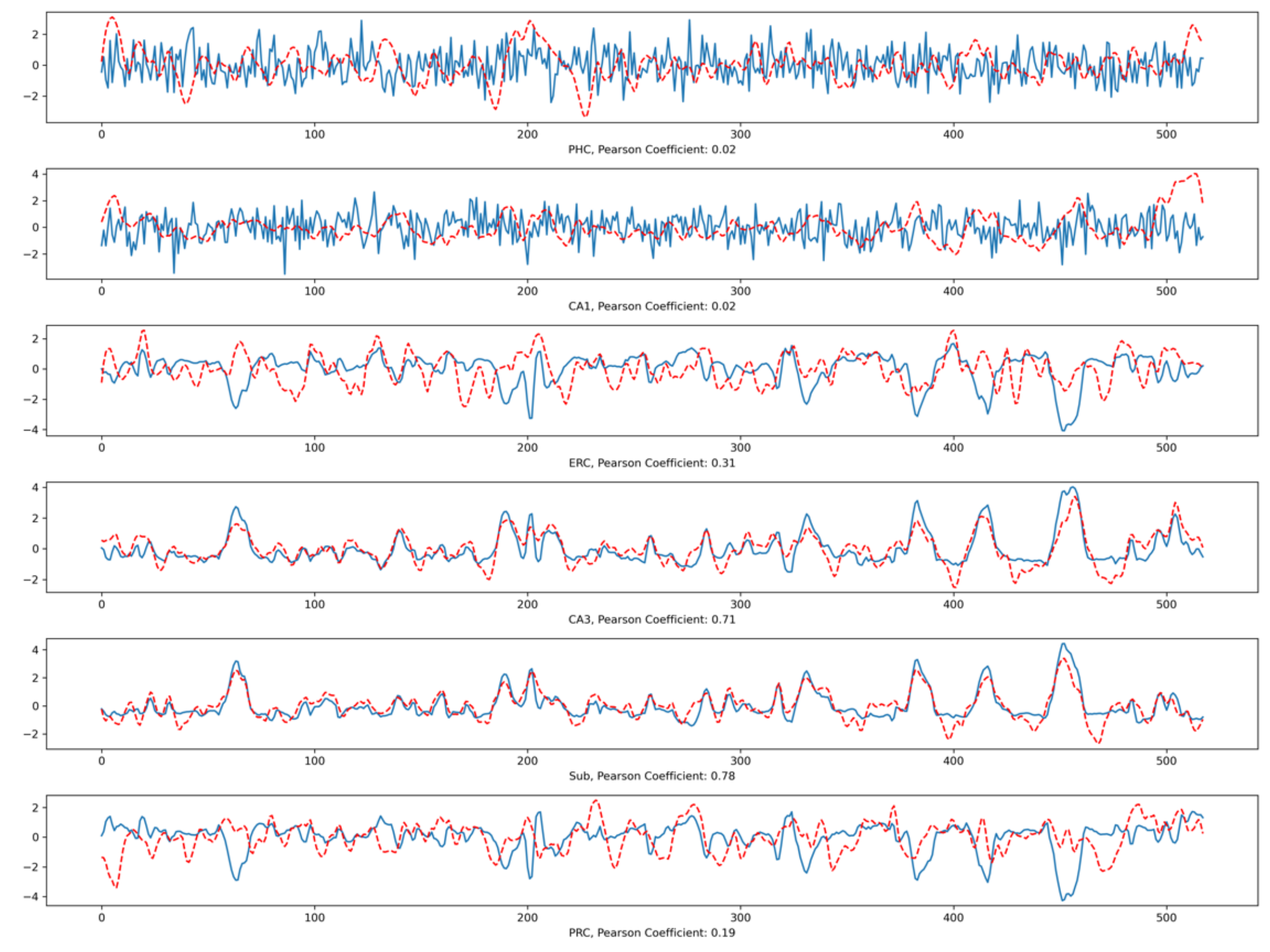}
  \caption{Recovered six signals (Blue) and the true ones (Red) by CausalVAE.}
  \label{fig:causalvaefmridata}
\end{figure}

Figures \ref{fig:fmridata}-\ref{fig:causalvaefmridata} show the recovered latent six signals (Blue) and the true ones (Red) within one day by the proposed \FancyName, iVAE, VAE, $\beta$-VAE, and CausalVAE, respectively. Regarding to implementation details, for fMRI data, again, we used the same network architecture for encoder (\ie, 3-layer fully connected network with 30 hidden nodes for each layer) and decoder (\ie, 3-layer fully connected network with 30 hidden nodes for each layer) parts in all these models. For prior model in the proposed \FancyName and iVAE, we use 3-layer fully connected network with 30 hidden nodes for each layer. We assign an 3-layer fully connected network with 30 nodes to generate the weights to model the relations among latent causal variables in the proposed \FancyName. For hyper-parameters, we set $\beta=4$, 25, 50 for $\beta$-VAE. For CausalVAE, we use the hyper-parameters setting as recommended in \citet{yang2020causalvae}, for generating observed $\mathbf{x}$, we use an invertible 2-layer multi-layer perceptron on the fMRI with different 5 random seeds.

\clearpage
\bibliography{sample}

\end{document}